\documentclass[twocolumn]{article}

\usepackage{arxiv}

\usepackage[numbers,square]{natbib}
\usepackage[utf8]{inputenc}	%
\usepackage[T1]{fontenc}	%
\usepackage{amsmath,amsfonts,amssymb,amsthm}
\usepackage[colorlinks = true,
            linkcolor = purple,
            urlcolor  = blue,
            citecolor = cyan,
            anchorcolor = black]{hyperref}
\usepackage{booktabs}
\usepackage{stfloats}
\usepackage{graphicx}
\usepackage[font=scriptsize, position=top]{subfig}
\usepackage[font=small,labelfont=bf]{caption}
\usepackage{url}

\usepackage{algorithm}
\usepackage{algpseudocode}

\newtheorem{property}{Property}
\newtheorem{remark}{Remark}

\usepackage{titlesec}
\titlespacing\section{0pt}{12pt plus 3pt minus 3pt}{1pt plus 1pt minus 1pt}
\titlespacing\subsection{0pt}{10pt plus 3pt minus 3pt}{1pt plus 1pt minus 1pt}
\titlespacing\subsubsection{0pt}{8pt plus 3pt minus 3pt}{1pt plus 1pt minus 1pt}

\usepackage{tikz}

\allowdisplaybreaks

\definecolor{lime}{HTML}{A6CE39}
\DeclareRobustCommand{\orcidicon}{
	\begin{tikzpicture}
	\draw[lime, fill=lime] (0,0)
	circle [radius=0.16]
	node[white] {{\fontfamily{qag}\selectfont \tiny ID}};
	\draw[white, fill=white] (-0.0625,0.095)
	circle [radius=0.007];
	\end{tikzpicture}
	\hspace{-2mm}
}
\foreach \x in {A, ..., Z}{\expandafter\xdef\csname orcid\x\endcsname{\noexpand\href{https://orcid.org/\csname orcidauthor\x\endcsname}
			{\noexpand\orcidicon}}
}

\title{Bridging Continuous-Time MAPF and Industrial AGV Fleet Coordination with Lazy Priority-Based Search}

\usepackage{xwatermark}
\newwatermark[firstpage,color=gray!90,angle=0,scale=0.25, xpos=0in,ypos=-5in,width=30in]{This work has been submitted to the IEEE for possible publication. Copyright may be transferred without notice, after which this version may no longer be accessible.}

\usepackage{authblk}

\author[1]{Federico Signorile\orcidA{}}
\author[1]{Raffaele Carli \orcidB{}}
\author[2]{Alessandro Bertogli}
\author[2]{Roberto Olmi}
\author[3]{Sven Koenig\orcidC{}}
\author[1]{Mariagrazia Dotoli \orcidD{}}

\affil[1]{Department of Electrical and Information Engineering, Politecnico di Bari, Bari, Italy}
\affil[2]{E80 Group S.p.A., Viano, Italy}
\affil[3]{Department of Computer Science, University of California, Irvine, CA, USA}

\begin{document}

\twocolumn[ %
  \begin{@twocolumnfalse} %

    \maketitle

    \begin{abstract}
    Continuous-time multi-agent path finding on roadmaps (MAPF$_R$) offers a principled way to coordinate robot fleets with geometric collision avoidance, yet its deployment on industrial automated guided vehicle (AGV) infrastructures remains challenging, as highly connected and redundant roadmaps amplify both safe-interval reasoning and multi-agent interaction handling. To bridge this gap, we formalize AGV-MAPF$_R$-CTC, a configuration-space variant of MAPF$_R$ with precomputed continuous-time conflicts that enables the reuse of MAPF$_R$ solvers for AGV fleet coordination with minimal changes. We then revisit continuous-time priority-based search (CPBS), identifying the boundary conditions required by its safe interval path planning (SIPP) low-level planner under priority-induced constraints and enabling earlier solution detection. As a central contribution, we analyze the structure of the priority search space, clarifying which agents are affected by each priority refinement and when replanning can be avoided. Building on this analysis, we introduce a novel algorithm, Lazy-CPBS, which materializes high-level nodes lazily: a child's plans are computed only when it is expanded, so siblings that are never expanded cost no low-level search. Experiments on standard MAPF benchmarks and on industrial AGV roadmaps imported from a proprietary digital-twin simulator show that CPBS-based solvers remain effective well beyond the agent counts reachable by optimal MAPF$_R$ solvers, while Lazy-CPBS substantially reduces CPBS overhead and offers the most consistent robustness/runtime trade-off across industrial layouts.
    \end{abstract}
    \vspace{0.35cm}

  \end{@twocolumnfalse} %
] %

\section{Introduction}
\label{sec:intro}

\subsection{From MAPF to Continuous-Time Roadmap Planning}
Multi-agent path finding (MAPF) has emerged as a central problem in both AI and robotics, providing algorithmic foundations for coordinating large fleets of robots under collision-avoidance constraints. Its techniques have been successfully applied in domains such as warehouse logistics \cite{wurman2008coordinating}, office service robots \cite{veloso2015cobots}, airport towing \cite{morris2016planning}, and video games \cite{sigurdson2018multi}, and have also inspired solutions to related problems including 3D pipe routing \cite{belov2020multi}, autonomous driving in urban environments \cite{li2023intersection}, and other complex routing and scheduling tasks.
In the classical MAPF formulation, the environment is discretized into a grid and agents move synchronously in discrete time steps, seeking non-colliding paths that minimize global objectives such as the sum of arrival times \cite{stern2019multifull, stern2019multi}. Over the last decade, this setting has led to a rich algorithmic landscape, ranging from optimal \cite{lam2025low} and bounded-suboptimal search \cite{li2021eecbs} to suboptimal methods that scale to hundreds or even thousands of agents in seconds~\cite{okumura2022priority, okumura2023lacam}. However, discrete-time, grid-based MAPF remains an abstraction: it disregards the continuous nature of time, the geometry of real roadmaps, and the kinematic constraints of industrial vehicles, limiting its direct applicability in real-world domains such as warehouses with automated forklifts and other car-like automated guided vehicles (AGVs).

To address these limitations, the continuous-time variant MAPF$_R$ has been proposed, in which agents move along a generic roadmap, collisions are detected geometrically rather than via grid-cell conflicts, and actions have arbitrary real-valued durations \cite{walker2018extended}. A key component in this context is safe interval path planning (SIPP) \cite{phillips2011sipp}, an A$^*$-based algorithm that performs continuous-time planning by encoding dynamic obstacles as time-varying safe intervals on vertices and edges. SIPP serves as the single-agent planner underlying most MAPF$_R$ algorithms, such as continuous-time conflict-based search (CCBS)~\cite{andreychuk2022multi}. CCBS and its recent refinements based on wait discretization \cite{li2025cbs} or improved branching rules \cite{combrink2025optimal} currently represent the main optimal and solution-complete baselines for MAPF$_R$.

\subsection{Industrial AGV Coordination and Roadmap Complexity}
Unlike autonomous mobile robots (AMRs), which navigate primarily through onboard intelligence~\cite{9247159}, industrial AGVs are designed to operate on fixed, predefined roadmaps~\cite{vis2006survey}.
Recent advances in MAPF$_R$ therefore appear as a natural algorithmic candidate for AGV fleet coordination, seemingly lowering the barrier to a direct integration of multi-agent planning into industrial systems.
In practice, however, most deployments still follow a decoupled architecture: each vehicle plans as if it were alone, while a separate traffic-management layer resolves interactions online~\cite{proia2026decentralized}.
Such controllers typically ensure safety and liveness through priority rules, local negotiation over shared resources~\cite{digani2015ensemble}, hierarchical control schemes~\cite{pratissoli2021hierarchical}, or segment-reservation policies~\cite{olmi2011efficient}, rather than through a globally coordinated multi-agent planner.
Although well established, these approaches can degrade in dense, customized warehouse layouts, where congestion, deadlocks, and suboptimal traffic patterns become increasingly frequent as fleet sizes grow~\cite{varambally2022mapf}.

A key reason why MAPF$_R$ techniques remain challenging to deploy at scale lies in the structure of industrial AGV roadmaps.
Such roadmaps are typically highly connected and redundant, providing strong connectivity, broad coverage of the free space, and multiple alternative routes between most locations~\cite{digani2015ensemble, 4059098}.
While this redundancy is desirable for robustness and for handling local disturbances, it amplifies the combinatorial complexity of both layers in MAPF$_R$ solvers.
At the single-agent level, many methods rely on SIPP, which represents temporal constraints by maintaining, for each vertex and edge, the time windows in which it can be safely occupied.
Large and edge-dense industrial roadmaps inflate the number of vertex/edge safe intervals, raising memory and bookkeeping overhead.
Moreover, redundancy offers many alternative outgoing edges that traverse similar regions of the roadmap, increasing the branching factor that the low-level planner must consider.
At the multi-agent level, the same structural property induces dense clusters of temporally overlapping interactions, which in CCBS-like approaches translates into an explosion in the number of conflicts to detect and resolve through branching and repeated replanning.
As a consequence, the scalability observed on standard MAPF$_R$ benchmarks often fails to transfer to real industrial layouts, and practitioners still overwhelmingly rely on faster coordination schemes.

\subsection{Priority-Based Algorithms and the Cost of Replanning}
Within this landscape, priority-based planning offers a natural compromise between the speed of decoupled coordination and the adaptability of conflict-driven MAPF solvers.
In classical prioritized planning, a total order over the agents is fixed in advance, and each agent plans its path while treating all higher-priority agents as moving obstacles \cite{silver2005cooperative}.
This yields a simple and fast framework, but it is well known to be incomplete: a fixed priority ordering may fail to admit any solution even when an alternative ordering would succeed \cite{vcap2015complete}.
To mitigate this limitation, restart-based variants repeatedly modify the priority order, either by random resampling~\cite{bennewitz2001optimizing} or by deterministic promotion rules for agents that fail to find a path~\cite{andreychuk2018two}.
A more principled alternative is priority-based search (PBS), which systematically explores the space of priority orderings and searches for a priority structure that is consistent with a conflict-free solution \cite{ma2019searching}.

Continuous-time priority-based search (CPBS) extends this idea to MAPF$_R$ by combining a PBS-style high-level search over partial priority relations with a SIPP-based low-level planner operating in continuous time~\cite{signorile2026continuous}.
In roadmap-based settings, CPBS can exploit precomputed continuous-time conflict annotations (CTC).
These annotations specify, offline, which pairs of roadmap actions or resources may lead to a collision for given relative execution times.
At runtime, they allow the trajectory of a higher-priority agent to be converted into time-varying reservations for lower-priority agents without performing geometric collision checks online. Related offline conflict-detection and annotation strategies are widely used in industrial AGV systems~\cite{olmi2011efficient} and have also been explored in roadmap-based MAPF planning~\cite{honig2018trajectory,kasaura2022prioritized}.

The resulting framework is attractive for industrial AGV coordination, but its performance is strongly affected by the cost of repeated replanning.
Each new priority refinement may require recomputing the path of the lower-priority agent and, through transitivity, the paths of all agents whose set of higher-priority predecessors has changed.
We refer to this computation of the multi-agent plan of a high-level node as its \textit{materialization}.
On large or highly connected industrial roadmaps, this operation can be expensive because each replanning step requires rebuilding priority-induced safe intervals and running SIPP over a roadmap with many alternative outgoing motions.
At the same time, the benefit of this effort depends on the interaction density of the instance.
In dense or highly constrained layouts, where bottlenecks generate many conflicts, revising priorities is essential because fixed-order prioritized planning is brittle.
In sparse layouts, instead, a simple priority ordering may already be sufficient, and eager materialization of high-level nodes can introduce overhead with limited conflict-resolution benefit.

These observations motivate a lazy variant of CPBS that avoids materializing nodes that are never expanded. Rather than replanning immediately for every generated child node, the proposed Lazy-CPBS delays materialization until node expansion, so that siblings that are never expanded cost no low-level search. The goal is to preserve the robustness of systematic priority revision in congested regimes while approaching the efficiency of prioritized planning when interactions are sparse.

\subsection{Contributions and Paper Outline}
This paper makes the following contributions. First, we analyze the priority-search space induced by partial priority orderings in priority-based continuous-time MAPF$_R$ solvers. The analysis clarifies how priority decisions propagate through the search, which agents are affected by a priority refinement, and why different priority branches can be reasoned about independently. This structural view also provides implementation-level consequences for CPBS-style solvers, including the identification of the SIPP boundary conditions required by priority-induced reservations and the possibility of detecting solutions earlier during node materialization.
Second, building on this structural analysis, we introduce \textit{Lazy-CPBS}, which delays node materialization until expansion, avoiding the low-level search for children that are never expanded. This reduces redundant SIPP calls and safe-interval updates while preserving the priority-search-space semantics of the eager formulation.
To ground these algorithmic contributions in industrial AGV coordination, we formalize AGV-MAPF$_R$-CTC as a configuration-space variant of MAPF$_R$ with precomputed continuous-time conflicts. We then evaluate
CPBS and Lazy-CPBS both on standard MAPF benchmarks converted to continuous-time roadmaps and on industrial AGV roadmaps imported from a proprietary digital-twin simulator.
Taken together, these results strengthen the connection between continuous-time MAPF and industrial AGV coordination, while identifying priority-search-space structure as a concrete lever for reducing replanning overhead and enabling further acceleration.

The remainder of this paper is organized as follows.
Section~\ref{sec:problem} formalizes the continuous-time multi-agent problem, introducing the MAPF$_R$-CTC model and its AGV-specific instantiation.
Section~\ref{sec:cpbs} revisits the CPBS framework, describes a SIPP-based low-level planner that ensures sound behavior under priority constraints, and formalizes structural properties of the CPBS high-level search that support efficient implementations.
Section~\ref{sec:lazycpbs} presents the novel Lazy-CPBS algorithm and discusses its relation to standard CPBS.
Section~\ref{sec:experiments} reports the experimental setup and results on standard MAPF$_R$ benchmark instances. Section~\ref{sec:industrial} then presents the industrial evaluation on warehouse roadmaps imported from a proprietary digital-twin simulator and derived from real AGV installations. Section~\ref{sec:ablation} presents an ablation study that isolates the contributions of lazy materialization and child ordering to the speed-up of Lazy-CPBS over CPBS.
Finally, Section~\ref{sec:conclu} concludes the paper and outlines directions for future work.

\section{Problem Statement}
\label{sec:problem}
\begin{figure*}
    \centering
    \includegraphics{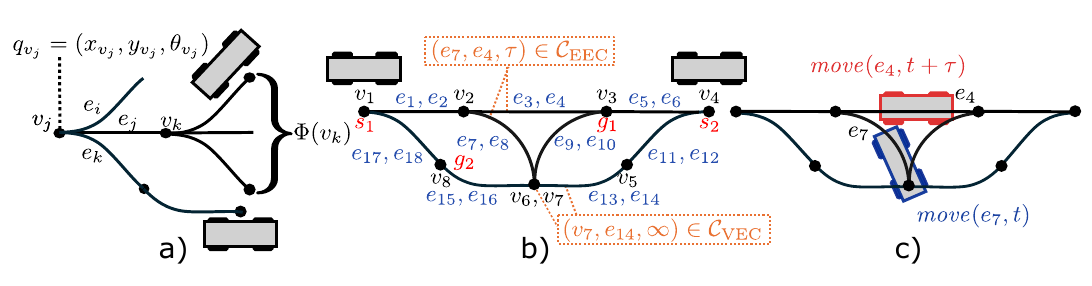}
    \caption{Illustration of the AGV-MAPF$_R$-CTC formulation.
(a) Excerpt of a directed configuration-space roadmap $\overrightarrow{\mathcal{R}}$, where each vertex $v\in V$ represents an admissible AGV stopping configuration and the outgoing edges $\Phi(v)\subseteq E$ encode the feasible motion primitives from that configuration.
(b) Example of an AGV-MAPF$_R$-CTC instance, with sample precomputed continuous-time conflict annotations. Geometrically coincident vertices may still represent different states if they correspond to different headings, as illustrated by $v_6$ and $v_7$. The annotation $(v_7,e_{14},\infty)\in \mathcal{C}_\mathrm{{VEC}}$ means that, as long as an agent occupies $v_7$, any other agent starting $\mathit{move}(e_{14},\cdot)$ would cause a collision.
(c) Interpretation of the relative-delay semantics of an edge-edge CTC annotation: if one agent executes $\mathit{move}(e_7,t)$, then $t+\tau$ is the earliest safe start time for another agent to execute $\mathit{move}(e_{4},\cdot)$; in the illustrated case, any start time in the interval $[t,t+\tau)$ would lead to a collision.}
\label{fig:problem}
\end{figure*}
The scope of this work spans both the MAPF$_R$ literature and the specific requirements of industrial AGV coordination.
We therefore
introduce the problem setting in three steps. We first recall the
classical continuous-time MAPF problem on generic roadmaps
(MAPF$_R$). We then introduce MAPF$_R$-CTC, where continuous-time
conflict annotations are precomputed and provided as part of the
problem input. Finally, we formalize AGV-MAPF$_R$-CTC, a
configuration-space variant tailored to AGV fleets operating on
fixed industrial infrastructures. This modeling structure is illustrated
in Fig.~\ref{fig:problem}.

This separation between the general MAPF$_R$ model and its industrial variant clarifies both the algorithmic context of our approach and the operational constraints under which our methods will be evaluated.

\subsection{MAPF$_R$}
\label{subsec:mapfr}
In the classical MAPF$_R$ problem, agents are identical---typically modeled as non-rotating discs of equal radius---and move at constant nominal speed along a predefined roadmap.
Formally, let $\mathcal{A} = \{1,...,n\}$ be the set of agents. An instance of the MAPF$_R$ problem is a tuple $(G, \mathcal{T})$, where $G = (V, E)$ is a weighted graph representing the roadmap and
$\mathcal{T} = \{(s_i, g_i)\}_{i=1}^n$ assigns each of the $n \in \mathbb{N}$ agents a start vertex $s_i \in V$ and a goal vertex $g_i \in V$, with no two agents sharing start and goal positions: $\forall i \in \mathcal{A} \; \nexists j \mid i \neq j \wedge \; (\; s_i = s_j \; \vee \; g_i = g_j)$.
Each vertex $v \in V$ corresponds to a point in the planar workspace $\mathcal{M} \subseteq \mathbb{R}^2$, and each edge $e = (u, v) \in E$ represents a collision-free geometric path in $\mathcal{M}$ that can be traversed at any departure time.
Edges are assumed to be straight segments with associated traversal time $w(e) \in \mathbb{R}_{\geq 0}$, and $G$ is symmetric.

Agents may either wait at a vertex or traverse an outgoing edge.
A traversal of edge $e$ initiated at time $t \in \mathbb{R}_{\ge 0}$ is denoted $\mathit{move}(e, t)$ and lasts $w(e)$, while a waiting action at vertex $v$ for duration $\Delta t \in \mathbb{R}_{\ge 0}$ is denoted $\mathit{wait}(v, t, \Delta t)$.

A solution of the problem is a collection of individual continuous-time trajectories
${\Pi} = \{\pi_1, \dots, \pi_n\}$, one per agent, where each $\pi_i$ is a temporally ordered sequence of admissible actions connecting $s_i \in V$ to $g_i \in V$.
The arrival time of agent $i$---the time at which it reaches its goal vertex for the last time---is denoted $T_i \in \mathbb{R}_{\geq 0}$, and the objective is to minimize the sum of arrival times $\sum_{i=1}^n T_i$, known as the sum of costs (SoC) of the solution.

Trajectories must be mutually collision-free: for any pair of agents $(i,j), i \neq j$ and any time $t \in \mathbb{R}_{\geq 0}$, their occupied regions in $\mathcal{M}$ must be disjoint.
For disc-shaped agents of radius $r \in \mathbb{R}_{> 0}$, this requires
$\| m_i(t) - m_j(t) \|_2 > 2r$,
where $m_i(t)$ and $m_j(t)$ denote the centers of agents $i$ and $j$ at time $t \in \mathbb{R}_{\geq 0}$.

\subsection{MAPF$_R$-CTC}
\label{subsec:mapfr-ctc}
The classical MAPF$_R$ formulation defined in Section \ref{subsec:mapfr} assumes that planners have no prior
knowledge of which pairs of actions may lead to collisions: any potential interaction
between agents must be detected online through geometric collision checks.
In this sense, MAPF$_R$ solvers are \textit{conflict-blind}: all continuous-time
interference between agents must be inferred dynamically during planning.

In many roadmap-based domains, however---including automated warehouses,
factory logistics, and other infrastructures where the roadmap is fixed---collision
relationships between roadmap elements can be computed offline, before
planning begins.
Providing this information as part of the problem input yields a richer model,
which we refer to as MAPF$_R$-CTC.

A defining feature of the MAPF$_R$-CTC formulation is that the roadmap is annotated offline with continuous-time conflicts (CTC).
For every vertex--edge and edge--edge pair, preprocessing computes a maximal,
time-translation-invariant unsafe interval of relative start times: if two actions begin
within this interval, they are guaranteed to collide, whereas any larger delay is
guaranteed to be safe~\cite{kasaura2022prioritized}.
All non-empty intervals are stored as tuples $(v,e,I_{\max}(v,e))$ and $(e_1,e_2,I_{\max}(e_1,e_2))$ in the vertex--edge and edge--edge conflict sets $\mathcal{C}_\mathrm{{VEC}}$ and $\mathcal{C}_\mathrm{{EEC}}$, which are treated as part of the problem input.

A tuple $(e_1,e_2,I_{\max}(e_1,e_2))\in\mathcal{C}_\mathrm{{EEC}}$ means that if one agent performs
$\mathit{move}(e_1,t)$ and another executes $\mathit{move}(e_2,t+\tau)$, then a collision is inevitable for every $\tau\in I_{\max}(e_1,e_2)$, whereas any delay
$\tau \ge \sup I_{\max}(e_1,e_2)$ is guaranteed to avoid the collision.
An analogous interpretation holds for $(v,e,I_{\max}(v,e))\in\mathcal{C}_\mathrm{{VEC}}$,
relating $\mathit{wait}(v,\cdot,\cdot)$ and $\mathit{move}(e,\cdot)$.
At runtime, these conflict sets allow planners to enforce collision avoidance through
simple interval checks rather than online geometric tests. See~\cite{kasaura2022prioritized} for the full construction of $\mathcal{C}_{\mathrm{VEC}}$ and $\mathcal{C}_{\mathrm{EEC}}$ and a representative roadmap-annotation example with circular agents.

\subsection{AGV-MAPF$_R$-CTC}
\label{subsec:agv-mapfr-ctc}
We now introduce a variant of MAPF$_R$ tailored to industrial AGV fleets.
Unlike holonomic disc agents, AGVs follow steering-constrained
trajectories, have direction-dependent motion primitives, and must
satisfy non-holonomic continuity between successive segments~\cite{siciliano2009robotics}.
Each vehicle is modeled as a rigid rectangle of dimensions $L_{\mathrm{AGV}}\times W_{\mathrm{AGV}}$, moving in a planar workspace $\mathcal{M}\subseteq \mathbb{R}^2$ with configuration space $\mathcal{C}=\mathbb{R}^2\times S^1$~\cite{lavalle2006planning}.
These constraints are captured by representing the navigation infrastructure as a directed, generally non-planar and  non-strict roadmap embedded in configuration space. Directionality accounts for
the fact that kinematic feasibility and nominal speed may depend on the
traversal direction. Moreover, geometrically coincident positions with
different headings correspond to distinct vertices, while multiple
kinematically distinct motion primitives may connect the same pair of
configurations.

More formally, an AGV-MAPF$_R$-CTC instance is defined as a tuple $(\overrightarrow{\mathcal{R}},\mathcal{T})$, where $\overrightarrow{\mathcal{R}}=(V,E)$ is a directed weighted roadmap embedded in configuration space and
$\mathcal{T}=\{(s_i,g_i)\}_{i\in\mathcal{A}}$ assigns each agent
$i\in\mathcal{A}=\{1,\ldots,n\}$ a start and goal vertex in $V$.
Each vertex $v\in V$ represents an admissible stopping configuration
\begin{equation}
    q(v) = (x_v,y_v,\theta_v) \in \mathcal{C},
\end{equation}
where $(x_v,y_v) \in \mathbb{R}^2$ is the footprint-center position and $\theta_v \in S^1$ is the vehicle heading (Fig. \ref{fig:problem}(a)).
The rectangular footprint associated with $q(v) \in \mathcal{C}$ is obtained by rotating an axis-aligned $L_{\mathrm{AGV}}\times W_{\mathrm{AGV}}$ rectangle by angle $\theta_v \in S^1$ in the workspace.
Each directed edge $e=(u,v)\in E$ encodes a physically realizable motion primitive from configuration $q(u) \in \mathcal{C}$ to $q(v) \in \mathcal{C}$ with travel time $w(e) \in \mathbb{R}_{\geq0}$; the primitive implicitly fixes the traversal mode (e.g., forward or reverse) and its admissible velocity profile.
By constructing $\overrightarrow{\mathcal{R}}$ directly in configuration space, the non-holonomic continuity constraints are embedded into the adjacency structure. The outgoing set
\begin{equation}
    \Phi(v) = \{\,e=(v,u)\in E\,\},
\end{equation}
therefore contains exactly the motion primitives that may follow a stop at configuration $q(v) \in \mathcal{C}$.
Thus, any roadmap path selected by the planner is kinematically consistent by construction.

As in MAPF$_R$, an agent may execute either a traversal action $\mathit{move}(e,t)$ or a waiting action $\mathit{wait}(v,t,\Delta t)$.
A feasible solution is a set of $n$ collision-free trajectories $\Pi=\{\pi_1,\dots,\pi_n\}$ connecting each pair $(s_i,g_i)$, and the objective is to find a SoC-minimal solution $\Pi^\star$.
Following the MAPF$_R$-CTC framework, the roadmap is enriched with offline continuous-time conflict annotations derived from the vehicle geometry and the curvature-constrained motion primitives.
For each relevant pair of roadmap elements, these annotations specify the time-translation-invariant unsafe intervals of relative start times that would unavoidably lead to a collision.
We denote by $\mathcal{C}_\mathrm{VEC}$ and $\mathcal{C}_\mathrm{EEC}$ the vertex--edge and edge--edge conflict sets, respectively. Examples of such annotations are shown in Fig.~\ref{fig:problem}(b)-(c).
This model, denoted \textit{AGV-MAPF$_R$-CTC}, preserves the continuous-time
geometric structure of MAPF$_R$ while capturing the kinematic specificity and infrastructural regularity characteristic of industrial AGV systems.
\begin{remark}
The sets $\mathcal{C}_\mathrm{VEC}$ and $\mathcal{C}_\mathrm{EEC}$ are part of the
AGV-MAPF$_R$-CTC input. If not available, they can be computed offline by checking pairwise roadmap
actions; for rectangular AGVs, collision queries are constant-time via
the separating axis theorem, and unsafe delay intervals can be sampled
over relative start times~\cite{ericson2004real,walker2019collision}.
\end{remark}
\begin{remark}
AGV-MAPF$_R$-CTC generalizes MAPF$_R$-CTC: the roadmap becomes directed and possibly non-strict, with pose vertices. Since geometry and kinematics are compiled into the roadmap and
into $\mathcal{C}_\mathrm{VEC}$ and $\mathcal{C}_\mathrm{EEC}$, both problems share the same input, a weighted
directed graph with conflict sets, so any MAPF$_R$-CTC solver that does not rely on roadmap
symmetry applies without modifying its search logic.
\end{remark}

\section{Continuous-Time Priority-Based Search}
\label{sec:cpbs}
Continuous-time priority-based search (CPBS) \cite{signorile2026continuous} is an incomplete, suboptimal MAPF$_R$-CTC (and AGV-MAPF$_R$-CTC) solver that extends the classical priority-based search (PBS) framework \cite{ma2019searching} to continuous-time settings.
It follows a two-level architecture: the high level explores partial priority orderings among agents, while the low level performs single-agent replanning using a SIPP-based planner that treats higher-priority agents as dynamic
obstacles through time-varying reservations on roadmap resources.

A key difference between CPBS and CCBS lies in how conflicts are handled.
When CPBS imposes a precedence constraint $i \prec j, \; i,j \in \mathcal{A}$, the meaning is global:
agent $j  \in \mathcal{A}$ must find a trajectory that avoids every possible continuous-time
interaction with the current trajectory of agent $i  \in \mathcal{A}$.
To enforce this, CPBS must treat the full timed path of $i$ as a dynamic obstacle
for $j$, inserting time-stamped reservations not only on the vertices and edges
directly occupied by $i$, but also on all roadmap elements that are geometrically
blocked by those resources.
These reservations can be constructed online by checking
candidate roadmap actions against the complete timed trajectories
of higher-priority agents. Without annotations, repeated low-level
searches may require recomputing forbidden intervals for many
resources, which can be expensive on large, highly connected
roadmaps. CTC annotations avoid repeating the geometric computation
by translating precomputed conflict relations into time-stamped
reservations.

CCBS instead introduces agent-specific action-time constraints
from selected conflicts between current plans. Its low-level
planner enforces these constraints without treating other agents'
complete trajectories as fixed obstacles. CCBS can use online
geometric conflict detection or precomputed annotations. Priority-based solvers likewise remain applicable to MAPF$_R$
without CTC annotations. We adopt MAPF$_R$-CTC to reuse precomputed
collision geometry across replanning calls, priority branches,
and instances on a fixed roadmap, while constructing
trajectory-dependent reservations online.

In the remainder of this section, we denote by $\overrightarrow{\mathcal{R}}$ a generic CTC-annotated roadmap. Depending on the domain, $\overrightarrow{\mathcal{R}}$ may correspond either to the standard MAPF$_R$-CTC roadmap $G$ or to the AGV configuration-space roadmap $\overrightarrow{\mathcal{R}}$. The CPBS and Lazy-CPBS search logic is unchanged in both cases.

\subsection{CPBS High-Level Search}
\label{subsec:cpbs-high-level}
\begin{figure}
    \centering
    \includegraphics[width=0.48\textwidth]{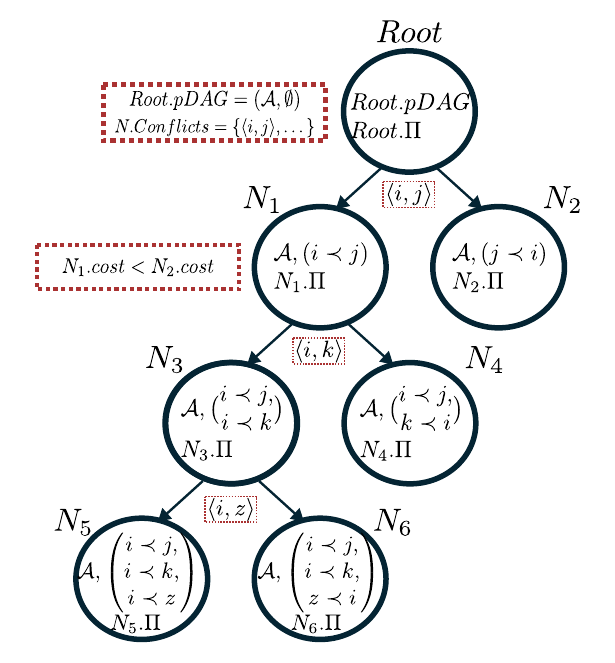}
    \caption{High-level CPBS priority tree.}
    \label{fig:cpbs_high_level}
\end{figure}
Unlike CCBS-style methods, which maintain and expand a constraint tree, CPBS relies exclusively on priority relations to solve conflicts, incrementally refining a \textit{priority directed acyclic graph} ($\mathrm{pDAG}$) that encodes the local precedence constraints required to eliminate inter-agent collisions.

As illustrated in Fig.~\ref{fig:cpbs_high_level}, CPBS performs a depth-first search \cite{peter2021modern} over the \textit{priority tree} (PT), whose nodes encode
partial priority structures and their associated multi-agent plans. A PT node $N$ is defined by a tuple $N = (N.\mathrm{pDAG},\; N.\Pi,\; N.\mathrm{cost}, \; N.Conflicts)$,
where $N.\mathrm{pDAG}$ is a priority directed acyclic graph,
$N.\Pi = \{\pi_1,\ldots,\pi_n\}$ is a set of continuous-time paths that comply with $N.\mathrm{pDAG}$, $N.\mathrm{cost}$ is the SoC of $N.\Pi$, and $N.Conflicts$ is the set of unresolved pairwise conflicts in the current solution $N.\Pi$.

Formally, a priority graph is represented as
\mbox{$\mathrm{pDAG}=(\mathcal{A},{\prec})$}, where
$\mathcal{A}=\{1,\ldots,n\}$ is the set of agents and
$\prec\subseteq\mathcal{A}\times\mathcal{A}$ specifies precedence constraints.
An ordered pair $(i,j)\in \; \prec$ indicates that agent $i$ has higher priority than agent $j$,
denoted $i\prec j$.
By construction, $\mathrm{pDAG}$ must be acyclic.
Acyclicity ensures that $\prec$ defines a valid partial order,
and therefore admits at least one topological ordering \cite{kahn1962topological} of the agents.
If a cycle existed---e.g., $i\prec j\prec k\prec i$---the induced ordering would be contradictory:
each agent in the cycle would be required to defer to another, making the set of constraints infeasible.

The set of single-agent paths $N.\Pi$ is said to \textit{respect} $N.\mathrm{pDAG}$ if every path $\pi_i$ was produced by the low-level SIPP planner while treating each higher-priority agent $j$ with $j\prec i$ as a dynamic obstacle.
This mechanism ensures that $N.\Pi$ is consistent with $N.\mathrm{pDAG}$ by construction and that collisions
between higher- and lower-priority agents are systematically avoided.

\begin{algorithm}[t!]
\caption{High-level CPBS search}
\label{alg:cpbs-high-level}
\begin{algorithmic}[1]
\Require $\overrightarrow{\mathcal{R}}, \mathcal{T}, \mathcal{C}_{\mathrm{VEC}}, \mathcal{C}_{\mathrm{EEC}}$
\State $Root.\mathrm{pDAG} \gets (\mathcal{A}, \emptyset)$, $\quad Root.\Pi \gets \emptyset$
\ForAll{$ i \in \mathcal{A}$}
 \State $success \gets \textsc{UpdatePlan}(Root, i)$
 \If{$\lnot success$}
     \Return ``no solution''
\EndIf
\EndFor

\State $Root.Conflicts \gets \textsc{FindConflicts}(Root.\Pi, \mathcal{C}_{\mathrm{VEC}}, \mathcal{C}_{\mathrm{EEC}})$
\If{$Root.Conflicts = \emptyset$}
     \Return $Root.\Pi$ \EndIf
\State STACK $\gets \{Root\}$
\While{STACK $\neq \emptyset$ }
\State{$N \gets $ top node in STACK}
\State STACK $\gets$ STACK $\setminus \{N\}$
\State $\langle i,j \rangle \gets \textsc{SelectConflict}(N.Conflicts)$
\For{$x \in \{\,i,j\,\}$}
    \State $y \gets$ other agent in $\{i,j\} \setminus \{x\}$
    \State $N' \gets$ copy of $N$
    \State $N'.\mathrm{pDAG} \gets N.\mathrm{pDAG} \cup \{\,x \prec y\,\}$
    \State $success \gets \textsc{UpdatePlan}(N', y)$
    \If{$success$}
\State $N'.Conflicts \gets$
    \Statex \hspace{\algorithmicindent}\hspace{\algorithmicindent}\qquad \quad
    $\textsc{FindConflicts}(N'.\Pi,
    \mathcal{C}_{\mathrm{VEC}}, \mathcal{C}_{\mathrm{EEC}})$
\If{$N'.Conflicts = \emptyset$}
     \Return $N'.\Pi$ \EndIf
        \State insert $N'$ into STACK so that the lower-cost sibling is on top
    \EndIf
\EndFor
\EndWhile
\State \Return ``no solution''
\Statex

\Function{UpdatePlan}{$N',\, y$}
    \State $\text{order} \gets \textsc{TopoFrom}(N'.\mathrm{pDAG},\, y)$
    \ForAll{$a \in \text{order}$}
        \State $\mathcal{I} \gets \textsc{ResetFullSafe}(\overrightarrow{\mathcal{R}})$
        \State $H \gets \textsc{Predecessors}(N'.\mathrm{pDAG},\, a)$
        \State $\mathcal{I} \gets$ \textsc{InsertHigherReservations}$(H, N'.\Pi)$
        \State $\pi_a' \gets \text{SIPP}(\overrightarrow{\mathcal{R}}, s_a,\, g_a,\, \mathcal{I}, \mathcal{H})$
        \If{$\pi_a' = $ failure}
             \Return \textbf{false}
        \EndIf
        \State $N'.\Pi[a] \gets \pi_a'$
    \EndFor
    \State $N'.\mathrm{cost} \gets \textsc{SoC}(N'.\Pi)$
    \State \Return \textbf{true}
\EndFunction

\end{algorithmic}
\end{algorithm}

Algorithm~\ref{alg:cpbs-high-level} illustrates the high-level  search procedure of CPBS.
The algorithm initializes the root PT node with an edge-empty $Root.\mathrm{pDAG} = (\mathcal{A},\emptyset)$ and an initially empty multi-agent plan $Root.\Pi$ (line~1). It then constructs the root plan by calling \textsc{UpdatePlan} once per agent $i \in \mathcal{A}$, ensuring that the initial multi-agent plan is computed in a temporally unconstrained environment, with constraints introduced only through explicit precedence relations in the $\mathrm{pDAG}$.
Since no precedence relations are enforced yet, these initial paths treat all other agents as absent; if any SIPP call fails, the instance is deemed infeasible and the algorithm terminates immediately (lines~2--6).

Once the initial multi-agent plan \(Root.\Pi\) has been computed, the algorithm tests it for conflicts (line~7).
If no collisions are present, the root node already represents a valid solution and the search halts.
Otherwise, CPBS initiates a depth-first search over the priority tree: a stack is created and the root node is pushed onto it (line~10), after which nodes are extracted and expanded in LIFO order (lines~11--13).
For each extracted PT node \(N\), the algorithm selects one conflict according to a deterministic conflict selection heuristic
(in our implementation, earliest-conflict-time policy, line~14).
Let the selected conflict involve agents \(i\) and \(j\).
To solve it, CPBS branches on the two possible local precedence assignments $i \prec j$ and $j \prec i$, thereby generating two child nodes (lines~15--18).
Each child inherits the parent's $\mathrm{pDAG}$ and adds exactly one new edge: the child priority graph is obtained by extending the parent $\mathrm{pDAG}$ with the corresponding directed edge choice $x \prec y$.

Since enforcing \(x \prec y\) makes agent \(y \in \mathcal{A}\) responsible for avoiding \(x \in \mathcal{A}\), CPBS invokes \textsc{UpdatePlan} on \(y\), which recomputes its trajectory and, if necessary, those of all agents that depend on it through transitivity. This replanning step materializes the child node:
the full multi-agent plan associated with the updated $\mathrm{pDAG}$ is generated
immediately, and its conflicts can therefore be evaluated on the spot.
Unlike~\cite{signorile2026continuous, ma2019searching}, where the goal test is performed at expansion, this anticipates it to the moment a child is materialized.
As a consequence, if a child node already yields a conflict-free plan, the solver can recognize it as a valid solution without exploring the sibling branch. We refer to this as the \textit{early goal test}.

The \textsc{UpdatePlan} procedure (lines~29--42) operates as follows.
Given a child node \(N'\) and the agent \(y\) whose priorities have worsened, a topological ordering of the subgraph induced by \(y\) and all its descendants in the $\mathrm{pDAG}$ is computed with a standard linear-time topological sorting algorithm \cite{kahn1962topological} (line~30).
This ordering ensures that agents are replanned in a sequence consistent with the current precedence structure: a descendant agent is never replanned before all its higher-priority predecessors have been fixed.
For each agent \(a \in \text{order}\) appearing in this ordering, the algorithm resets all safe intervals (line~32), retrieves the set \(H \subset \mathcal{A}\) of agents that dominate \(a\) in the $\mathrm{pDAG}$ (line~33), and inserts their time-stamped reservations into the safe-interval tables of the roadmap (line~34).
This process effectively removes from the temporal search space all time intervals during which \(a\) would collide with a higher-priority agent, thereby constraining its admissible continuous-time motions.
A new SIPP query is then executed for agent \(a\) (line~35), producing a new trajectory that avoids all such reservations.
If any replanning step fails, \textsc{UpdatePlan} returns \textbf{false} and the corresponding branch of the priority tree is discarded.
Otherwise, after all affected agents have been replanned, the cost of the resulting plan is updated (line~40), and the algorithm resumes the high-level search.

If the child node \(N'\) remains feasible and is not a solution, it is pushed onto the stack for further exploration (lines~20--24); feasible siblings are pushed so that the one with the lower cost is expanded first.
Due to the depth-first nature of the search, CPBS prioritizes deepening along one branch of the priority space before backtracking, allowing it to rapidly discover feasible orderings when they exist.
The search continues until either a conflict-free node is found or all branches of the priority tree are exhausted.

\subsection{SIPP}
\label{subsec:sipp}
\begin{figure}
    \centering
    \includegraphics[width=0.48\textwidth]{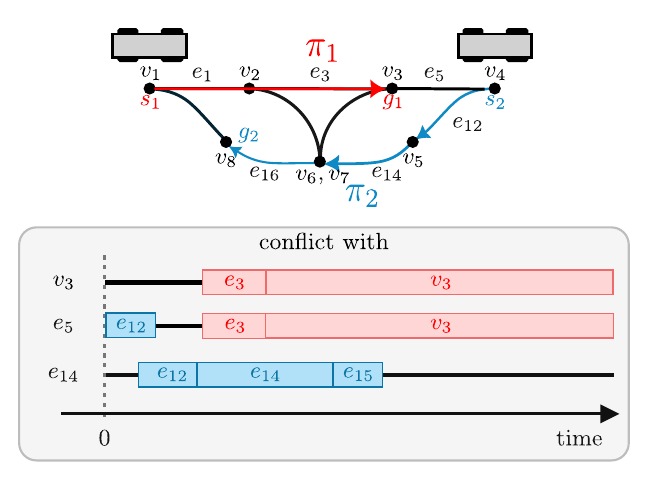}
    \caption{Example of priority-induced safe intervals used by SIPP.}
    \label{fig:sipp_safe_intervals}
\end{figure}

CPBS relies on a modified version of SIPP \cite{phillips2011sipp} as its single-agent planner.
SIPP extends A$^*$ search \cite{4082128} by associating each vertex $v \in V$ and edge $e \in E$ with a set of safe intervals ($\mathcal{I}_v \in \mathcal{I}$ and $\mathcal{I}_e \in \mathcal{I}$, respectively), i.e., maximal time windows during
which the resource can be used without colliding with any higher-priority
agent.  A state is a pair $(v,I)$, where $v\in V$ is a vertex and
$I=[I_s,I_e]\in \mathcal{I}_v$ is one such interval; its cost $g(v,I)$ is the
earliest feasible arrival time at $v$ within $I$, and its priority key is
$f(v,I)=g(v,I)+h(v,g_a)$, where $h(v,g_a) \in \mathcal{H}$ is a precomputed admissible heuristic. In our implementations we use the exact shortest-path travel time on the static roadmap with unbounded resource availability, precomputed once by running backward Dijkstra search from each goal vertex over the entire roadmap, which is admissible by construction. The OPEN list is ordered by increasing $f$, and
dominated states are pruned: if a state $(v,I)$ is reached with an arrival
time no better than a previously observed one, it is discarded \cite{gonzalez2012using}. Fig.~\ref{fig:sipp_safe_intervals} illustrates how priority-induced reservations are translated into safe intervals for the low-level planner.

To ensure soundness under priority-based reasoning, CPBS requires two boundary conditions on the SIPP low-level planner that are not necessarily enforced in CCBS-style implementations.
First, the planner must start from the \textit{earliest} safe interval of $s_a$,
which must intersect $t=0$; allowing SIPP to start from a later interval
(e.g., $[45,+\infty)$ when $[0,33]$ exists) would introduce an initial wait that
may already violate reservations imposed by higher-priority agents.
Second, a plan is accepted as a goal only if it terminates in the \textit{unbounded}
safe interval of $g_a$: reaching $g_a$ inside a finite interval, such as
$[33,190]$, would imply that the agent remains at its goal beyond the end of that interval,
thereby postponing an unavoidable future conflict that CPBS cannot repair
within the same priority assignment.

These conditions are specific to the semantics of priority-induced reservations. In CCBS-style solvers, SIPP is typically used inside an incremental repair loop: if a later conflict is detected, the high level can introduce an additional constraint and replan. In CPBS, instead, a precedence relation is interpreted as eliminating all interactions between the replanned agent and its higher-priority predecessors in a single low-level call. Therefore, starting from a late safe interval or accepting a finite goal interval may produce a plan that is valid for a CCBS repair step but invalid under CPBS semantics. Enforcing the two boundary conditions above ensures that SIPP preserves the intended meaning of the \(\mathrm{pDAG}\).

Given a state $(v,I)$, successors are generated by considering each outgoing
edge $e=(v,u)$ and computing the earliest departure time compatible with
the current interval $I$, the safe intervals of $e$, and those of $u$.  A
successor $(u,J)$ is added only if the traversal lies entirely within the
corresponding edge and vertex intervals; for each interval $J$, only the
earliest feasible arrival is retained, since later departures cannot improve the
solution under a monotone time metric \cite{gonzalez2012using}.
This SIPP variant is thus both sound for CPBS and efficient, matching the
structure of standard SIPP while enforcing the additional temporal
constraints required by priority-based continuous-time planning.

\subsection{Structural Properties and Implementation Implications}
\label{subsec:cpbs_properties}
This subsection discusses structural properties of the CPBS high-level search and outlines how they can guide future multi-threaded implementations aimed at further accelerating CPBS-style solvers.
Note that a CPBS node $N$ is represented by its priority graph $N.\mathrm{\mathrm{pDAG}}$: for each agent $a$, the low-level constraints depend only on the set of higher-priority agents
\begin{equation}
H_N(a)=\{b\in \mathcal{A} \mid b\prec_N^* a\},
\end{equation}
where $\prec_N^*$ denotes the transitive closure of $N.\mathrm{pDAG}$. Thus, a $\mathrm{pDAG}$, or equivalently its transitive closure, provides a canonical representation of the high-level priority state.

\begin{property}[\textit{Conflicts involve incomparable agents}]
\label{prop:incomparable-conflicts}
Let $N$ be a materialized CPBS node whose plan $N.\Pi$ respects $N.{\mathrm{pDAG}}$. If two agents are comparable in $N.{\mathrm{pDAG}}$, then the lower-priority one has been planned while treating the higher-priority one as a dynamic obstacle. Hence, under sound SIPP planning and correct CTC reservations, selected conflicts can only involve agents that are incomparable in the current $N.\mathrm{pDAG}$.
\end{property}

\begin{property}[\textit{Acyclicity preservation}]
\label{prop:acyclicity-preservation}
Let $N$ be a valid CPBS node with an acyclic $\mathrm{pDAG}$, and let $\langle i,j\rangle$ be the selected conflict. By Property~\ref{prop:incomparable-conflicts}, neither $i\prec_N^* j$ nor $j\prec_N^* i$ holds. Therefore, adding either $i\prec j$ or $j\prec i$ cannot create a directed cycle, and both generated children preserve the $\mathrm{pDAG}$ invariant.
\end{property}

\begin{property}[\textit{Affected-subgraph replanning}]
\label{prop:affected-subgraph-replanning}
Let $N'$ be obtained from $N$ by adding $x\prec y$. Only $y$ and its descendants in the updated $\mathrm{pDAG}$ may have changed low-level constraints. Therefore, the agents to be replanned are exactly
\begin{equation}
\label{eq:replan-list}
D_{N'}(y)=\{y\}\cup\{a\in \mathcal{A} \mid y\prec_{N'}^* a\},
\end{equation}
processed in a topological order of the updated $\mathrm{pDAG}$.
\end{property}

\begin{property}[\textit{Disjoint priority subtrees}]
\label{prop:disjoint-priority-subtrees}
Let $N$ be a CPBS node and let $\langle i,j\rangle$ be the conflict selected for branching. The two children add opposite precedence relations, $i\prec j$ and $j\prec i$. Since CPBS only extends $\mathrm{pDAG}$s monotonically, every descendant of the first child contains $i\prec^* j$, whereas every descendant of the second contains $j\prec^* i$. Hence, under the acyclicity invariant, the two subtrees cannot contain the same valid canonical high-level state.
\end{property}

\begin{property}[\textit{Replan list from the parent}]
\label{prop:replan-from-parent}
Let $N'$ be obtained from $N$ by adding $x\prec y$. Then $D_{N'}(y)=D_N(y)$, and any topological order of $N.\mathrm{pDAG}$ restricted to $D_N(y)$ is a valid replanning order in $N'$. Indeed, the new edge enters $y$, so it creates no new path leaving $y$, and $x\notin D_N(y)$ by Property~\ref{prop:incomparable-conflicts}. Hence the child's replan list and its size $|D_N(y)|$ can be computed on the parent without updating or re-sorting the child $\mathrm{pDAG}$s.
\end{property}

\begin{figure}
    \centering
    \includegraphics[width=0.48\textwidth]{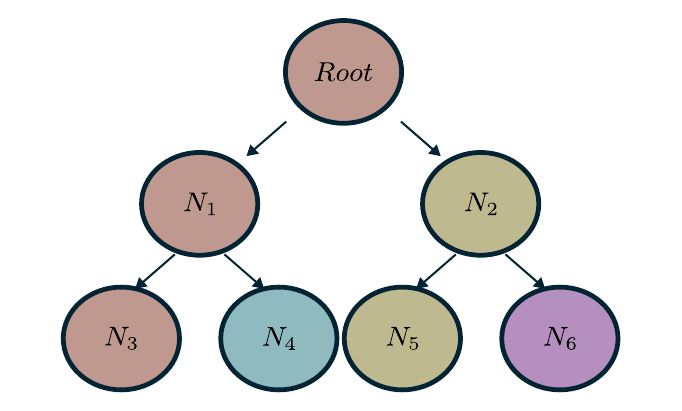}
    \caption{Parallel CPBS high-level exploration from independent frontier nodes.}
    \label{fig:parallel_cpbs_frontier}
\end{figure}

These properties also suggest a natural parallel implementation strategy.
Fig.~\ref{fig:parallel_cpbs_frontier} sketches the corresponding frontier-parallel view: whenever a worker is available, an unprocessed frontier node is assigned to it, and the worker performs a local depth-first descent within the subtree rooted at that node; the worker terminates when it backtracks to its assigned root and cannot continue without entering subtrees already assigned to other workers.
Different workers could therefore expand and materialize distinct frontier nodes independently, while sharing only read-only data such as the roadmap $\overrightarrow{{\mathcal{R}}}$, the CTC annotations $\mathcal{C}_{\mathrm{VEC}}$ and $\mathcal{C}_{\mathrm{EEC}}$, and the heuristic table used by SIPP.
Mutable low-level structures, such as safe-interval tables and conflict calendars, should instead be local to each worker.
The only global mutable structures would be a concurrent frontier and an atomic generated-state table indexed by canonical $\mathrm{pDAG}$ keys.
Such an implementation is left for future work.

\section{Lazy-CPBS}
\label{sec:lazycpbs}
\begin{figure}
    \centering
    \includegraphics[width=0.48\textwidth]{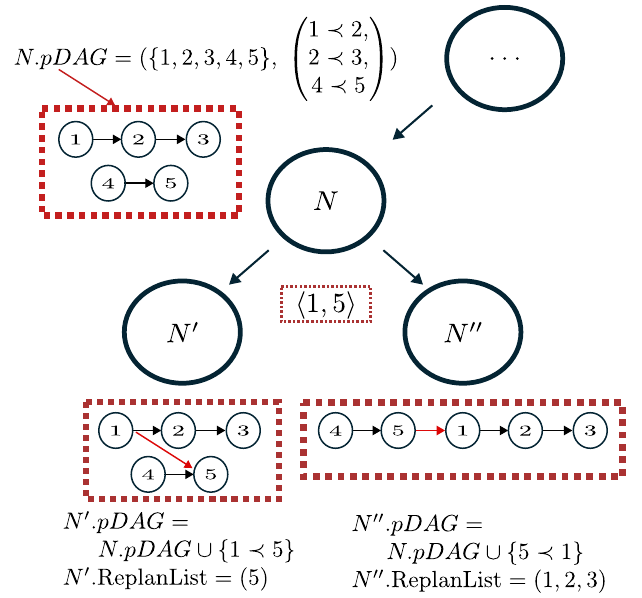}
    \caption{Replan-list construction in Lazy-CPBS.
A conflict between agents $1$ and $5$ generates two children with opposite precedence relations.
Adding $1\prec 5$ requires replanning only agent $5$, whereas adding $5\prec 1$ affects agent $1$ and its descendants in the $\mathrm{pDAG}$, producing $N''.\mathrm{ReplanList}=(1,2,3)$. Since Lazy-CPBS prioritizes the child with the smaller $\mathrm{ReplanList}$, the branch induced by $1\prec 5$ is expanded first.}
    \label{fig:lazy_replan_list}
\end{figure}
In this section we introduce \textit{Lazy-CPBS}, a novel algorithm that explores the PT while postponing replanning until nodes are expanded.

The CPBS solver described in Section \ref{subsec:cpbs-high-level} already exploits a depth-first exploration of the PT. At each
high-level step, CPBS incrementally refines the current $\mathrm{pDAG}$
by adding precedence edges between conflicting agents, thereby
constructing a search tree over feasible priority relations.
However, whenever a conflict between two agents $\langle i,j \rangle$
is selected, CPBS immediately \textit{materializes} both children.
For each child, it augments the $\mathrm{pDAG}$ with either $i \prec j$ or
$j \prec i$ and then triggers a full low-level replanning cycle via \textsc{UpdatePlan}. This procedure first resets all safe intervals on the roadmap to their default ``fully safe'' state and then scans the agents in a topological order of the updated $\mathrm{pDAG}$, repeatedly shrinking safe intervals and calling SIPP to replan the lower-priority agent and any agent that becomes reachable from it in $\mathrm{pDAG}$~\eqref{eq:replan-list}.
As a result, a single high-level branching may induce multiple SIPP calls and global passes over the safe-interval structure.

This eager strategy becomes particularly costly on sparse instances. Instance density plays a crucial role in the behavior of priority-based methods: as the interaction density grows, CPBS remains robust by systematically revising priority decisions through precedence constraints. However, on large layouts the same number of agents induces much sparser interactions. In such regimes, even a fixed priority ordering is frequently sufficient to solve the instance, while CPBS may still pay the full cost of repeatedly descending the priority tree. As a result, CPBS can spend a significant portion of its runtime materializing children that are never expanded, even on instances that a simple fixed ordering would already solve.

These observations motivate the introduction of \textit{Lazy-CPBS},
a lazy variant of CPBS that preserves the same search space over
priority orderings but avoids eagerly recomputing safe intervals
and single-agent plans at every high-level decision. Algorithm~\ref{alg:lazycpbs-high-level} shows the high-level
search of Lazy-CPBS. The PT has the same meaning as
in CPBS, but each node now stores different information: besides the current $\mathrm{pDAG}$, a node
$N$ keeps a multi-agent plan $N.\Pi$ and a replan list $N.\mathrm{ReplanList}$ containing exactly the
agents to be replanned (Property~\ref{prop:affected-subgraph-replanning}) when $N$ is expanded.

The search starts by creating the root node $Root$. Its $\mathrm{pDAG}$ contains all agents but no precedence edges, representing a fully unconstrained priority ordering. No low-level paths are computed at this point: the root carries an unmaterialized plan, and the node is simply pushed onto the \textit{STACK} (lines~1--3). When a node $N$ is popped from \textit{STACK} for expansion (lines~5--6), it is
first materialized. The current multi-agent plan $N.\Pi$ (empty at the root) is updated by replanning exactly the agents in $N.\mathrm{ReplanList}$ (line~7).
If materialization fails, the node is discarded without branching and the high-level search backtracks; otherwise, $N.\Pi$ is updated and conflicts are then checked on the resulting plan (lines~7-9). If the current multi-agent plan $N.\Pi$ is collision-free, it is returned
as a valid solution. Otherwise, a conflict $\langle i,j \rangle$
is selected for expansion (lines~10--12). As in CPBS, our implementation adopts
the earliest-conflict-time policy.
The selected conflict $\langle i,j \rangle$ is then resolved by branching on the two possible local precedence relations. Lazy-CPBS creates two children, $N'$ and $N''$, whose $\mathrm{pDAG}$s extend $N.\mathrm{pDAG}$ with the corresponding precedence edge
(lines~13--15). For each child, the new constraint is injected into the $\mathrm{pDAG}$, the set of agents to be replanned is extracted in topological order from the parent (Property~\ref{prop:replan-from-parent}) and stored in $N'.\mathrm{ReplanList}$ and $N''.\mathrm{ReplanList}$ (lines~16--17). At this point, both children inherit a copy of $N.\Pi$ as a provisional multi-agent plan, which will only be updated when the node is expanded and materialized.
Finally, the two child nodes are pushed onto \textit{STACK} in an order that favors the node with the smaller replan list as illustrated in Fig.~\ref{fig:lazy_replan_list} (line 19). In this way, Lazy-CPBS not only performs low-level materialization for one node at a time, but also biases the depth-first search towards nodes that require less replanning effort, in terms of both safe-interval updates and SIPP calls, thereby accelerating the descent along the PT.

The \textsc{MaterializeNode} procedure plays the same role as \textsc{UpdatePlan} in CPBS and replans the same set of agents (Property~\ref{prop:affected-subgraph-replanning}), the only difference is that this set, together with its topological order, has already been computed on the parent (Property~\ref{prop:replan-from-parent}) and stored in $N.\mathrm{ReplanList}$. For each agent in this list, the procedure resets the safe
intervals, inserts the reservations of its higher-priority agents, and calls SIPP to compute a new collision-free path (lines 24--27). If all SIPP calls succeed, the multi-agent plan $N.\Pi$ is updated accordingly (line 30) and the node $N$ is considered fully materialized; otherwise, \textsc{MaterializeNode} returns \textbf{false} and the node is discarded (lines 28 and 7).

\begin{remark}
Lazy-CPBS inherits the structural Properties~\ref{prop:incomparable-conflicts}--\ref{prop:replan-from-parent}: it changes only when replanning is performed and which child is explored first, not the PT itself, and is therefore amenable to the same parallel high-level exploration strategy discussed in Section~\ref{subsec:cpbs_properties}.
\end{remark}

\begin{algorithm}[t!]
\caption{Lazy-CPBS high-level search}
\label{alg:lazycpbs-high-level}
\begin{algorithmic}[1]
\Require $\overrightarrow{\mathcal{R}}, \mathcal{T}, \mathcal{C}_{\mathrm{VEC}}, \mathcal{C}_{\mathrm{EEC}}$
    \State $Root.\mathrm{pDAG} \gets (\mathcal{A}, \emptyset)$, $Root.\Pi \gets \emptyset$
    \State $Root.\mathrm{ReplanList} \gets (1,\dots,n)$

    \State STACK $\gets \{Root\}$

    \While{STACK $\neq \emptyset$}
        \State $N \gets$ top node in STACK
         \State STACK $\gets$ STACK $\setminus \{N\}$
                 \If{$\neg \textsc{MaterializeNode}(N)$}
             \textbf{continue}
        \EndIf
         \State $\textit{conflicts} \gets  \textsc{FindConflicts}(N.\Pi, \mathcal{C}_{\mathrm{VEC}}, \mathcal{C}_{\mathrm{EEC}})$
         \If{$\textit{conflicts} = \emptyset$}  \Return $N.\Pi$
         \EndIf
        \State $\langle i,j \rangle \gets \textsc{SelectConflict}(\textit{conflicts})$
        \State $N' \gets$ new node, $N'' \gets$ new node
        \State $N'.\mathrm{pDAG} \gets N.\mathrm{pDAG} \cup \{\,i \prec j\,\}$
        \State $N''.\mathrm{pDAG} \gets N.\mathrm{pDAG} \cup \{\,j \prec i\,\}$
        \State $N'.\mathrm{ReplanList} \gets \textsc{TopoFrom}(N.\mathrm{pDAG},\, j)$
        \State $N''.\mathrm{ReplanList} \gets \textsc{TopoFrom}(N.\mathrm{pDAG},\, i)$
         \State $N'.\Pi \gets N.\Pi, \quad N''.\Pi \gets N.\Pi$
        \State push $N'$ and $N''$ onto STACK in non-increasing order of $|N'.\mathrm{ReplanList}|$ and $|N''.\mathrm{ReplanList}|$

    \EndWhile
    \State \Return ``no solution''
    \Statex
    \Function{MaterializeNode}{$N$}
    \ForAll{$a \in N.{\mathrm{ReplanList}}$}
        \State $\mathcal{I} \gets \textsc{ResetFullSafe}(\overrightarrow{\mathcal{R}})$
        \State $H \gets \textsc{Predecessors}(N.\mathrm{pDAG},\, a)$
        \State $\mathcal{I} \gets$ \textsc{InsertHigherReservations}$(H, N.\Pi)$
        \State $\pi_a' \gets \text{SIPP}(\overrightarrow{\mathcal{R}}, s_a,\, g_a,\, \mathcal{I}, \mathcal{H})$
        \If{$\pi_a' = $ failure}
             \Return \textbf{false}
        \EndIf
        \State $N.\Pi[a] \gets \pi_a'$
    \EndFor
    \State \Return \textbf{true}
\EndFunction
\end{algorithmic}
\end{algorithm}

\section{Experimental MAPF$_R$ Evaluation}
\label{sec:experiments}

To assess the performance of Lazy-CPBS and CPBS, we conducted two extensive experimental campaigns with a twofold objective: (i) providing reproducible results on standard benchmarks from the MAPF literature, and (ii) validating transferability and robustness in a realistic industrial setting.
Specifically, the evaluation comprises: (a) a benchmark campaign in the MAPF$_R$-CTC domain on classical MAPF instances, and (b) an industrial campaign on AGV-MAPF$_R$-CTC instances built from roadmaps and conflict annotations imported from a proprietary industrial simulator.
In the remainder of this section we focus on the standard benchmark campaign;
results on industrially derived AGV roadmaps are reported and discussed separately in Section~\ref{sec:industrial}.

\subsection{Simulation Setup}
\label{subsec:benchmark-setup}
\begin{figure}[t]
\centering

\subfloat[den]{
\includegraphics[width=0.32\linewidth]{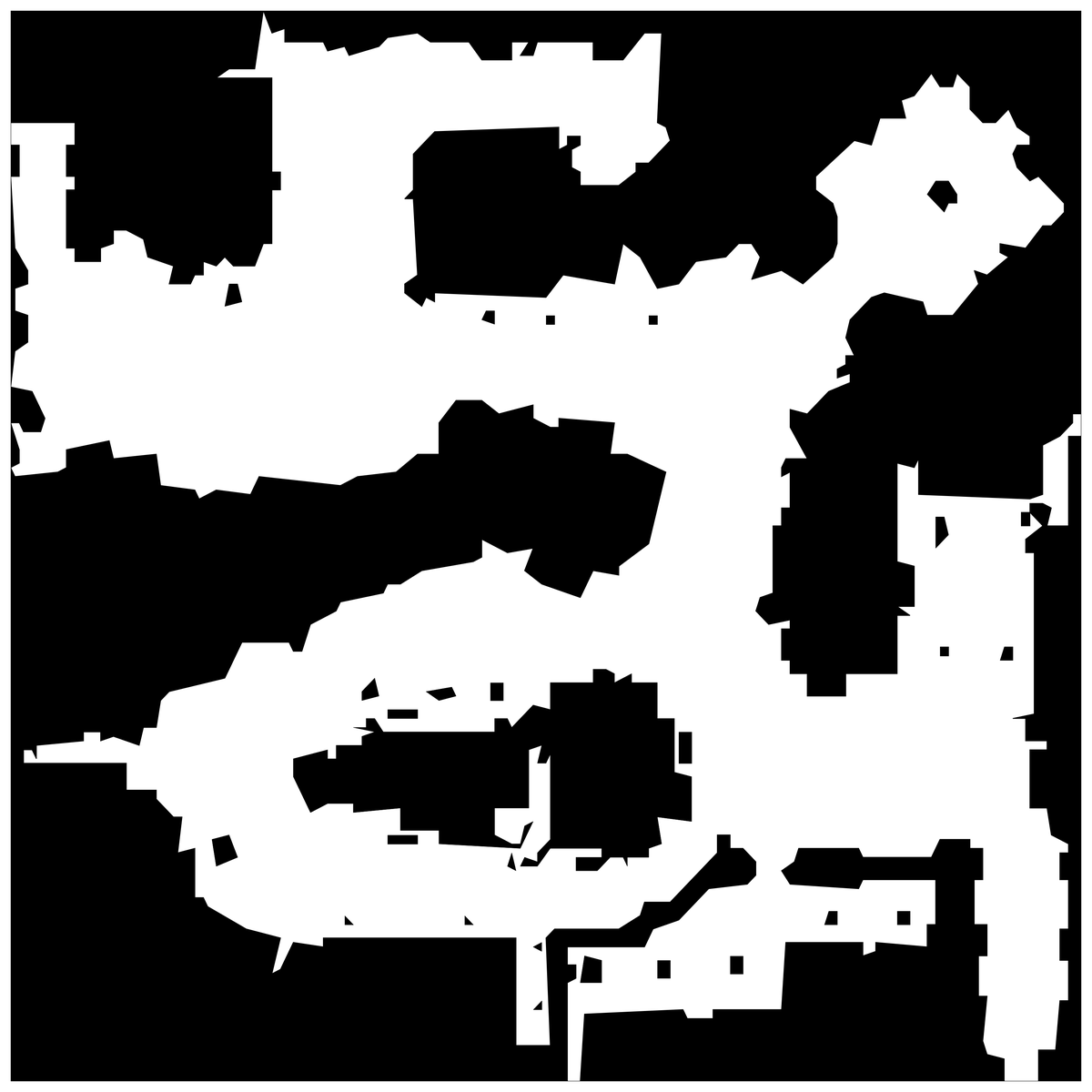}}
\subfloat[Berlin]{
\includegraphics[width=0.32\linewidth]{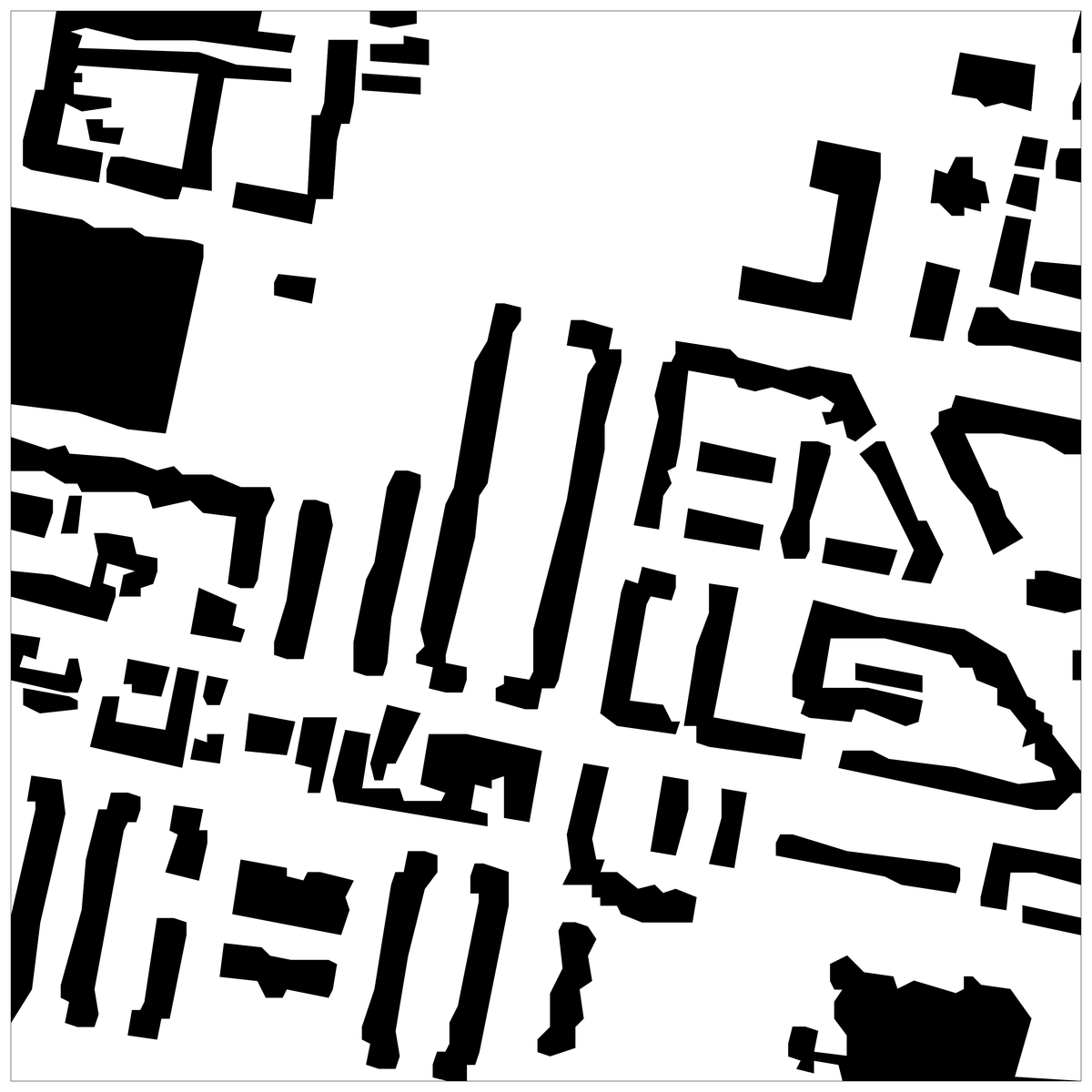}
}
\subfloat[maze]{
\includegraphics[width=0.32\linewidth]{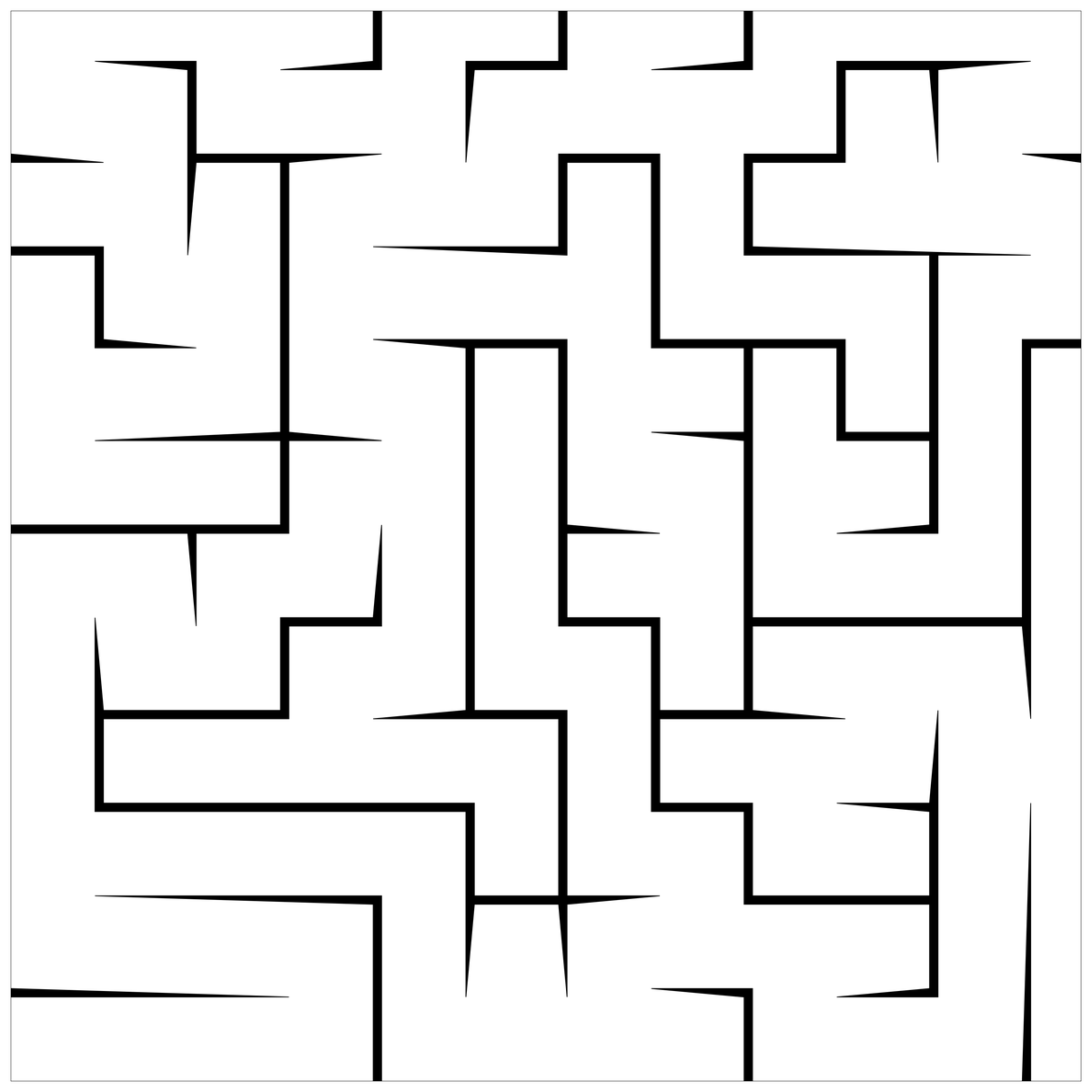}
}\\

\hfill
\subfloat[random]{
\includegraphics[width=0.33\linewidth]{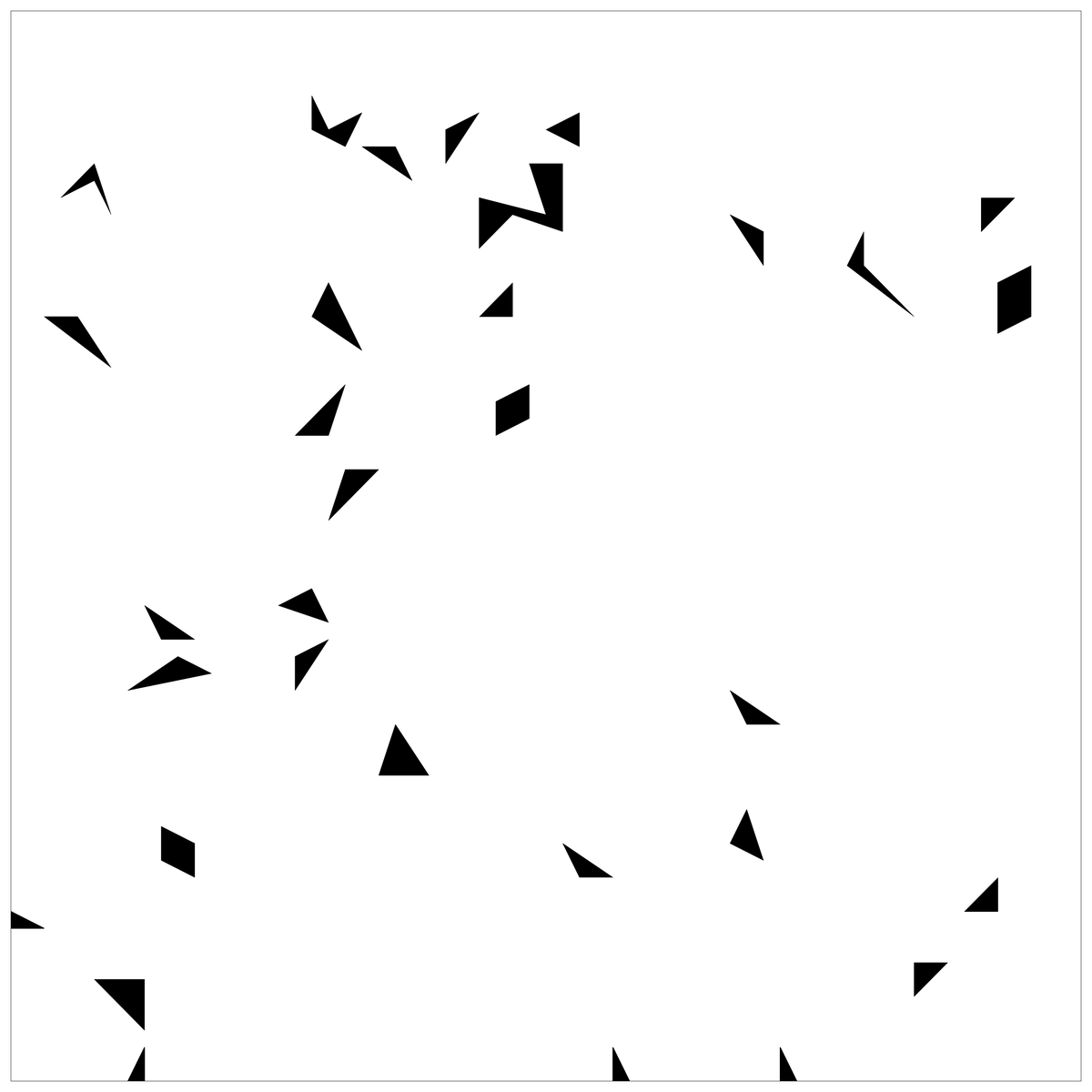}
}\hfill
\subfloat[warehouse]{
\includegraphics[width=0.33\linewidth]{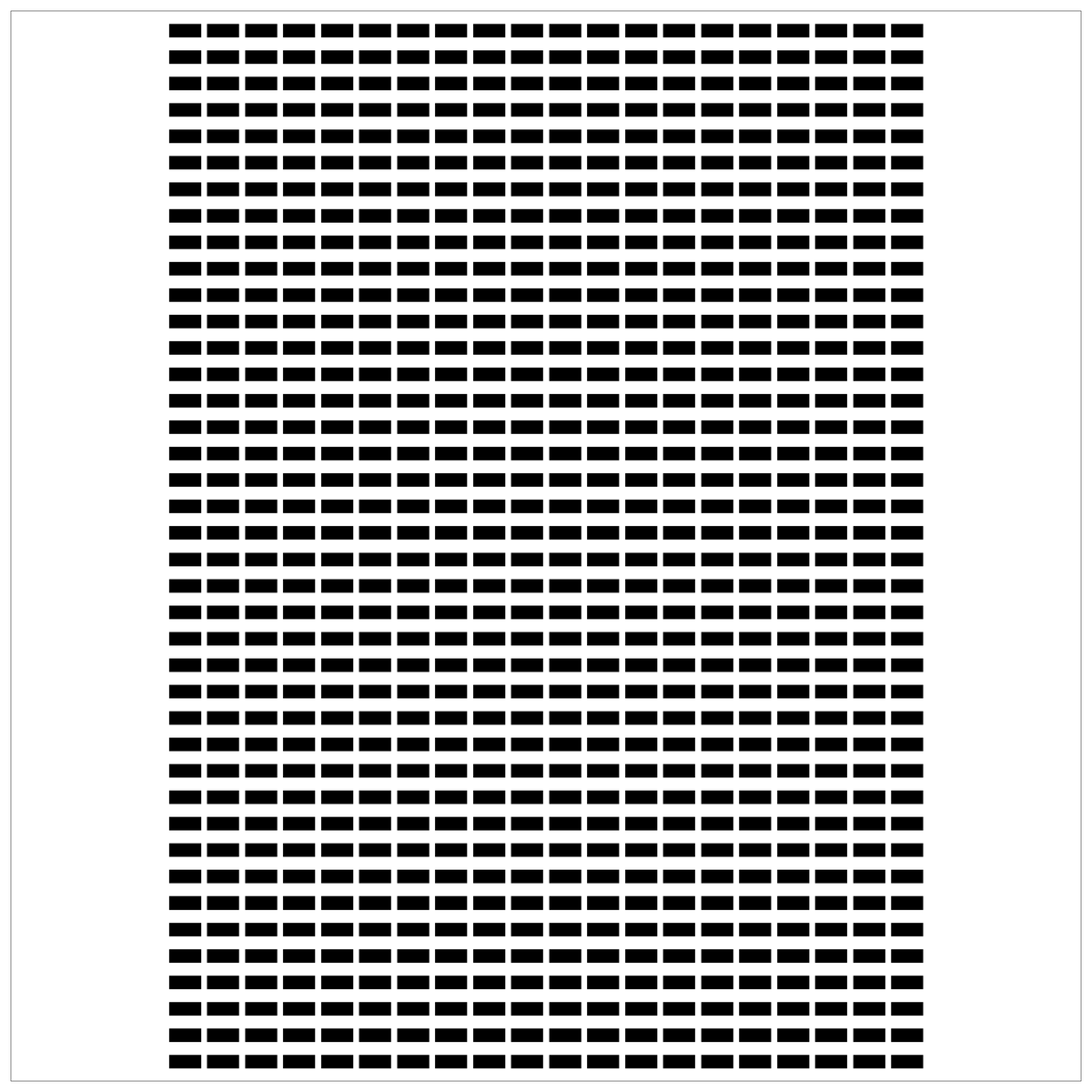}
}\\

\caption{Static environments used in the experiments.}
\label{fig:static_envs}
\end{figure}

The simulation setup follows the protocol in \cite{kasaura2022prioritized}. We consider a standard suite of environments from the MAPF benchmark, consistently with prior works \cite{sharon2013increasing, ma2019searching, andreychuk2021improving}. Starting from the original grid maps for discrete-time MAPF, we automatically generate corresponding roadmaps used in our continuous-time planning domain.
The evaluated environments are \texttt{den}, \texttt{Berlin}, \texttt{random}, \texttt{maze}, \texttt{warehouse}, \texttt{empty16}, and \texttt{empty256}. Table~\ref{tab:maps_stats} reports the main characteristics of the underlying grid maps, including their dimensions and obstacle density, while Fig.~\ref{fig:static_envs} provides a visual overview of the considered environments.

\begin{table}[h]
\centering
\caption{Characteristics of the MAPF benchmark environments.}
\label{tab:maps_stats}
\begin{tabular}{lcc}
\toprule
{Environment} & {Grid size} & {Obstacle density} \\
\midrule
\texttt{den}       & $256{\times}257$ & $57.2\%$ \\
\texttt{Berlin}    & $256{\times}256$ & $24.2\%$ \\
\texttt{random}    & $64{\times}64$     & $10.0\%$ \\
\texttt{maze}      & $128{\times}128$ & $33.7\%$ \\
\texttt{warehouse} & $340{\times}164$ & $30.5\%$ \\
\texttt{empty16}   & $16{\times}16$   & $0\%$ \\
\texttt{empty256}  & $256{\times}256$ & $0\%$ \\
\bottomrule
\end{tabular}
\end{table}

All experiments are conducted on a notebook equipped with an Intel Core i7-12700H CPU (2.30~GHz, 14 cores, 20 threads) and 16~GB DDR4 RAM. Following \cite{kasaura2022prioritized}, each grid environment is converted into a geometric roadmap $G=(V,E)$ for continuous-time planning by sampling vertices in free space and adding bidirectional edges with either k-nearest simplified probabilistic roadmap method (k-sPRM, $k=15$) \cite{karaman2011sampling} or constrained Delaunay triangulation (CDT) \cite{chew1987constrained}. Agents are modeled as discs of radius $r=0.5$ moving at constant speed along straight-line edges, with traversal time proportional to the Euclidean edge length, and collisions are checked in continuous time via exact geometric overlap of disc trajectories.

For every environment--roadmap combination we compare CPBS and Lazy-CPBS against PSIPP-CTC (hereafter also referred to as PP) and CCBS (augmented with the enhancements of \cite{andreychuk2021improving}), by considering an increasing number of agents: for each map we first sample a pool of $n_\mathrm{max}$ start-goal pairs $\{(s_i,g_i)\}_{i=1}^{n_\mathrm{max}}$ uniformly at random from roadmap vertices, enforcing unique start and goal locations, and then build instances with $n \in \{1,\ldots,n_\mathrm{max}\}$ agents by taking the first $n$ pairs from the pool so that all methods are evaluated on identical instance sets while progressively increasing the level of roadmap congestion. We repeat this procedure for 25 independent pools and impose a per-run runtime cap of 30\,s; an instance is counted as failed if no solution is found within the limit.
For CCBS, we consider \textit{online} constraint/safe-interval construction (i.e., not an augmented variant CCBS-CTC). In this disc-on-segment setting, geometric conflict checking and the induced safe-interval reconstruction are a marginal component of the total runtime; moreover, each high-level CCBS expansion branches on a single conflicting pair of moves and replans only one agent, so constant-time access to pre-annotated roadmap collision data (CTC) did not yield a noticeable speed-up in our experiments---unlike in solvers where many pairwise interactions must be repeatedly evaluated (e.g., PP, CPBS, and Lazy-CPBS). This is partly because, in this setting, collision checks reduce to inexpensive distance computations between disc trajectories, whereas in more complex settings (e.g., AGV-MAPF$_R$-CTC) collision detection is more costly and CCBS-CTC can provide gains.

Finally, for reproducibility, our C++ implementation is publicly available online\footnote{{\url{https://github.com/FedericoSignorile/Lazy-CPBS}}} and interfaces with the authors' official implementations of PSIPP-CTC\footnote{\url{https://github.com/omron-sinicx/PSIPP-CTC}.} and CCBS\footnote{\url{https://github.com/thaynewalker/CCBS}.}.

\subsection{Results}
\begin{figure}[t]
\centering

\subfloat[100 CDT]{%
  \includegraphics[width=0.24\linewidth]{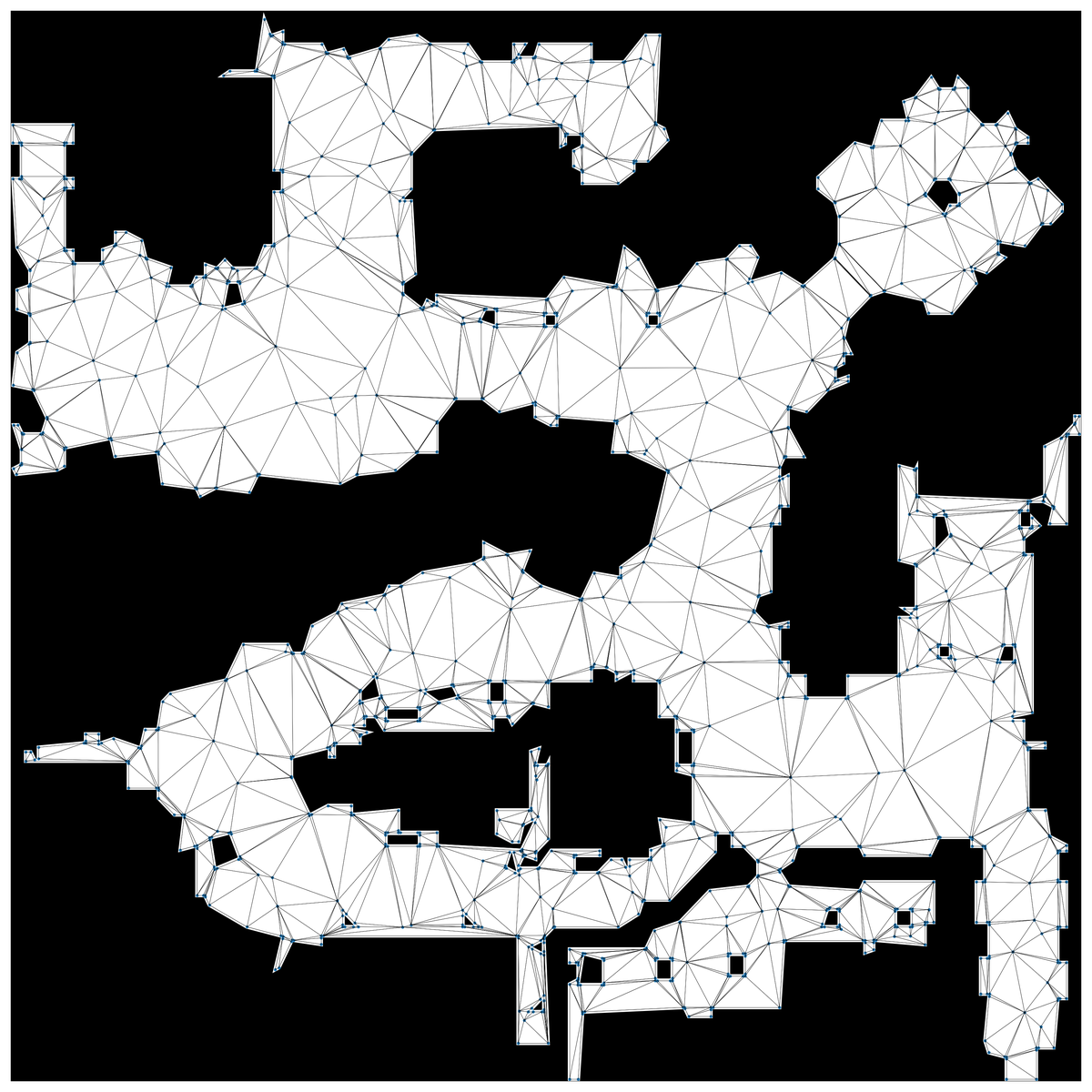}%
}\hfill
\subfloat[100 k-sPRM]{%
  \includegraphics[width=0.24\linewidth]{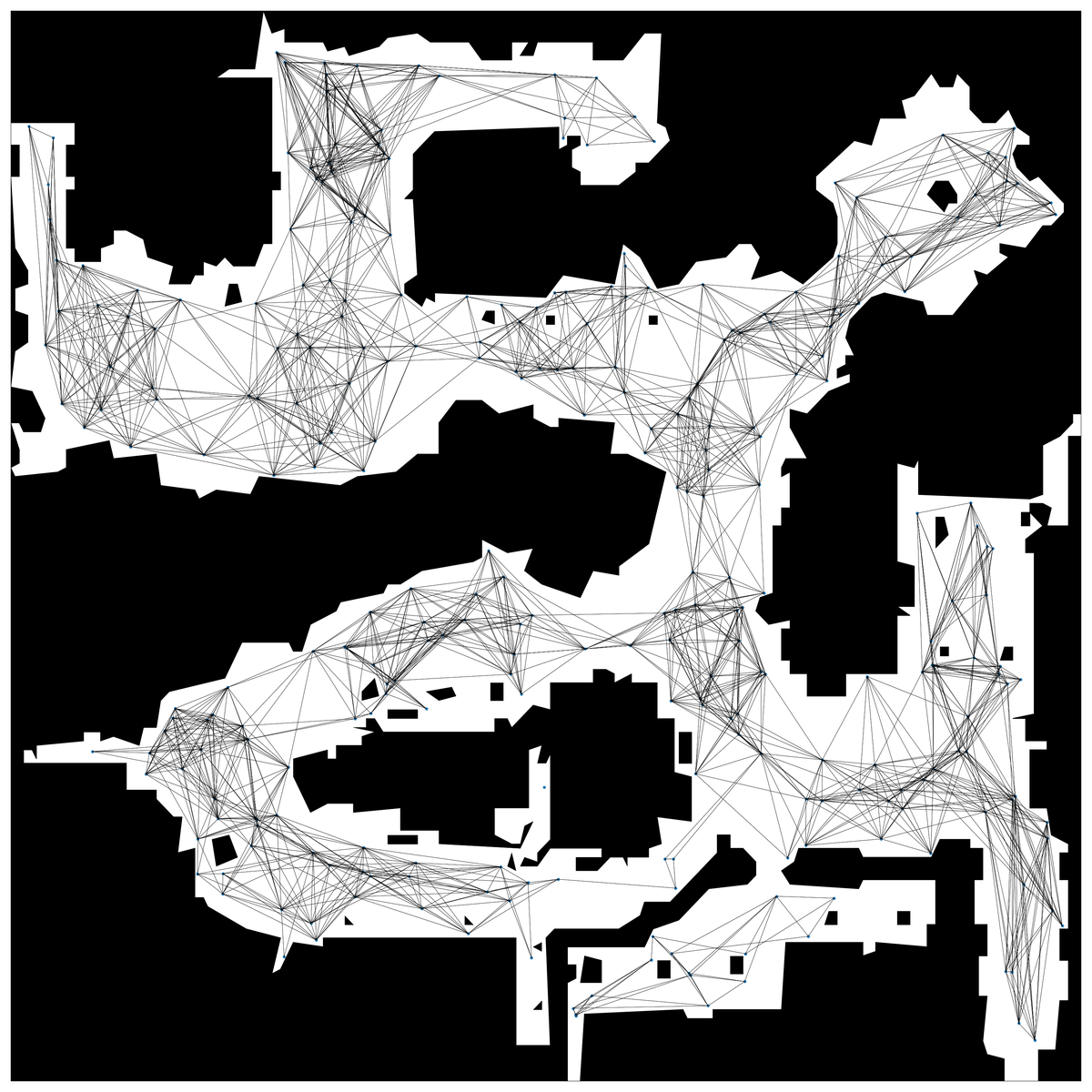}%
}\hfill
\subfloat[700 CDT]{%
  \includegraphics[width=0.24\linewidth]{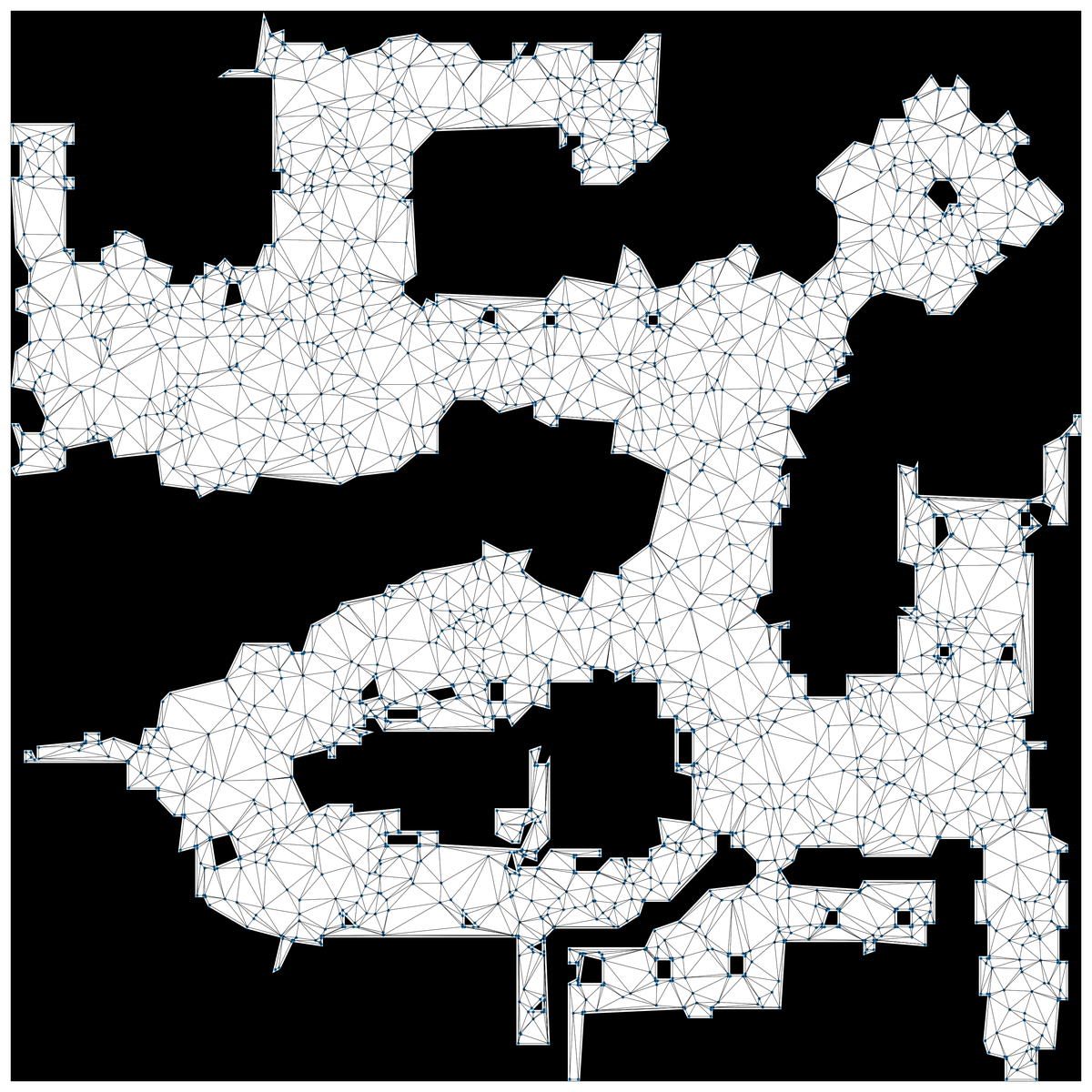}%
}\hfill
\subfloat[700 k-sPRM]{%
  \includegraphics[width=0.24\linewidth]{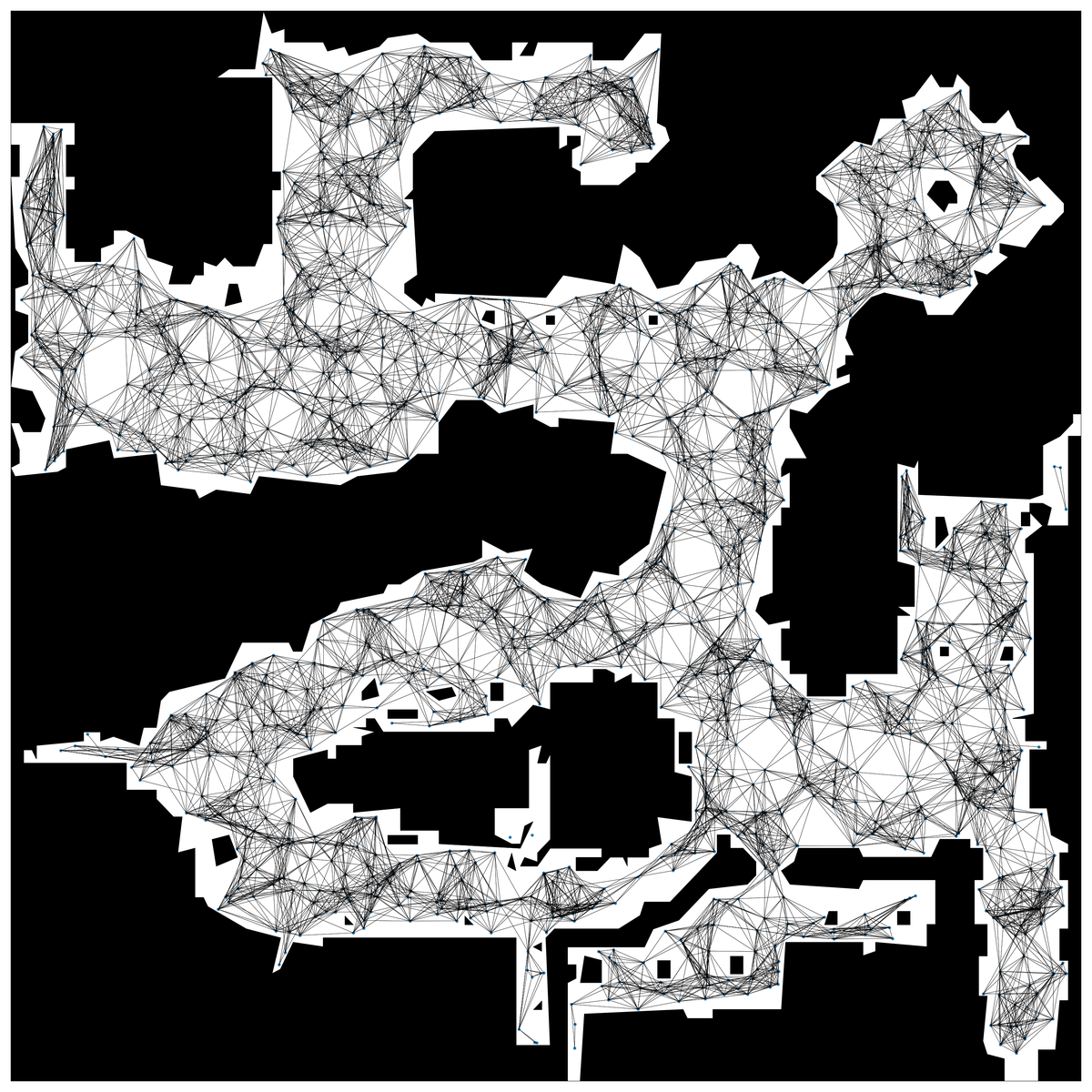}%
}

\caption{Geometric roadmaps for the \texttt{den} environment.}
\label{fig:den-roadmaps}
\end{figure}

As done in \cite{andreychuk2021improving,andreychuk2022multi,kasaura2022prioritized}, we use the \texttt{den} environment as our baseline testbed. For this environment, we consider four roadmaps obtained by combining the CDT and k-sPRM generators with two sampling budgets (100 and 700 vertices). Fig.~\ref{fig:den-roadmaps} visualizes the resulting roadmaps. Overall, k-sPRM produces roadmaps with a higher edge density due to the value of $k = 15$, offering many alternative routing possibilities. In contrast, CDT tends to place samples more effectively in cluttered regions, often yielding feasible transitions through narrow bottlenecks and improving connectivity in constrained passages.

\begin{figure}[t]
\centering

\subfloat[100 CDT]{
\includegraphics[width=0.32\linewidth]{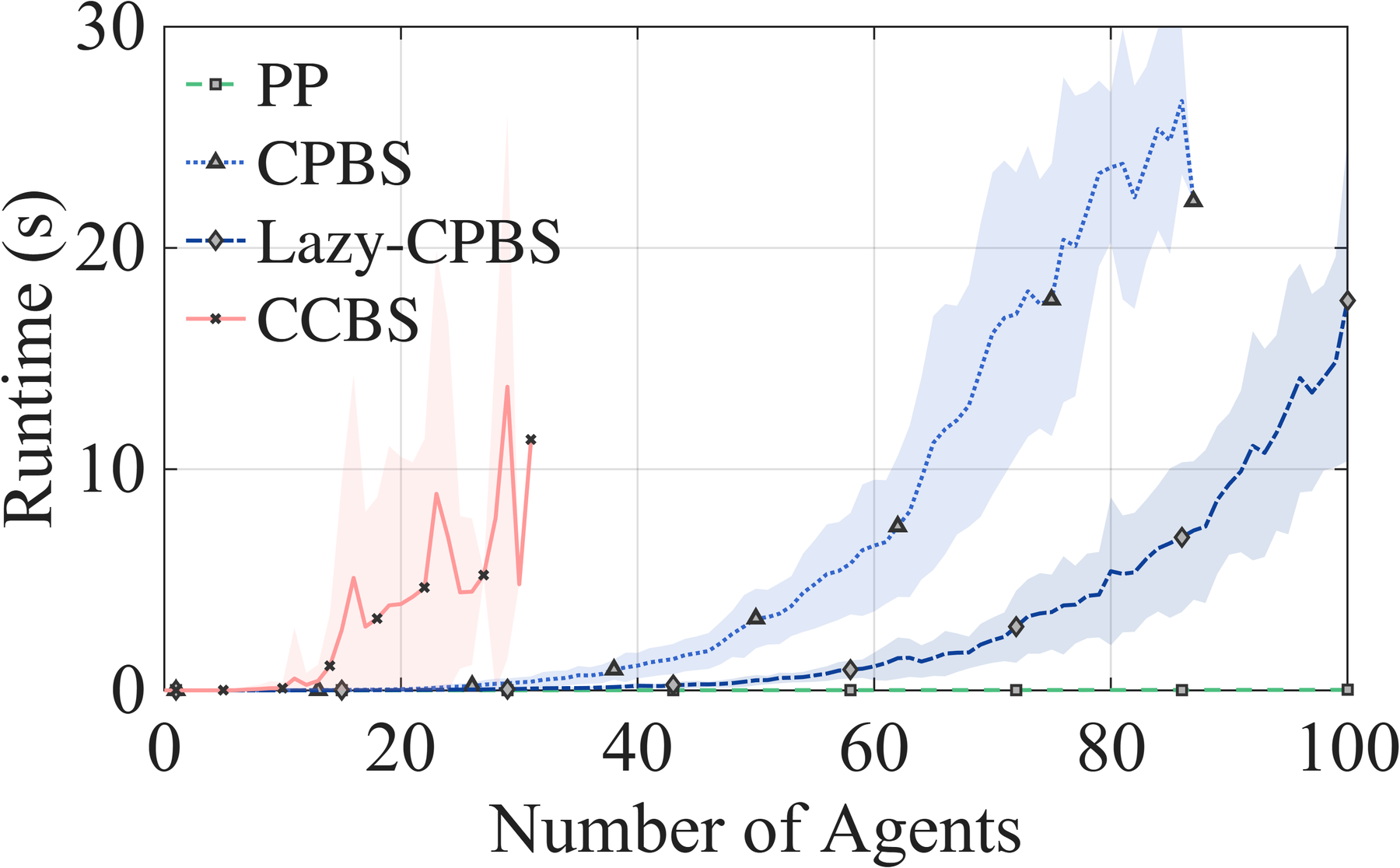}
\includegraphics[width=0.32\linewidth]{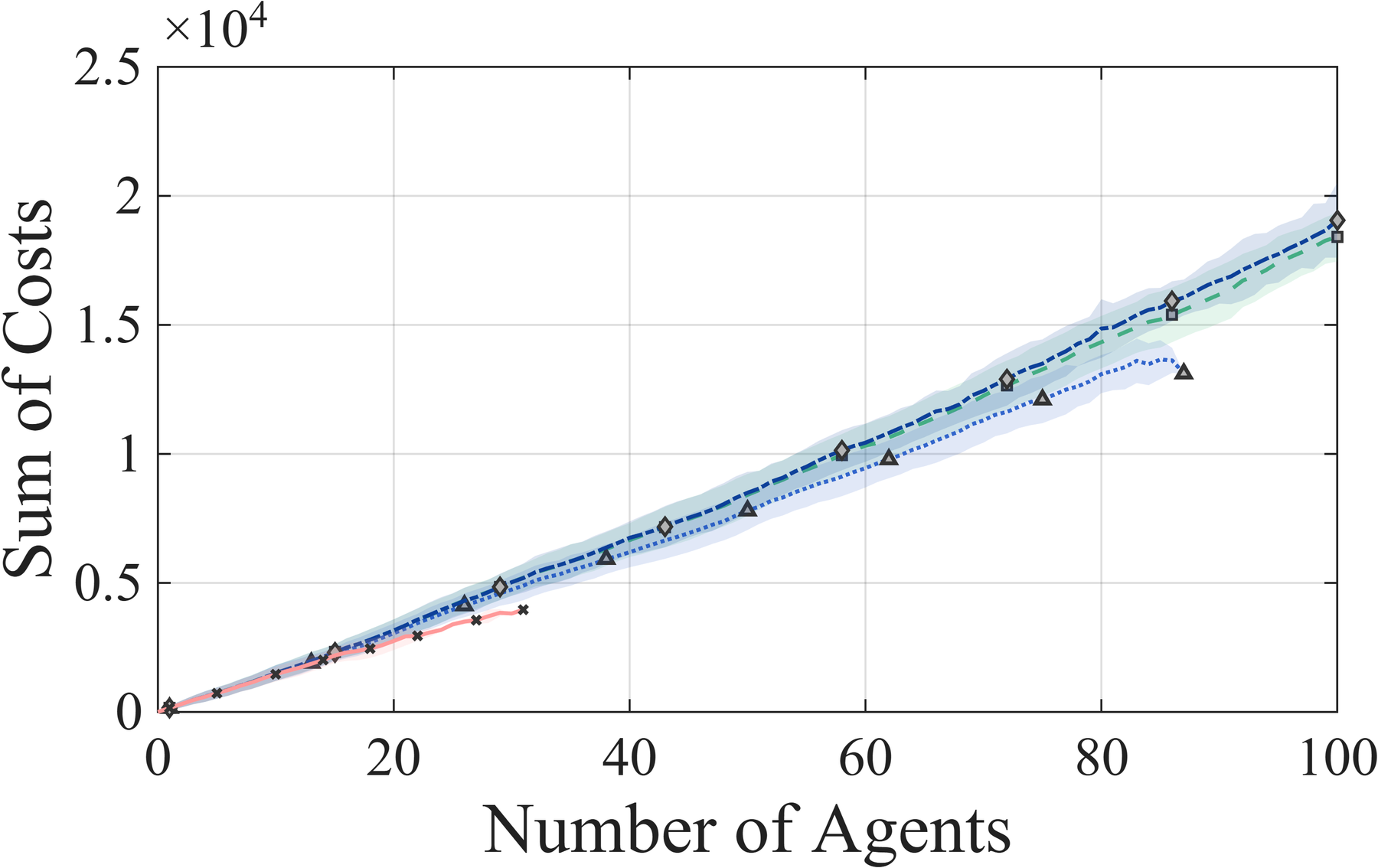}
\includegraphics[width=0.32\linewidth]{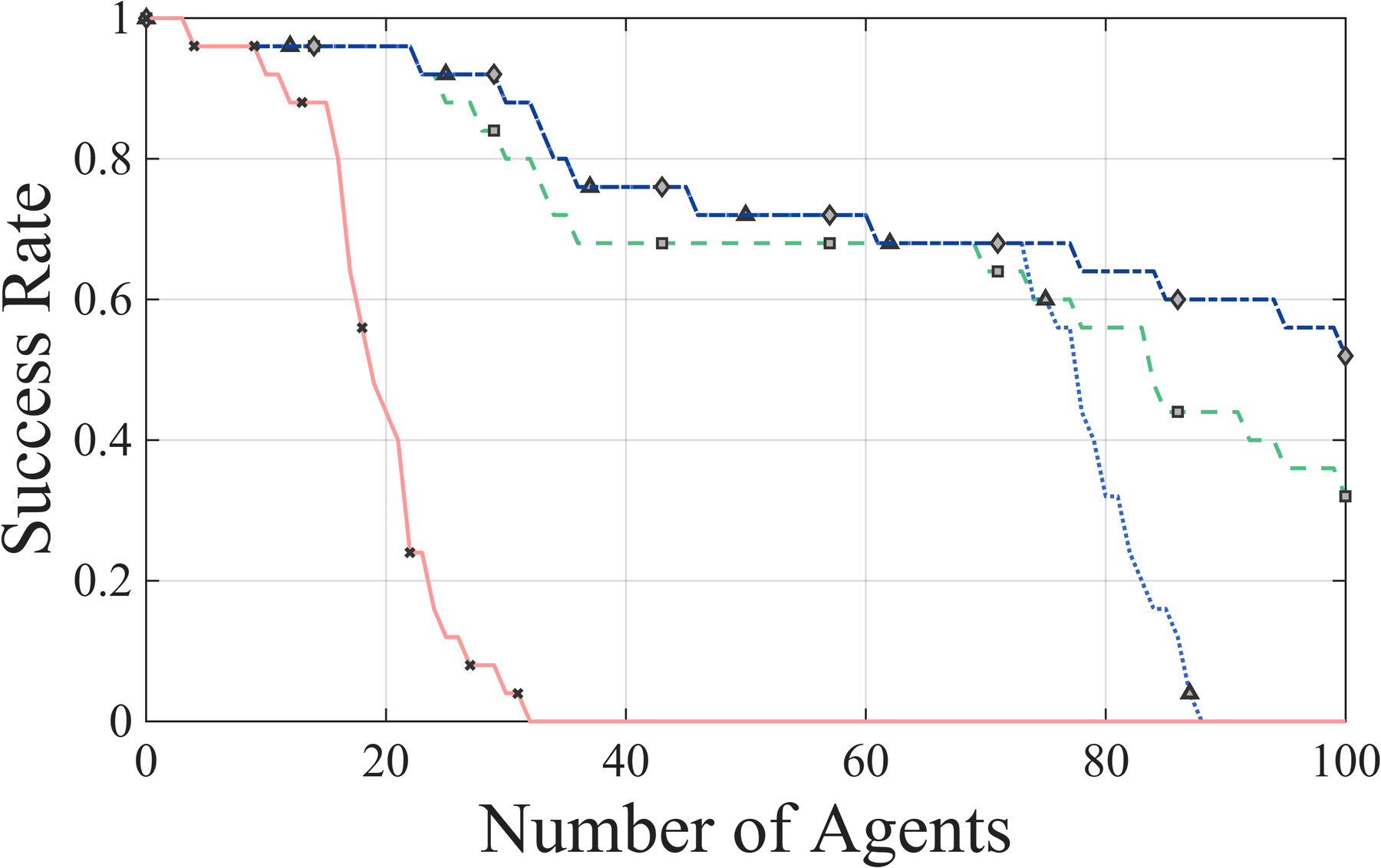}
} \\

\subfloat[100 k-sPRM]{
\includegraphics[width=0.32\linewidth]{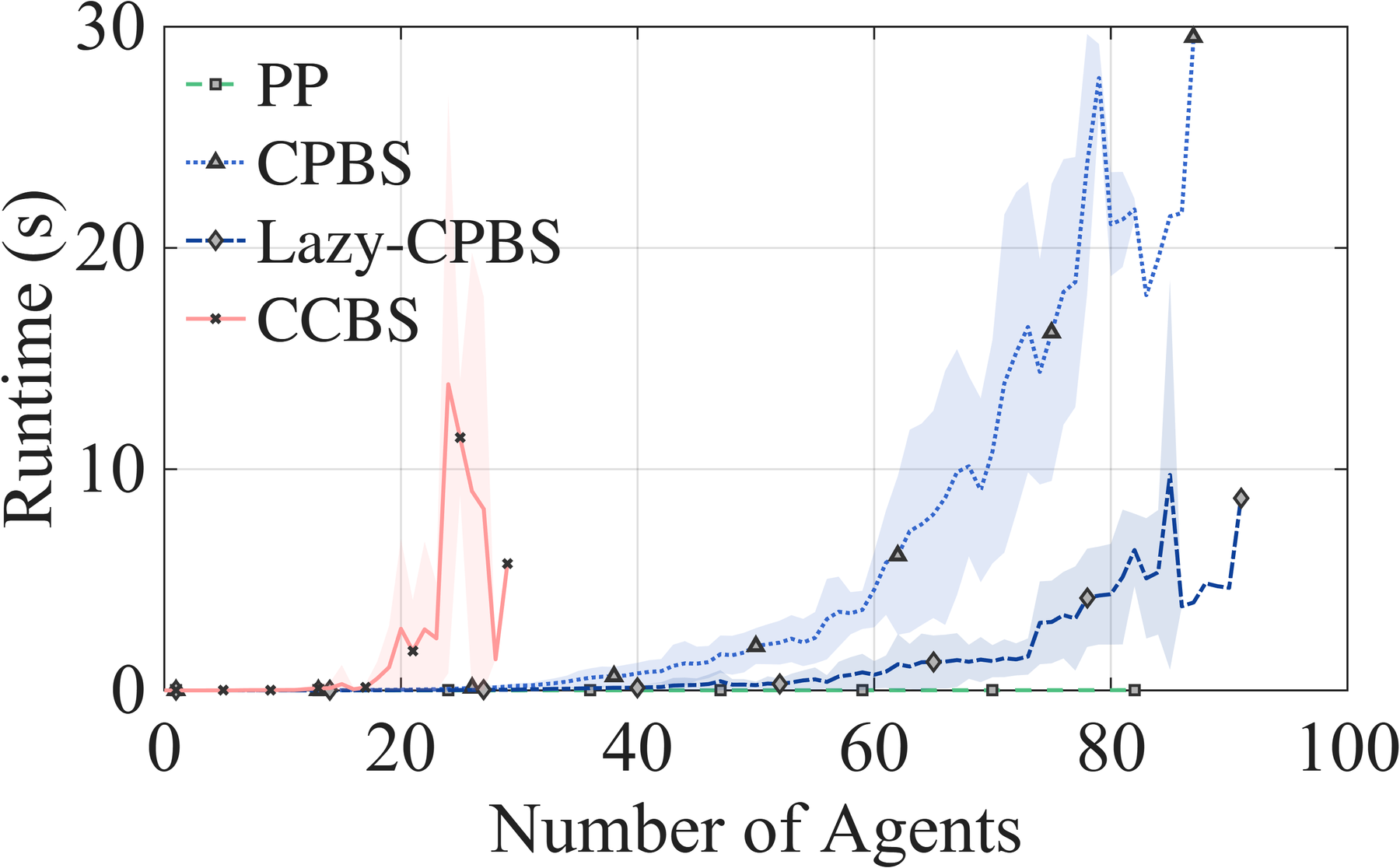}
\includegraphics[width=0.32\linewidth]{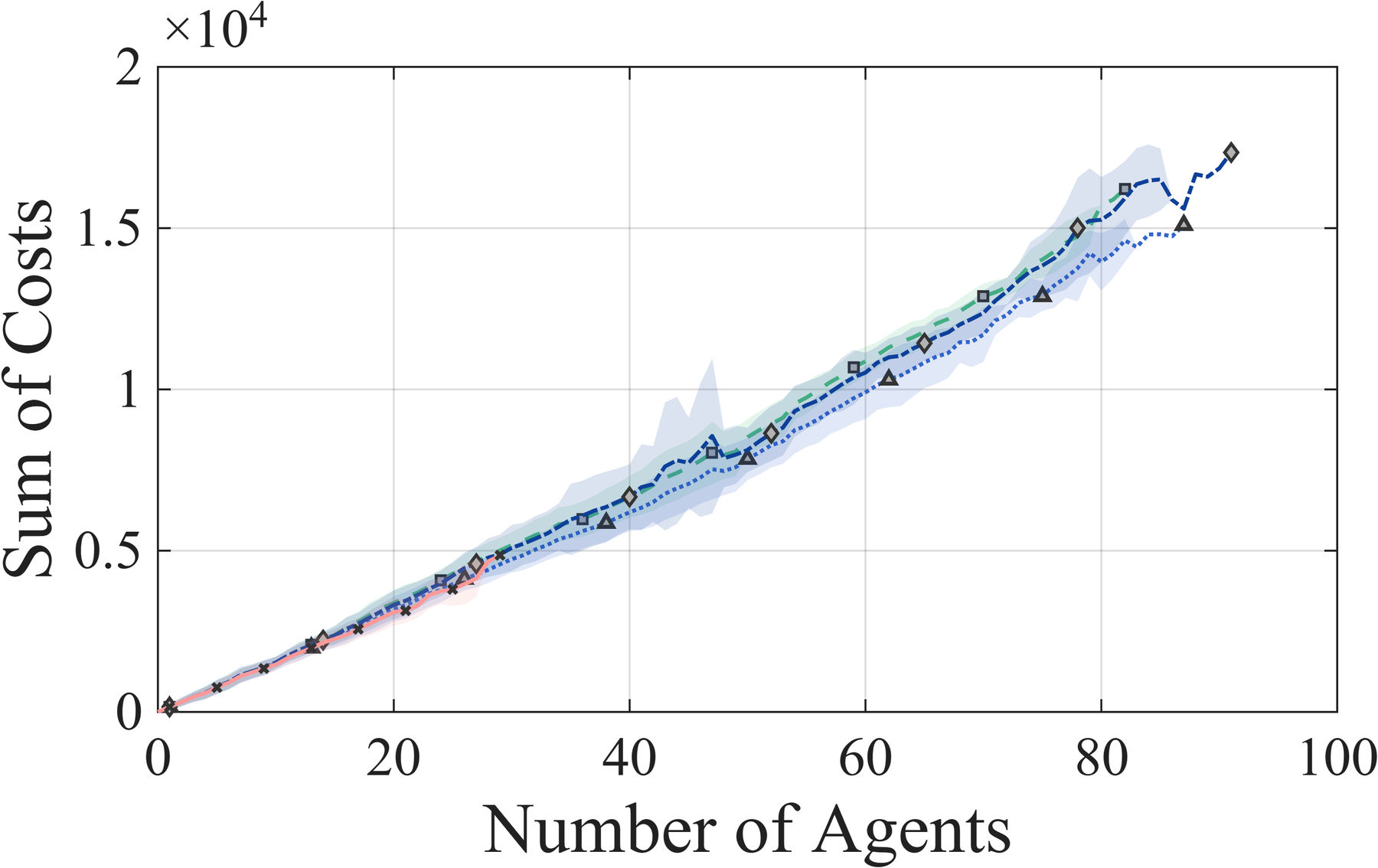}
\includegraphics[width=0.32\linewidth]{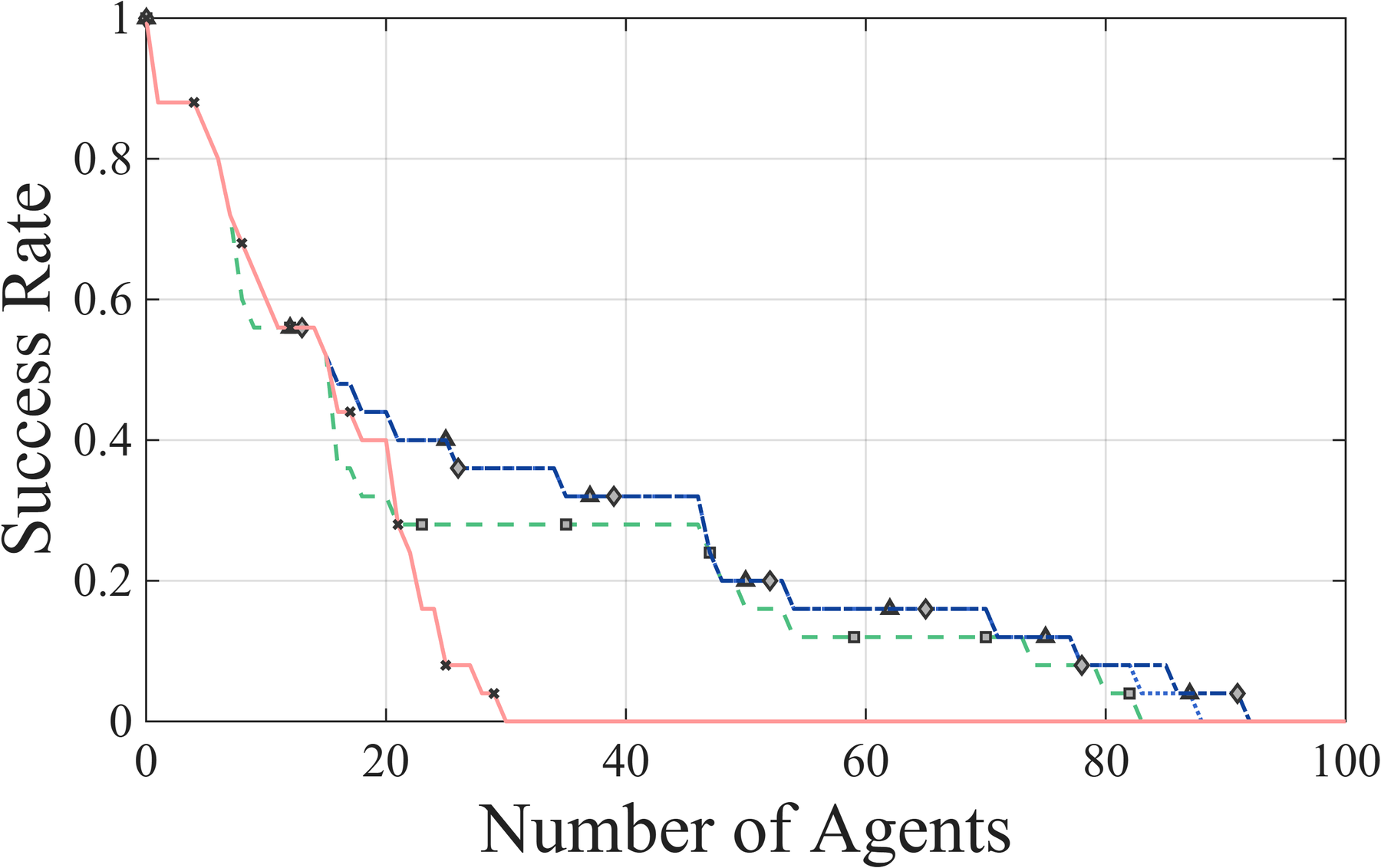}
}\\

\subfloat[700 CDT]{
\includegraphics[width=0.32\linewidth]{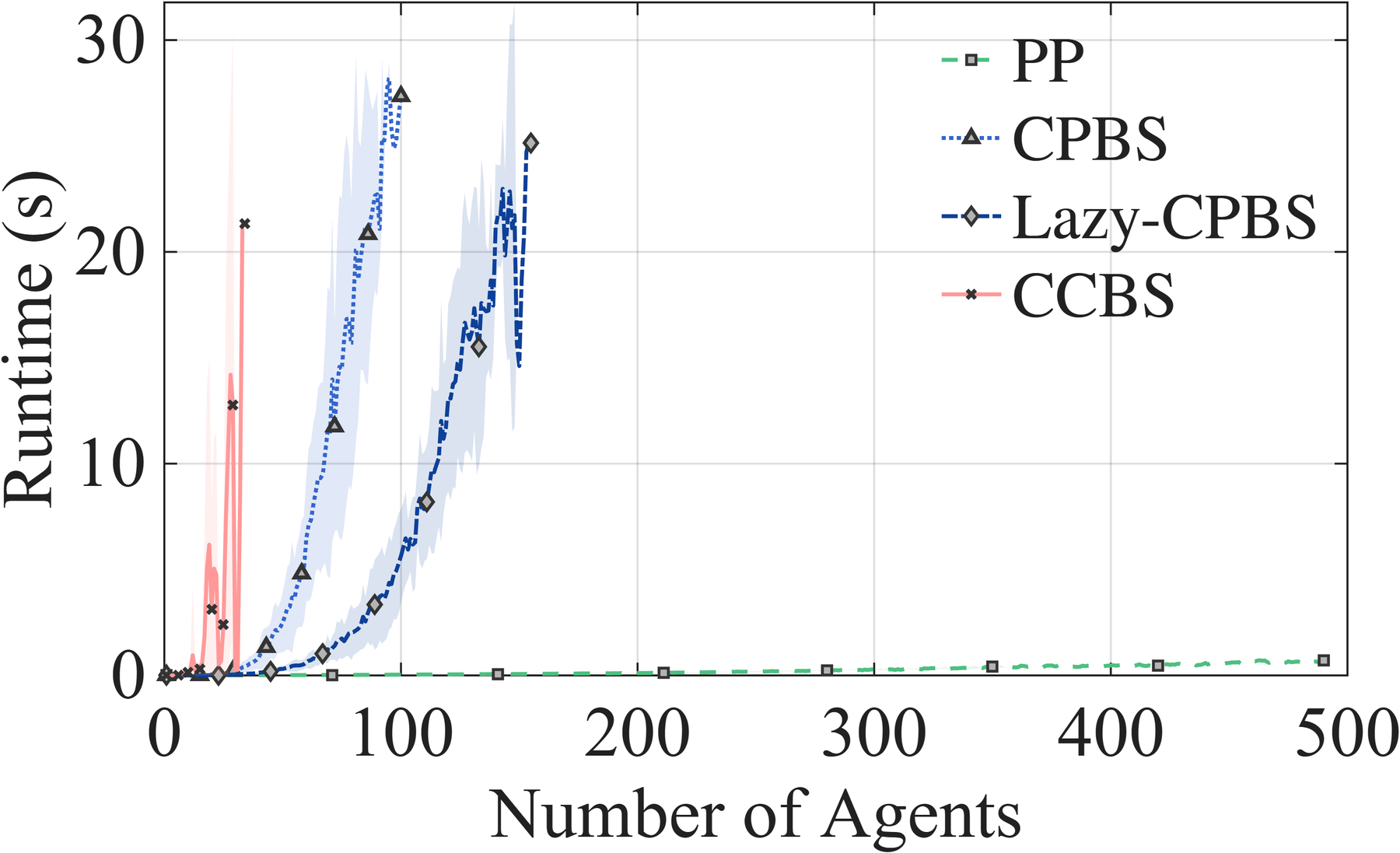}
\includegraphics[width=0.32\linewidth]{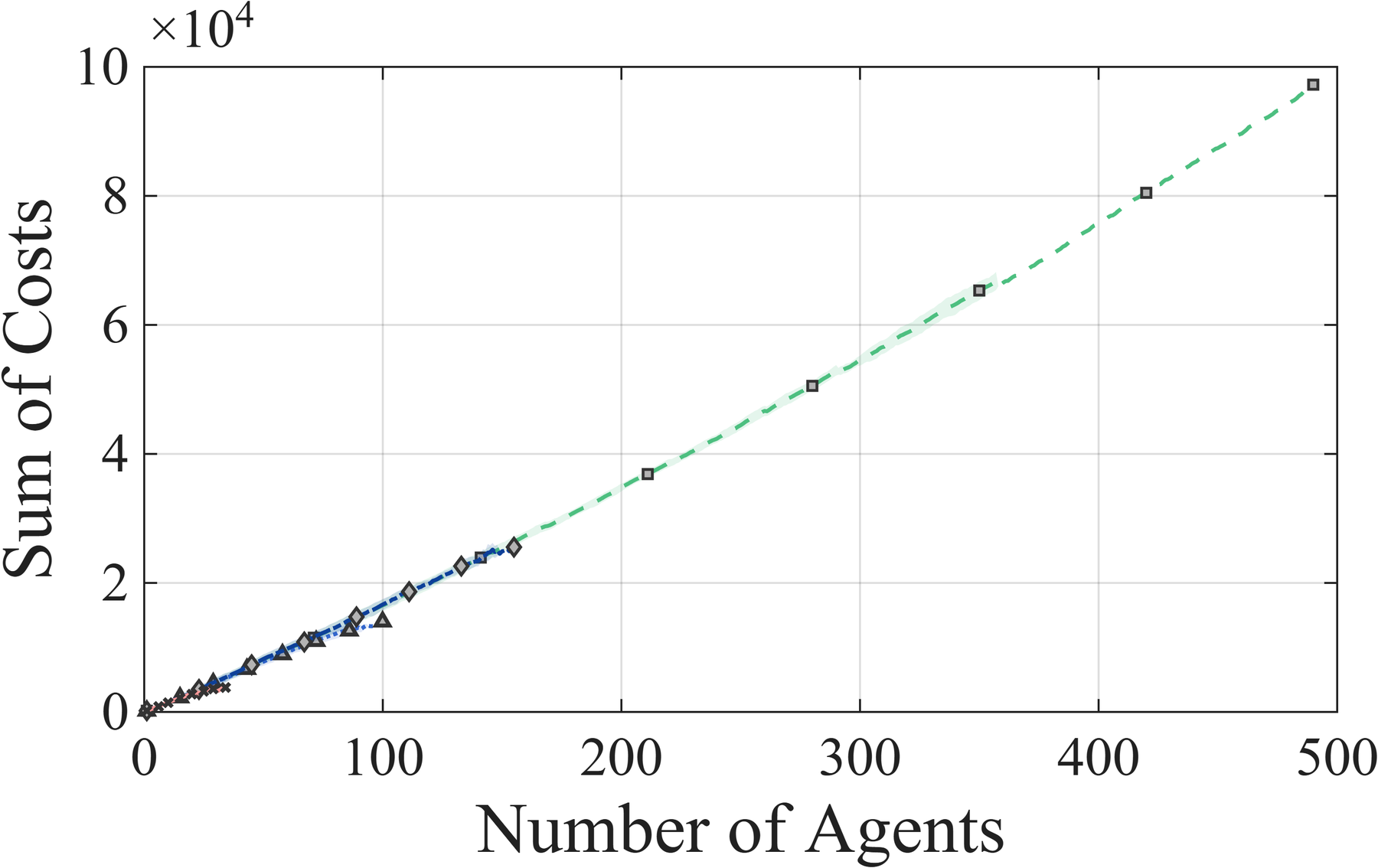}
\includegraphics[width=0.32\linewidth]{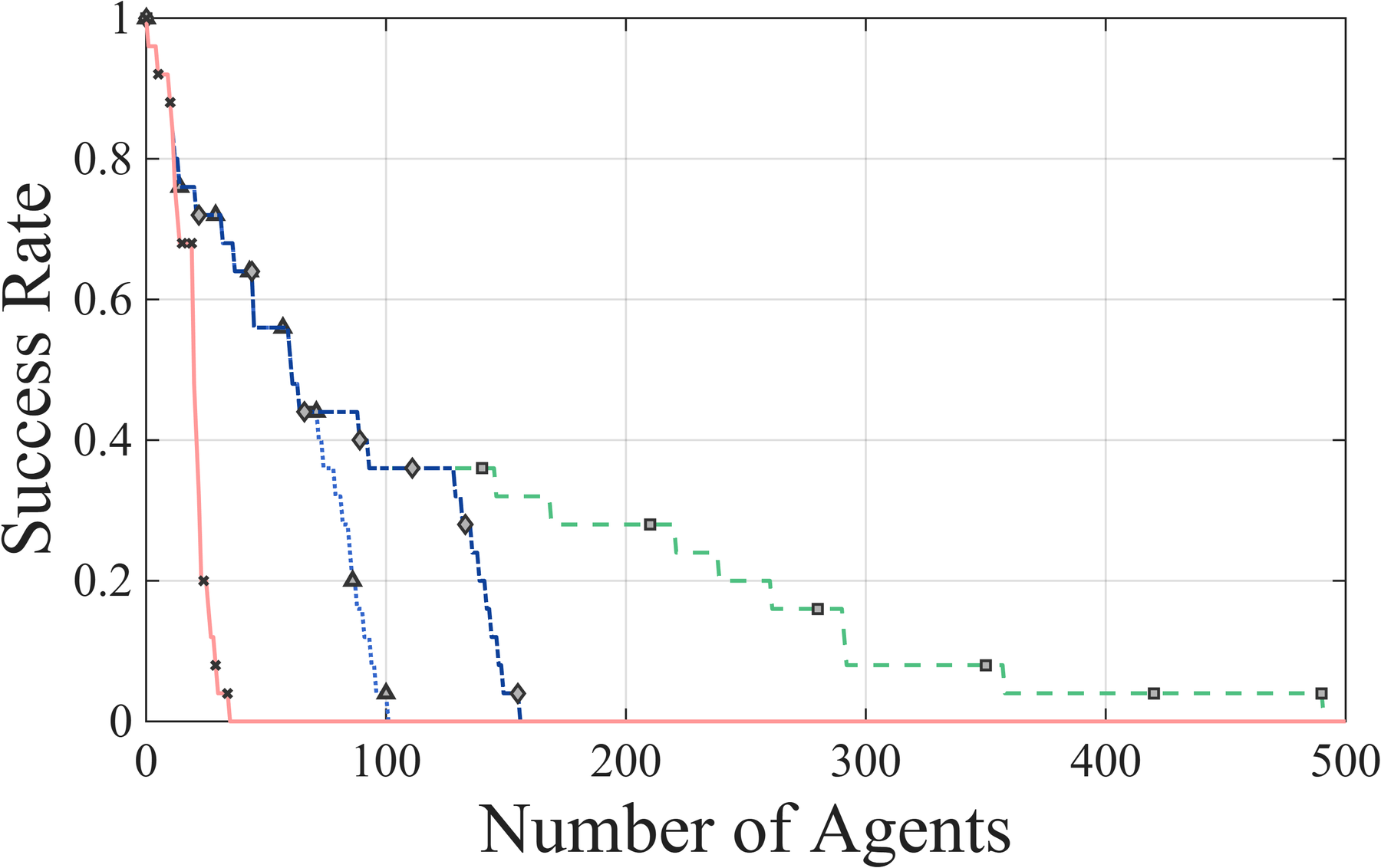}
}\\

\subfloat[700 k-sPRM]{
\includegraphics[width=0.32\linewidth]{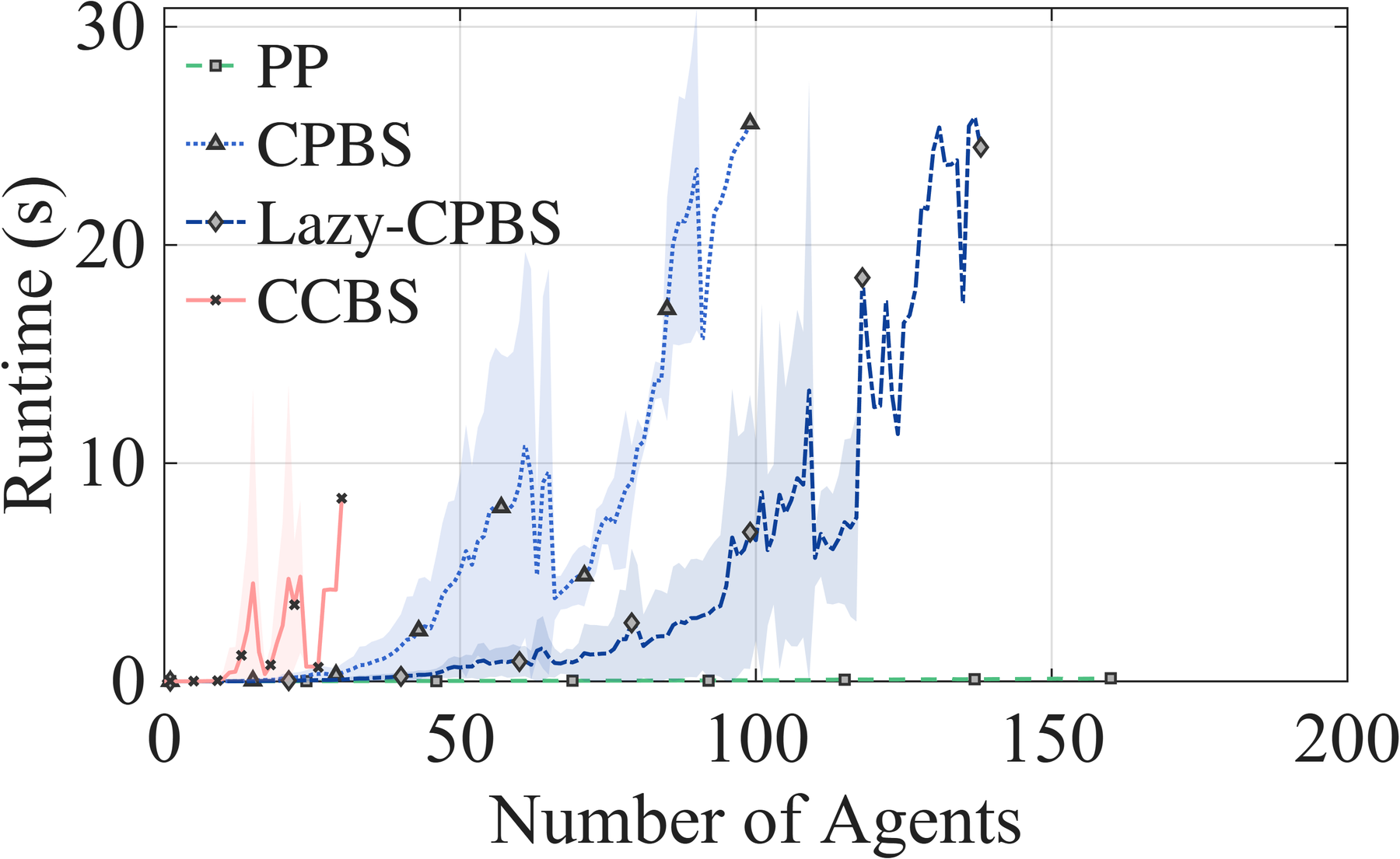}
\includegraphics[width=0.32\linewidth]{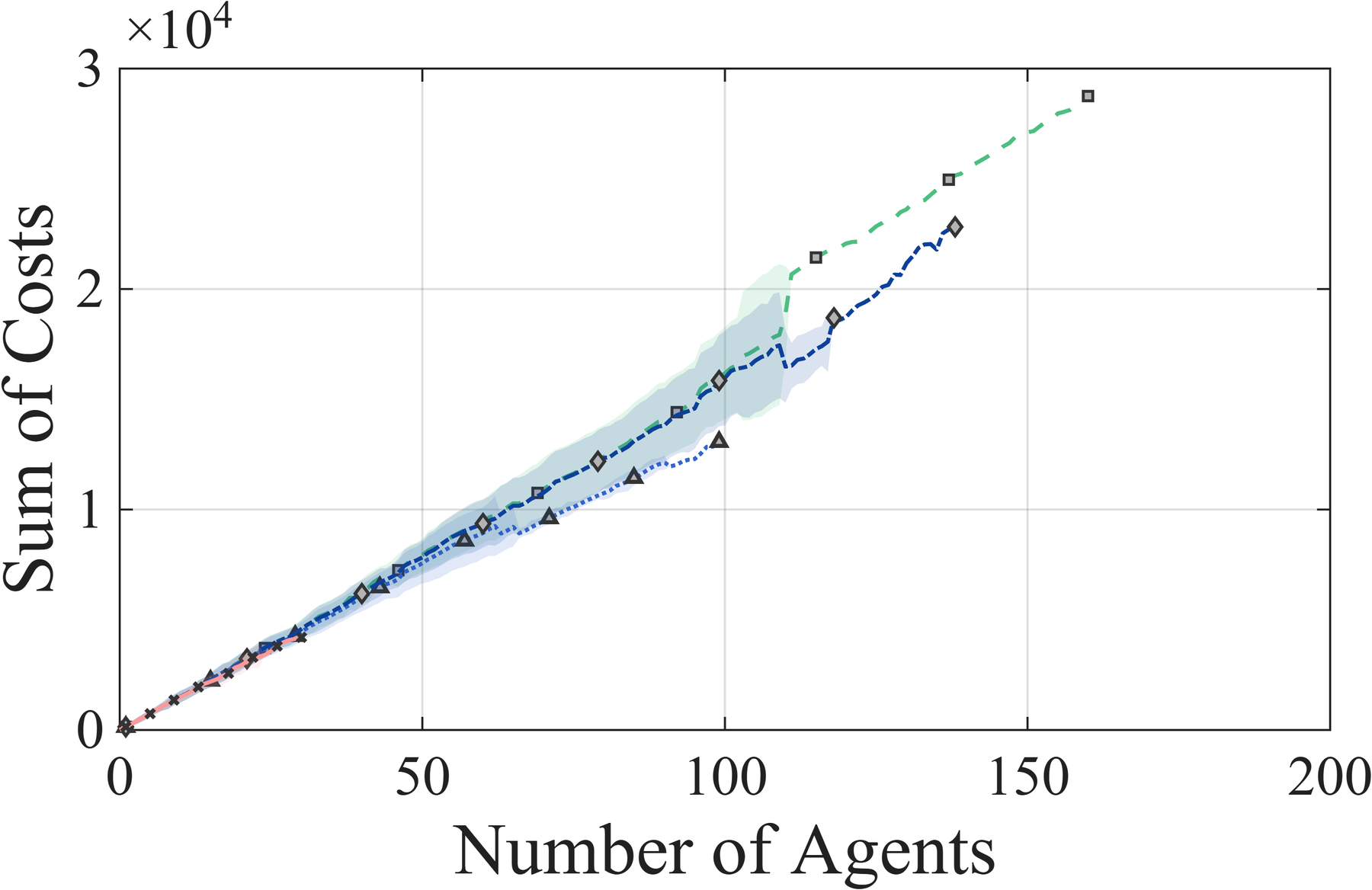}
\includegraphics[width=0.32\linewidth]{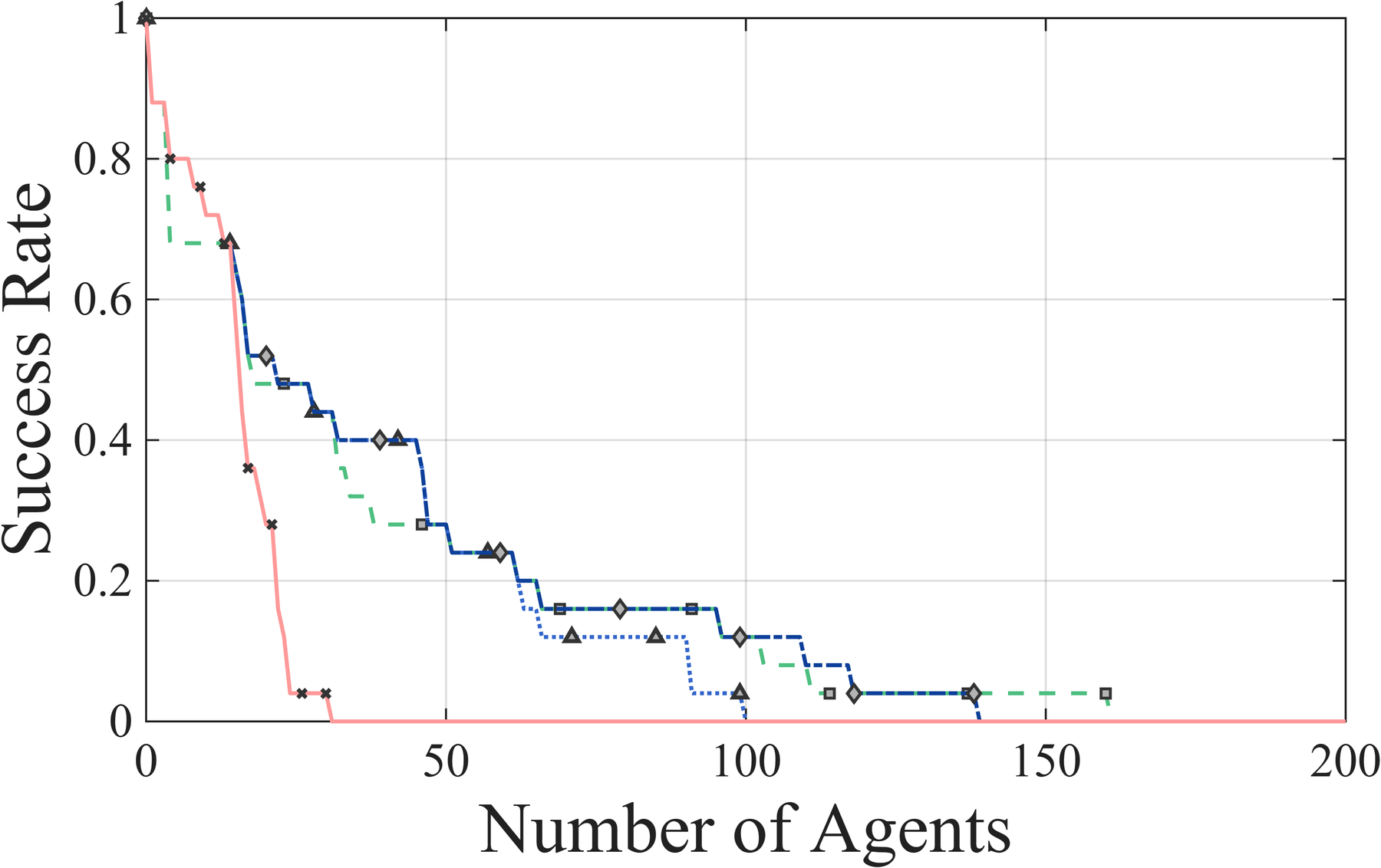}
}

\caption{Experimental results across \texttt{den} roadmaps. Each row shows runtime (left), SoC (middle) and success rate (right). Lines show the mean over solved instances and shaded bands $\pm 1$ standard deviation.}
\label{fig:den-results}
\end{figure}

The results of the simulations on the four roadmaps are reported in Fig.~\ref{fig:den-results}.
For each roadmap and algorithm, runtime, SoC of the computed solutions, and success rate are shown as a function of the number of agents.

\begin{figure*}
    \centering
    \includegraphics[width=0.98\linewidth]{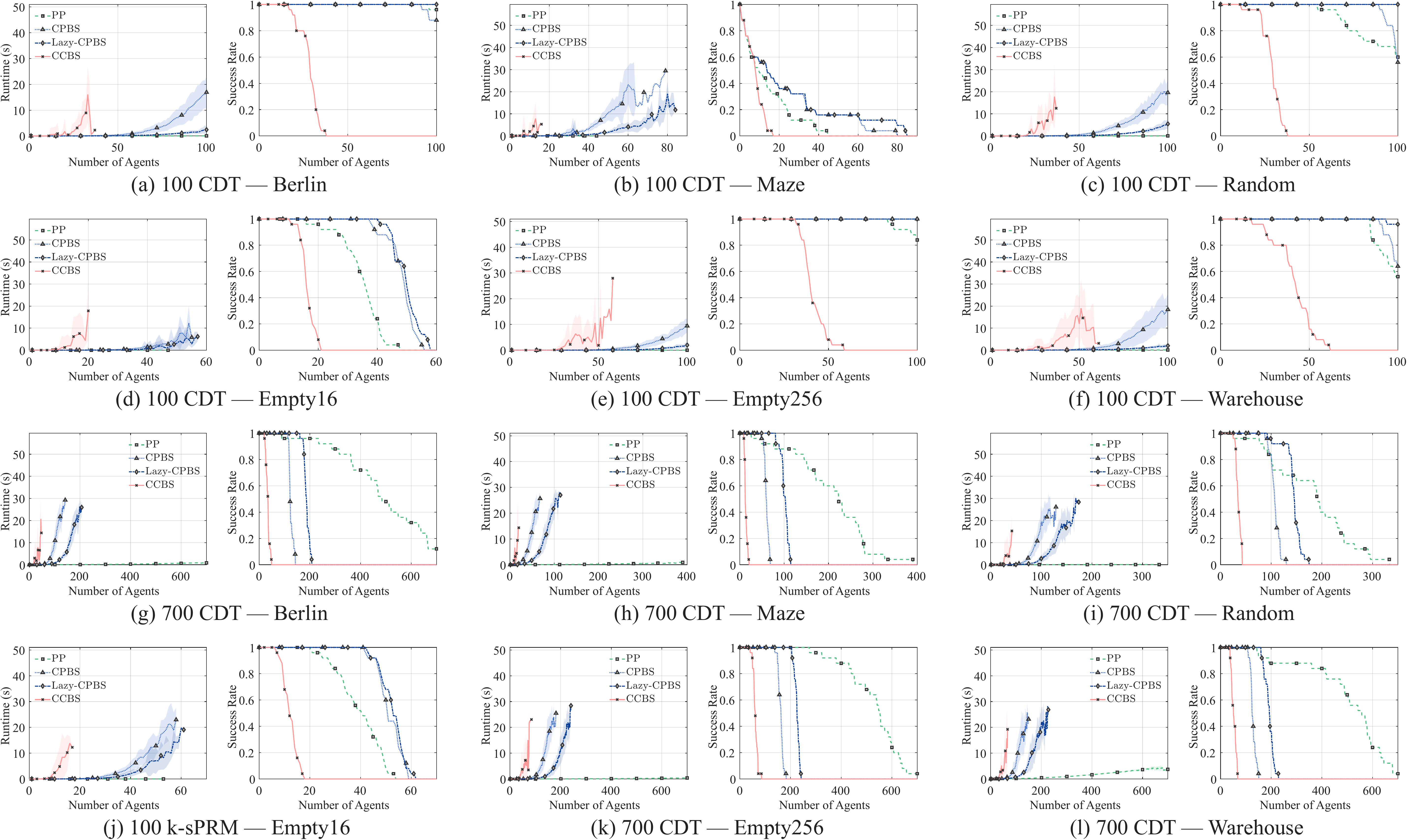}
    \caption{Runtime (left) and success rate (right) across benchmark environments and roadmap configurations. Lines show the mean over solved instances and shaded bands $\pm 1$ standard deviation.}
    \label{fig:benchmark-results}
\end{figure*}

Across all four \texttt{den} roadmaps, CCBS solves only instances with few agents and then quickly becomes impractical as conflicts proliferate. This is expected for CBS-style optimal solvers: when it succeeds it achieves the lowest SoC, but its scalability remains limited in this continuous-time roadmap setting.

Among the non-optimal solvers, PP ({PSIPP-CTC}) is consistently the fastest, but its fixed priority order makes it fragile once the problem instances become congested. In contrast, CPBS and Lazy-CPBS trade additional computation for robustness by revising priority orders, obtaining higher success deeper into the congested regime, until the runtime limit becomes dominant.
When all methods succeed, solution quality follows a stable ordering: CCBS yields the best (lowest) SoC, CPBS is typically close but slightly higher, and Lazy-CPBS is slightly worse than CPBS, often comparable to PP.

Success rates are highly sensitive to roadmap scale and to how quickly instances become dense. On the smaller 100-vertex roadmaps, Lazy-CPBS outperforms the other methods as congestion emerges early. On 700~CDT, instead, PP achieves the highest success: the larger roadmap provides enough spatial slack that feasibility is preserved longer, and PP's negligible runtime lets it solve many instances before timeout becomes a limiting factor for the more expensive solvers. On 700~k-sPRM it is possible to observe a clear trade-off: PP and Lazy-CPBS end up with comparable success, with neither method uniformly dominating across agent densities.

Since CDT yields consistently higher success rates, the remainder of the benchmark campaign focuses on CDT roadmaps and is extended to all remaining environments by reporting results on \texttt{100~CDT} and \texttt{700~CDT}. The only exception is \texttt{empty16}, whose limited free space makes 700-vertex roadmap instances impractical; for this map, results are reported on \texttt{100~CDT} and \texttt{100~k-sPRM}. Fig.~\ref{fig:benchmark-results} reports runtime and success rate on the remaining environments.

The density-driven pattern largely carries over to the remaining environments. For instance, in the smallest map \texttt{empty16} (Fig.~\ref{fig:benchmark-results}(d) and (j)),
instances become congested almost immediately even in the absence of obstacles. In this regime, the fixed-order policy of PP breaks down early, whereas CPBS and Lazy-CPBS remain effective over a wider range of agent counts.

More broadly, on \texttt{100~CDT} roadmaps Lazy-CPBS tends to achieve the best success rates overall. The advantage is most pronounced in smaller or more cluttered environments with tight bottlenecks (e.g., \texttt{maze} and \texttt{random}), where Lazy-CPBS clearly outperforms PP and CPBS typically follows closely until the highest agent counts. On larger and less constrained maps (e.g., \texttt{warehouse}, \texttt{Berlin}, and \texttt{empty256}), the gap narrows: the additional free space and connectivity delay the onset of congestion, so the suboptimal planners scale more smoothly and the robustness advantage of priority revision, while still visible, becomes less pronounced.

A similar but reversed picture emerges on \texttt{700~CDT}. Here PP benefits systematically from the larger roadmap, which reduces effective congestion and preserves feasibility for longer. In the largest and least bottlenecked environments, this translates into scalability to hundreds of agents, whereas CPBS and Lazy-CPBS typically become limited earlier as the cost of managing interactions grows and the runtime limit becomes the dominant failure mode. Lazy-CPBS nevertheless reaches 1.3--1.6$\times$ as many agents as CPBS before this limit.

\subsection{Overall Findings}
Across all environments, the observed behaviors are largely density-driven, where density refers to the effective congestion induced by agents competing for the same free space and, especially, the same bottlenecks. When instances become dense quickly (e.g., narrow corridors and tight passages), PP's fixed-order policy is brittle and often collapses early due to infeasibility, whereas priority-revising methods (CPBS and Lazy-CPBS) remain effective over a wider range of agent counts. Conversely, in environments that remain relatively sparse as $n$ grows (e.g., open layouts such as \texttt{empty256}), all suboptimal solvers retain high success for longer ranges and PP benefits the most from its low computational overhead, typically dominating runtime by a wide margin; CPBS/Lazy-CPBS become mainly beneficial near the highest densities, where revising priorities mitigates the fragility of fixed ordering.

Finally, it is important to note that these instances originate from discrete-time MAPF benchmarks, where scaling to hundreds or even thousands of agents is often feasible. In the continuous-time roadmap setting, feasibility and runtime are shaped by interaction handling: potentially conflicting motions require continuous-time collision checking and safe-interval/time-constraint updates, and planning occurs on geometric roadmaps rather than grids. As a result, once congestion emerges, computational overhead increases sharply and the practical ``hardness threshold'' is reached at substantially smaller agent counts than in the discrete domain.

\section{Industrial Evaluation on AGV-MAPF$_R$-CTC}
\label{sec:industrial}

After the benchmark evaluation on standard instances from the MAPF literature, we assess the proposed methods in a realistic industrial setting, in collaboration with E80 Group S.p.A., a leading international provider of automation solutions for logistics. As discussed in Section~\ref{subsec:agv-mapfr-ctc}, solvers tailored for MAPF$_R$-CTC can be deployed for the management and control of AGV fleets with minimal algorithmic changes: the key adaptation is to interpret the underlying roadmap as a discretization of the vehicles' configuration space.

\subsection{Simulation Setup}
In this experimental campaign, all algorithms are implemented in C\# within a custom MAPF evaluation framework developed in-house. The framework is interfaced with the company's proprietary simulator only to access the digital twin of deployed AGV systems and to import the industrial roadmaps $\overrightarrow{\mathcal{R}}$ used in real customer plants, together with the corresponding offline conflict annotations $\mathcal{C}_{\mathrm{VEC}}$ and $\mathcal{C}_{\mathrm{EEC}}$. Heuristic preprocessing, instance generation, continuous-time conflict handling, and multi-agent planning are performed outside the proprietary simulator, within our own evaluation pipeline. Experiments are executed on a notebook equipped with an Intel Core i7-12700H CPU (2.30~GHz, 14 cores, 20 threads) and 16~GB DDR4 RAM.

\begin{figure}
    \centering
    \includegraphics[width=0.48\textwidth]{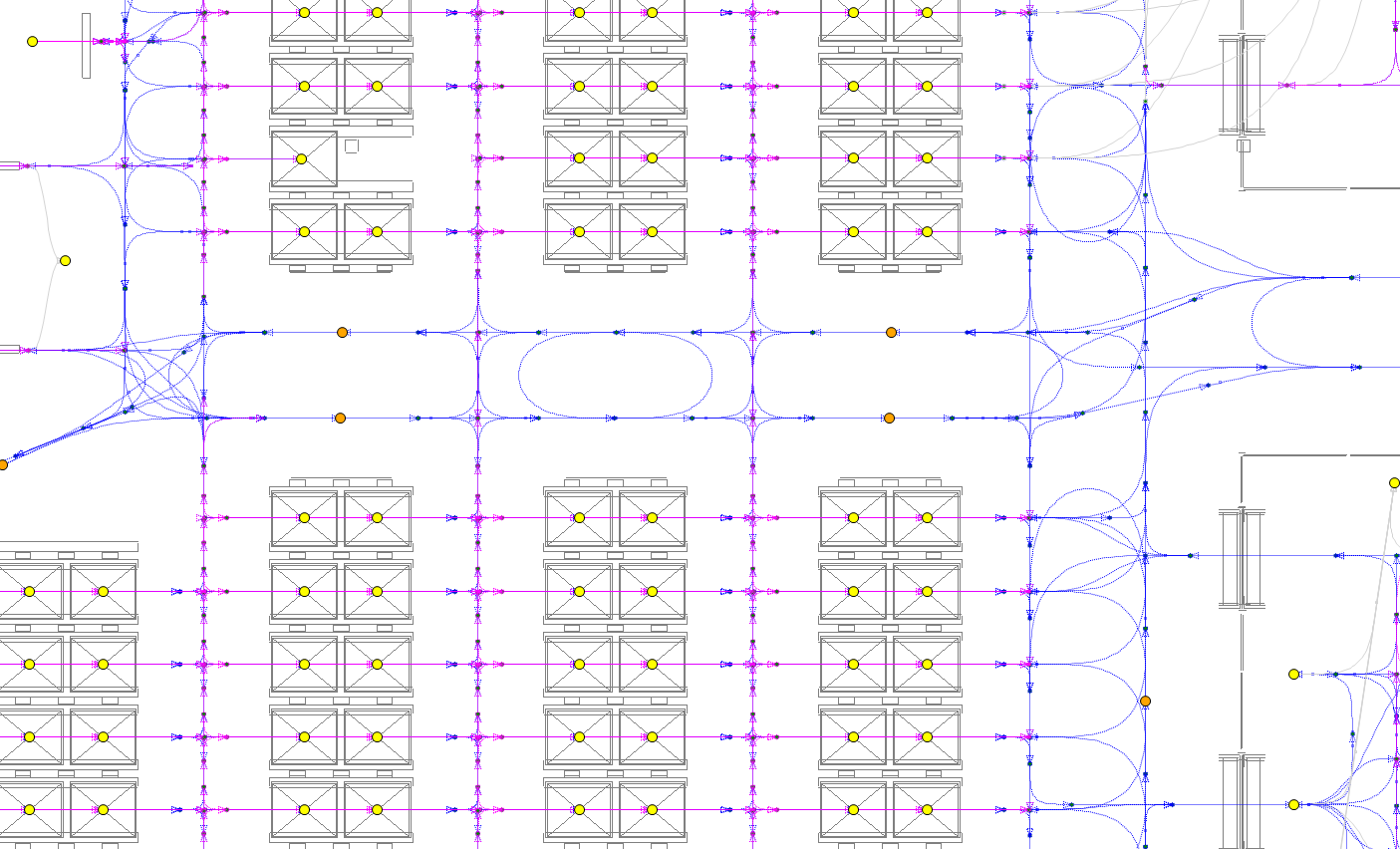}
    \caption{Portion of a roadmap in a real industrial plant (extracted from layout A).}
    \label{fig:roadmap_excerpt}
\end{figure}

We evaluate on five industrial warehouse layouts derived from real-world installations (Layouts A-E). Table~\ref{tab:e80_layouts} summarizes the corresponding roadmap statistics, reporting the number of vertices and edges, the average and maximum branching factor (i.e., out-degree), and the number of AGVs operating in each environment.

Layout A is compact, with redundant one-way lanes. Layout B is a regular grid of long parallel aisles lined with dead-end bays. Layout C is compact but dense: long single-lane aisles traversed in both directions, with perpendicular dead-end bays, so that the aisles act as bottlenecks.
Layout D is small and sparsely connected, with two clusters joined through a narrow junction.
Layout E has a comb-like topology, with two parallel spine corridors from which many dead-end bays branch on both sides. Fig.~\ref{fig:roadmap_excerpt} presents an excerpt of the roadmap $\overrightarrow{\mathcal{R}}$ extracted from layout A. For confidentiality reasons, the complete layouts and the industrial implementation cannot be disclosed.

\begin{table}[t]
\centering
\caption{Characteristics of the industrial layouts.}
\label{tab:e80_layouts}
\resizebox{\ifdim\width>\columnwidth\columnwidth\else\width\fi}{!}{%
\begin{tabular}{lrrrr}
\toprule
Layout & $|V|$ & $|E|$ & $b$ ($b_{\max}$) & Installed AGVs \\
\midrule
A & $\approx 1400$  & $\approx 3400$  & $\approx 2.3$ $(8)$ & $20$ \\
B & $\approx 8700$  & $\approx 19200$ & $\approx 2.1$ $(7)$ & $21$ \\
C & $\approx 1700$  & $\approx 4300$  & $\approx 2.5$ $(6)$ & $5$ \\
D & $\approx 250$   & $\approx 400$   & $\approx 1.5$ $(7)$ & $2$  \\
E & $\approx 1200$  & $\approx 2100$  & $\approx 1.7$ $(8)$ & $5$ \\
\bottomrule
\end{tabular}}
\end{table}

For each layout, we generate benchmark instances by sampling distinct start-goal pairs on the roadmap and progressively increasing the fleet size. For each agent count $n$, we run 25 random instances per layout and impose a per-run runtime cap of 30\,s; an instance is marked as failed if no solution is found within the limit.
Industrial roadmaps may become interaction-dense already at relatively small fleet sizes because of bottlenecks, narrow passages, and geometric constraints. In this regime, fixed-order prioritized planning can be brittle, while priority-revising methods are expected to remain effective deeper into congestion.

Motivated by this density-driven behavior, we compare:
(i) CPBS and Lazy-CPBS;
(ii) CCBS augmented with CTC annotations (CCBS-CTC, referred to as CCBS for brevity); and
(iii) prioritized planning baselines based on PSIPP-CTC (hereafter PP) and two restart enhancements:
PP+RR (random restart) and PP+DR (deterministic restart).
Random restart repeatedly samples new random priority orderings as long as runtime allows, whereas deterministic restart promotes the agent that fails to find a path to highest priority in subsequent runs.

\subsection{Results and Discussion}
\begin{figure}[t]
\centering

\subfloat[Layout A]{
\includegraphics[width=0.32\linewidth]{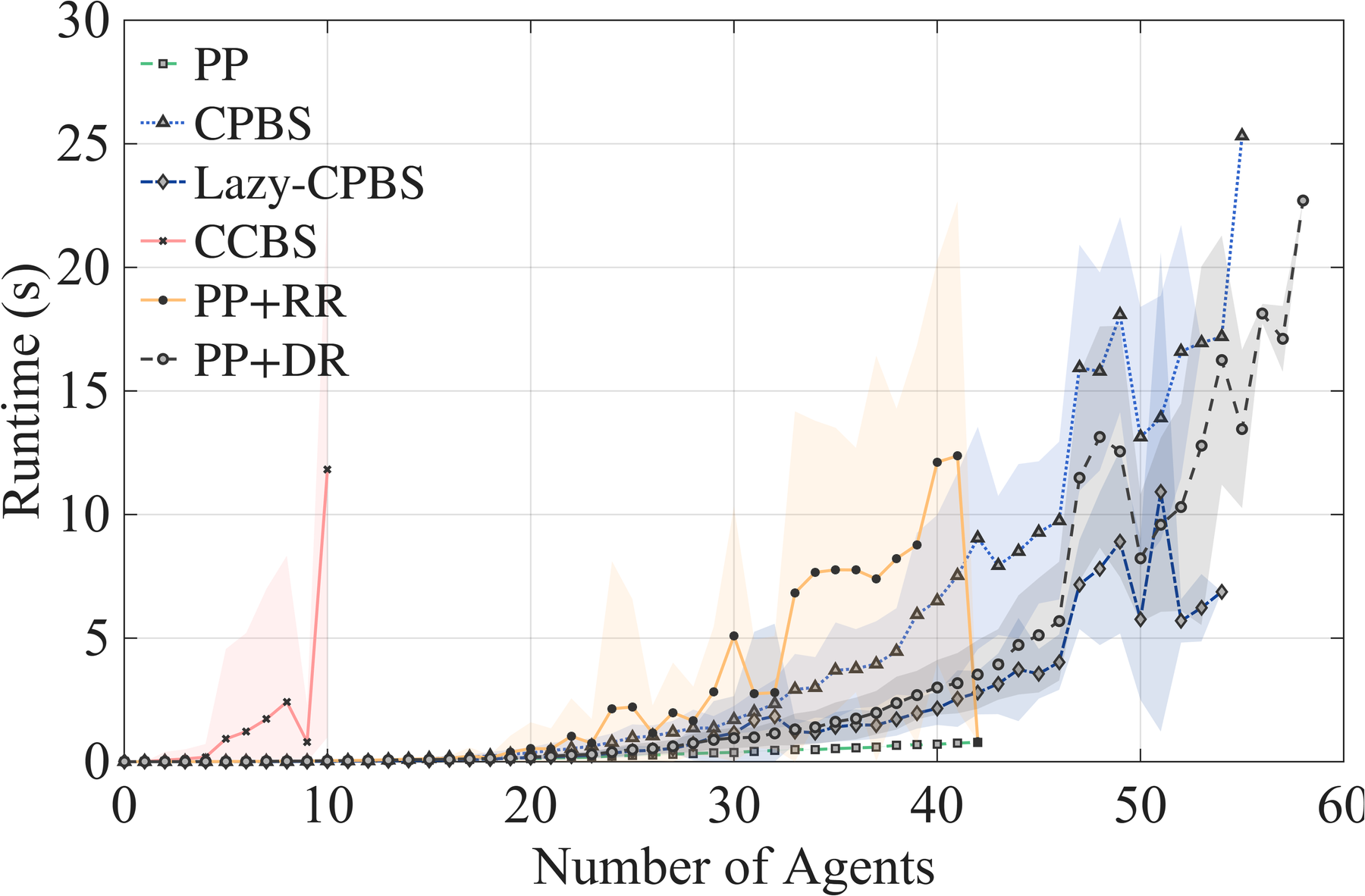}
\includegraphics[width=0.32\linewidth]{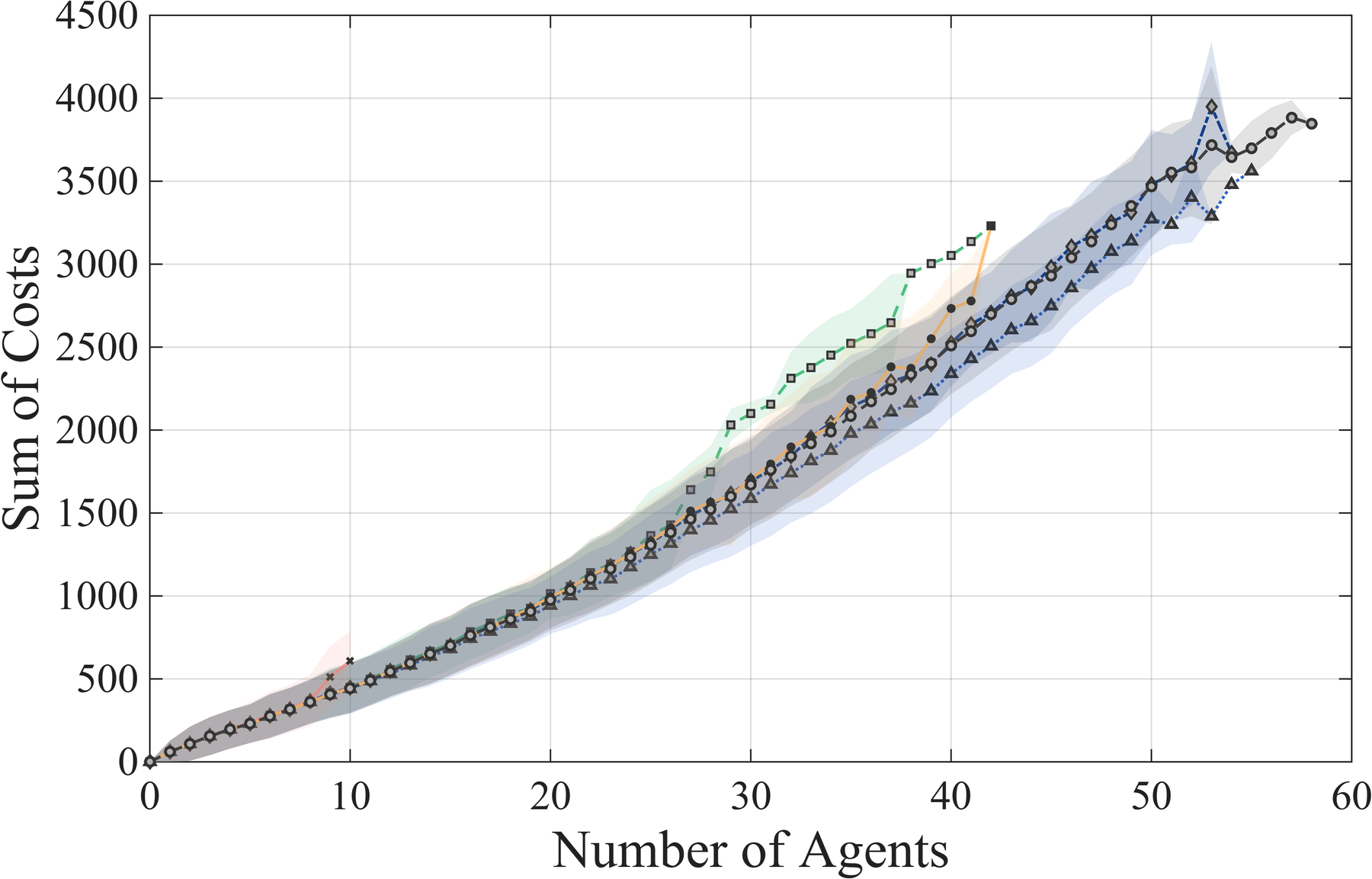}
\includegraphics[width=0.32\linewidth]{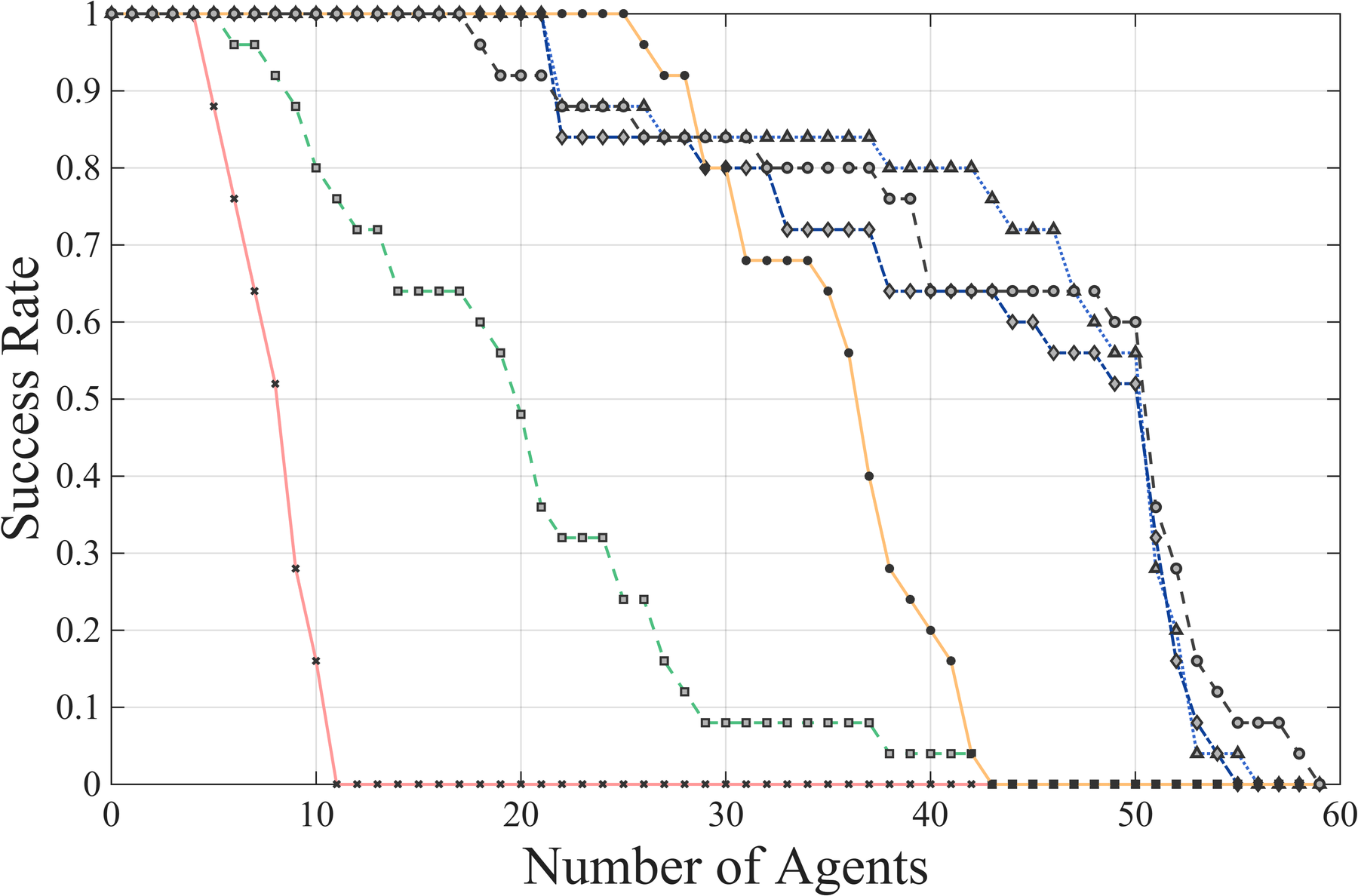}
} \\

\subfloat[Layout B]{
\includegraphics[width=0.32\linewidth]{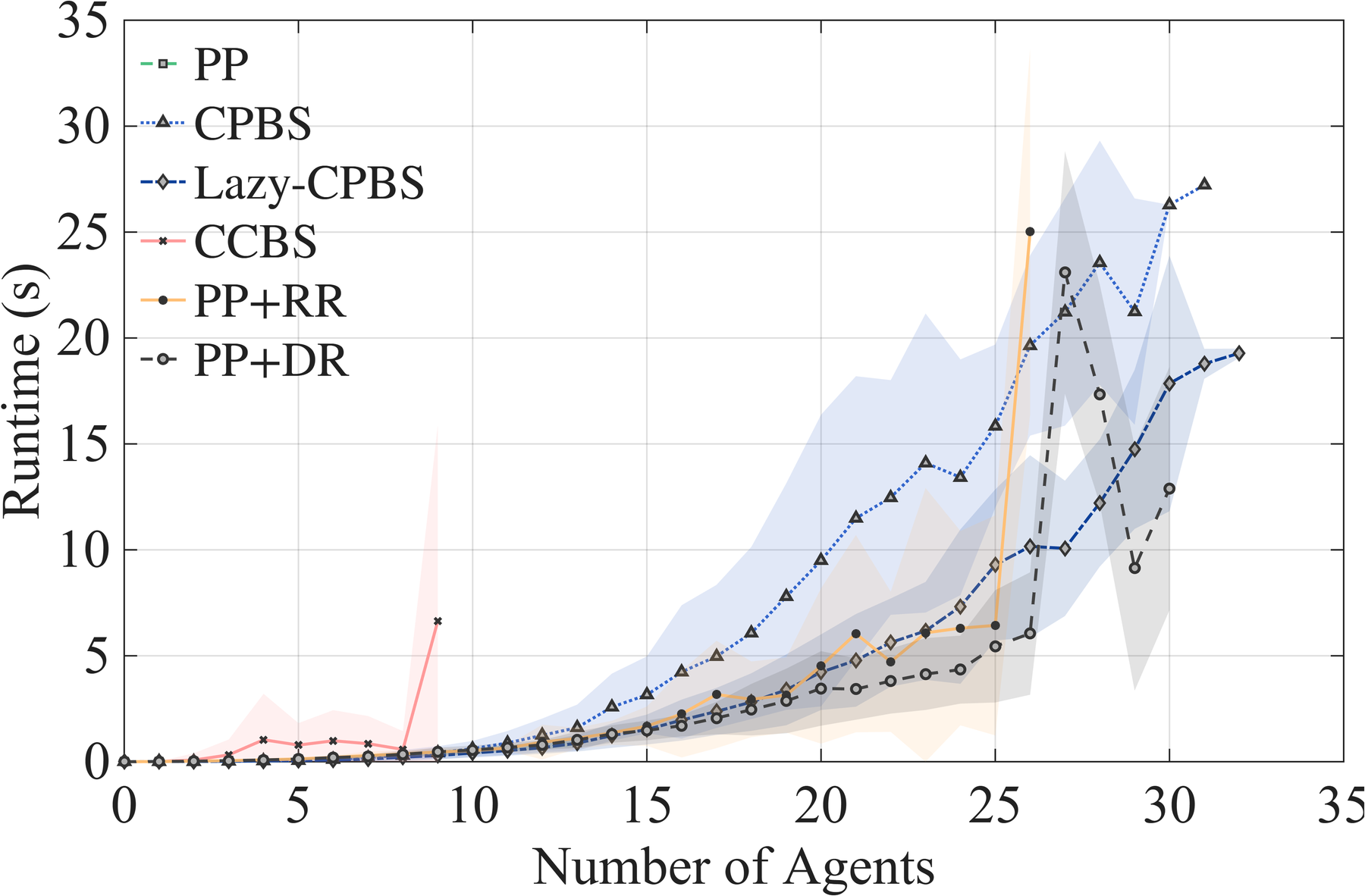}
\includegraphics[width=0.32\linewidth]{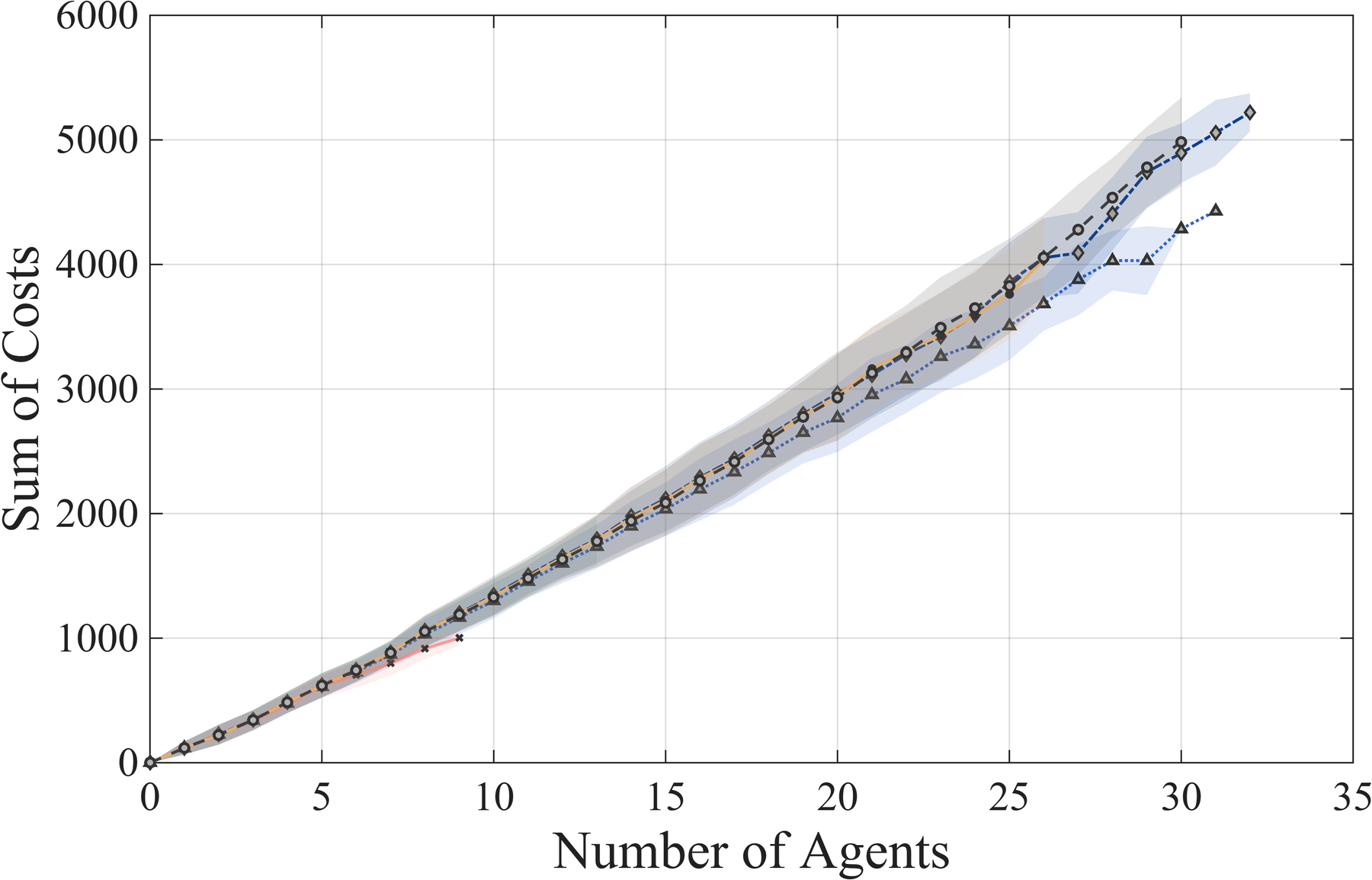}
\includegraphics[width=0.32\linewidth]{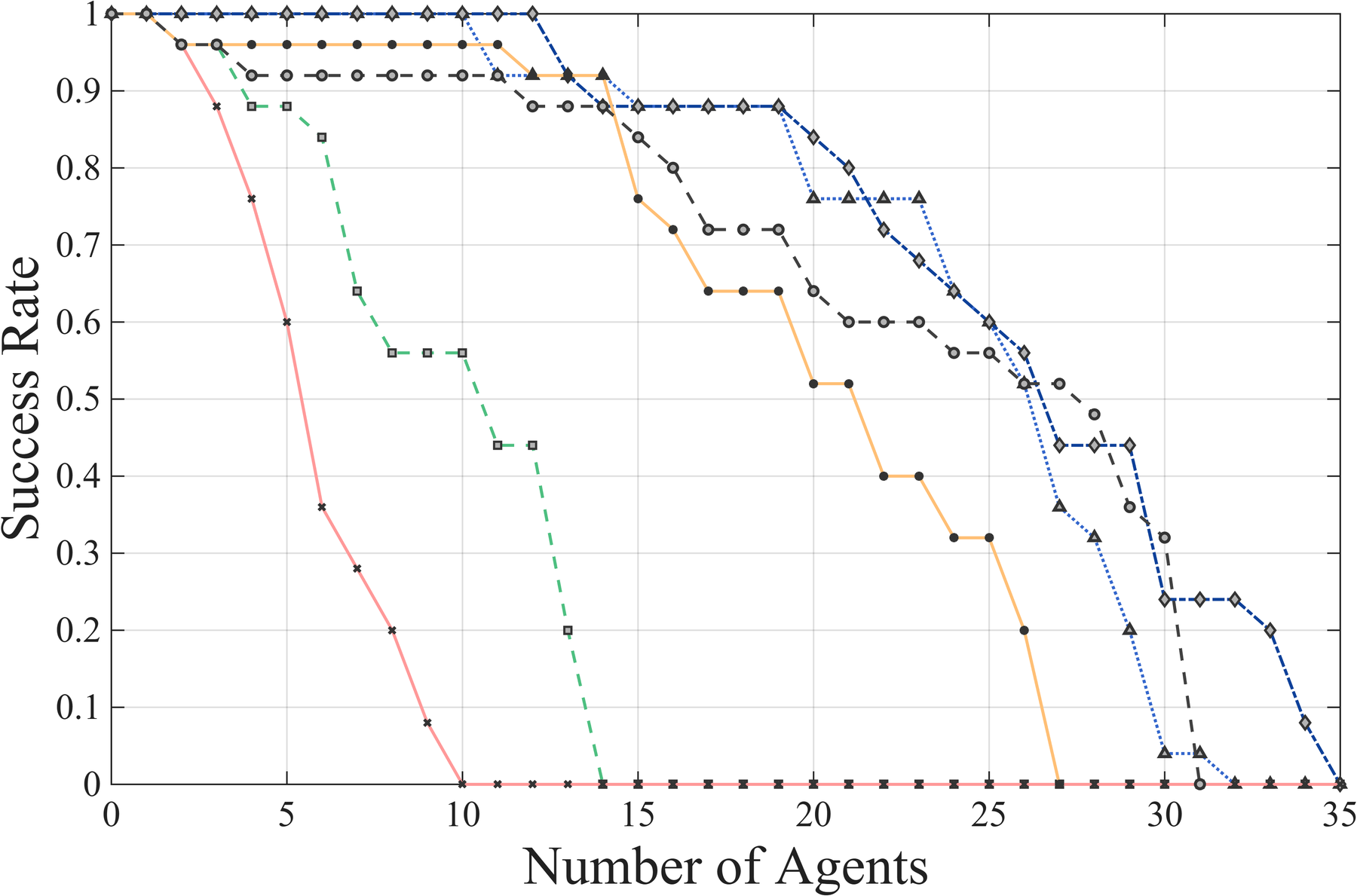}
} \\

\subfloat[Layout C]{
\includegraphics[width=0.32\linewidth]{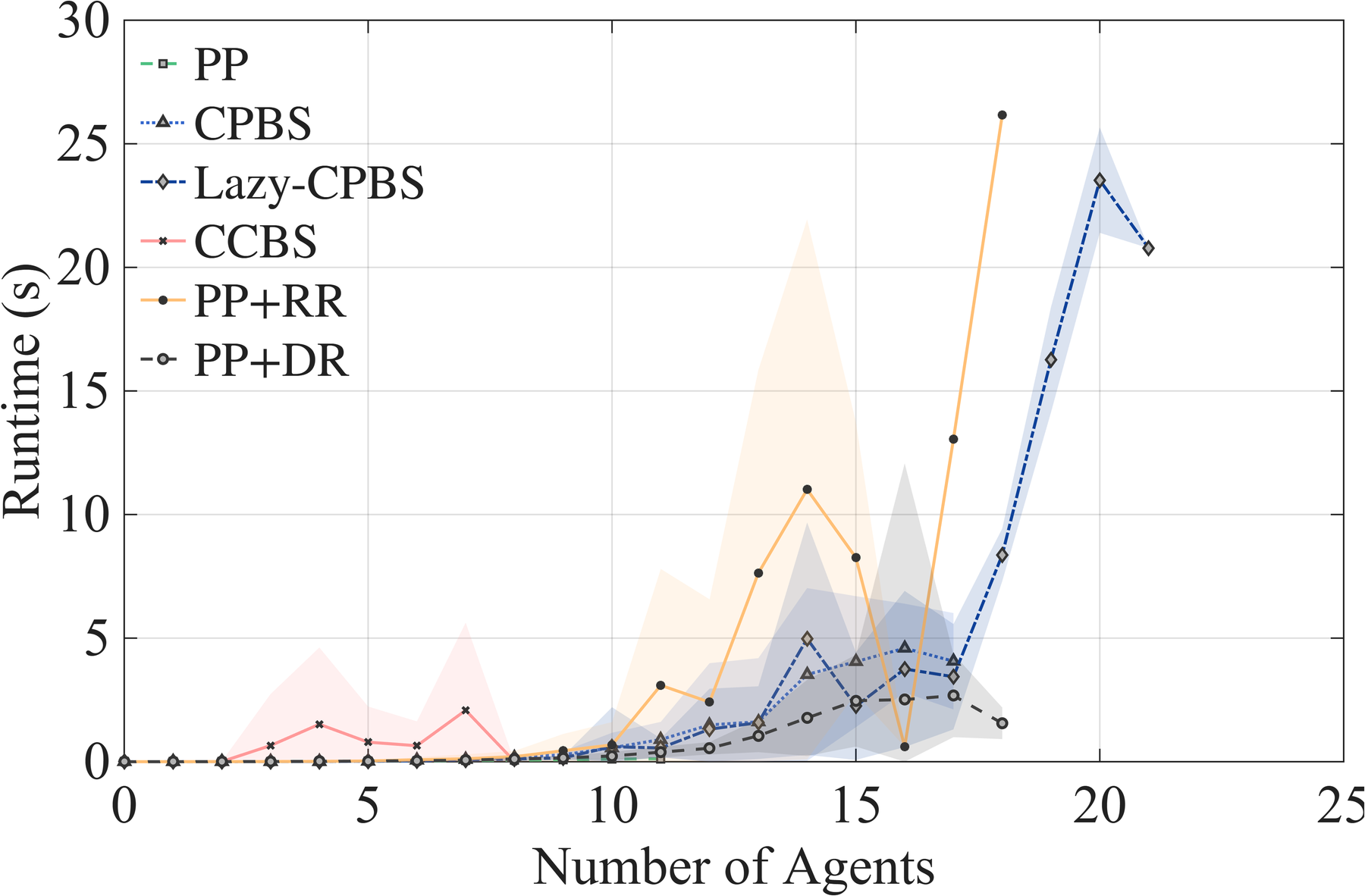}
\includegraphics[width=0.32\linewidth]{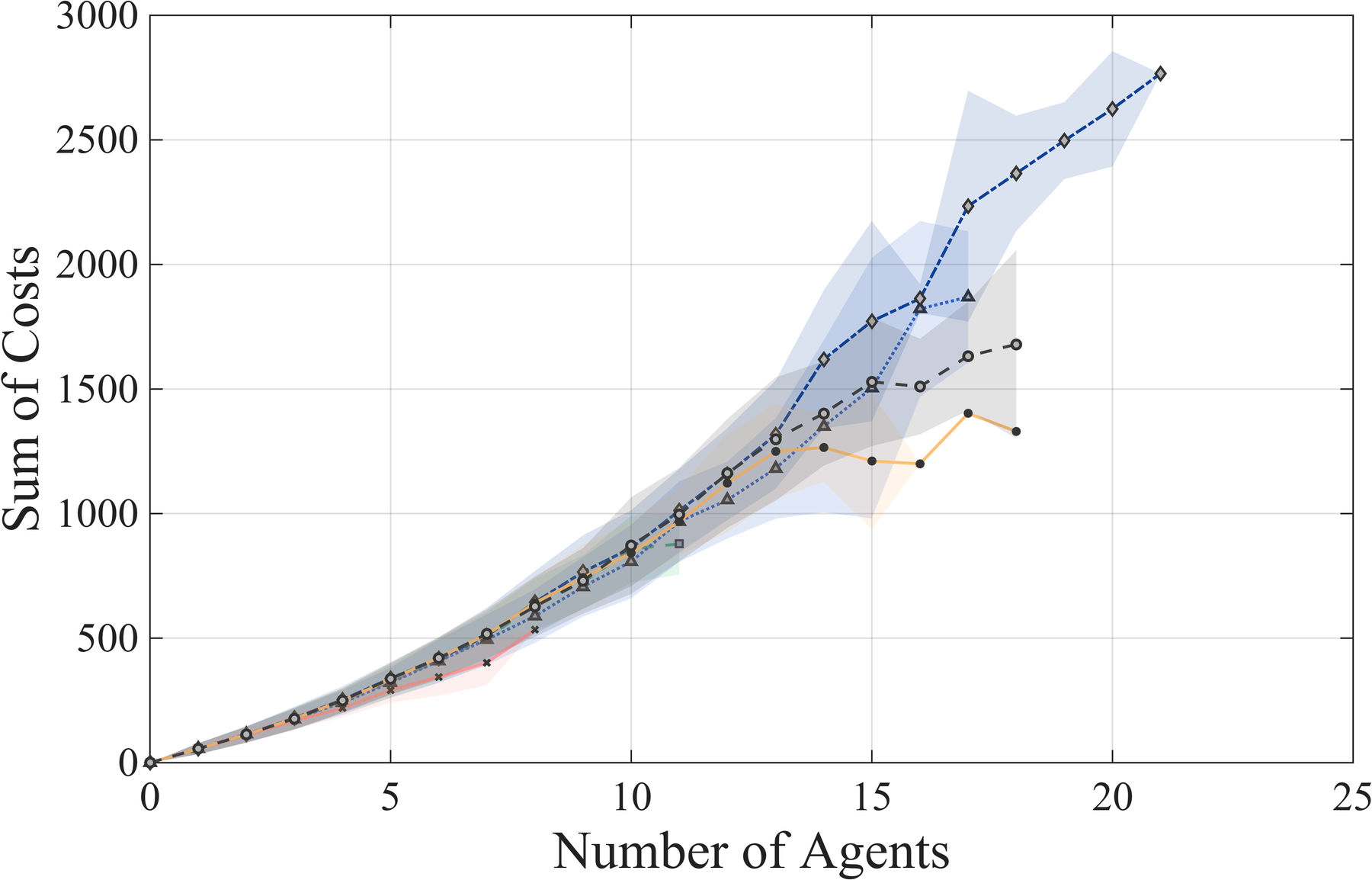}
\includegraphics[width=0.32\linewidth]{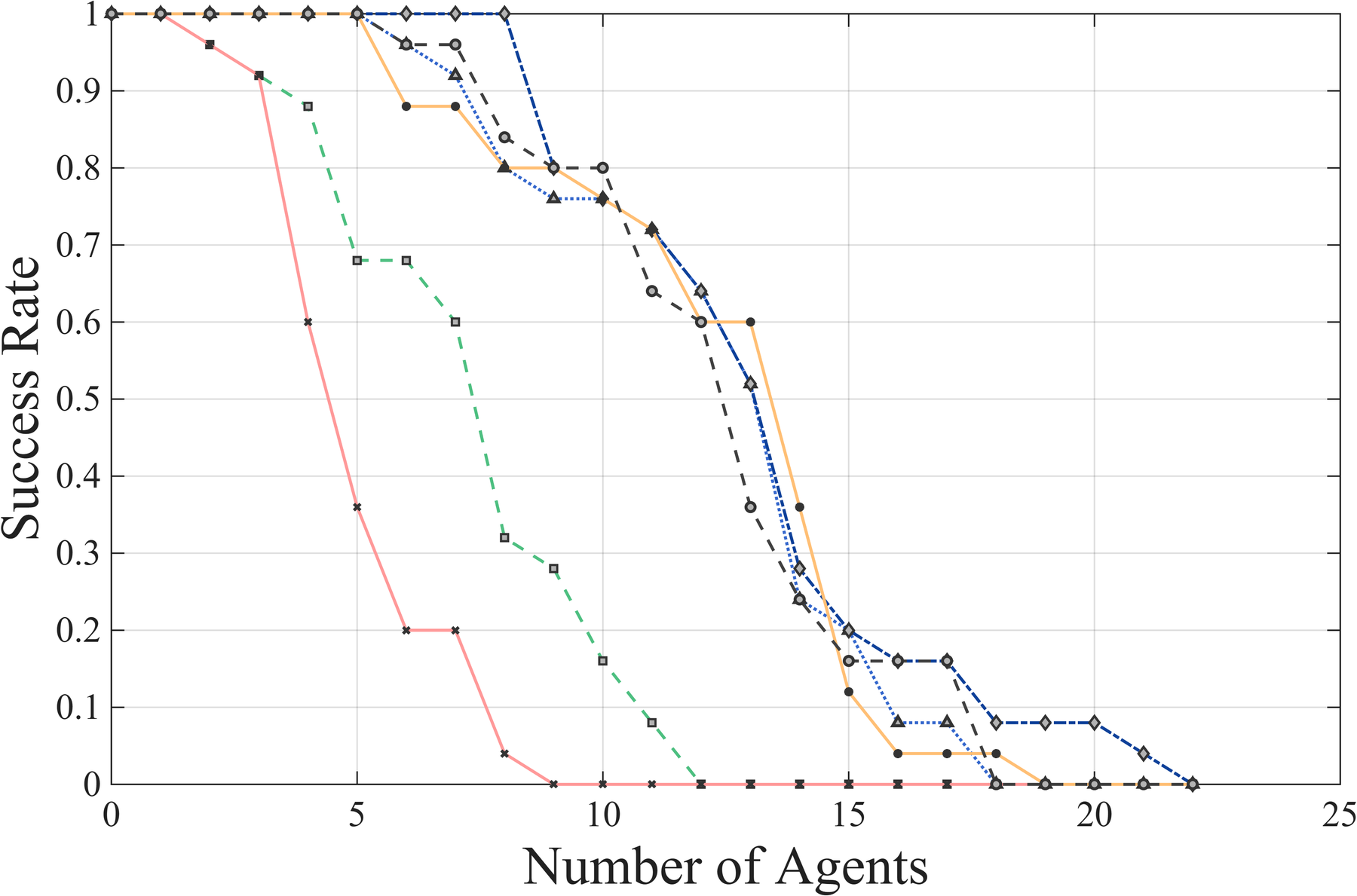}
} \\

\subfloat[Layout D]{
\includegraphics[width=0.32\linewidth]{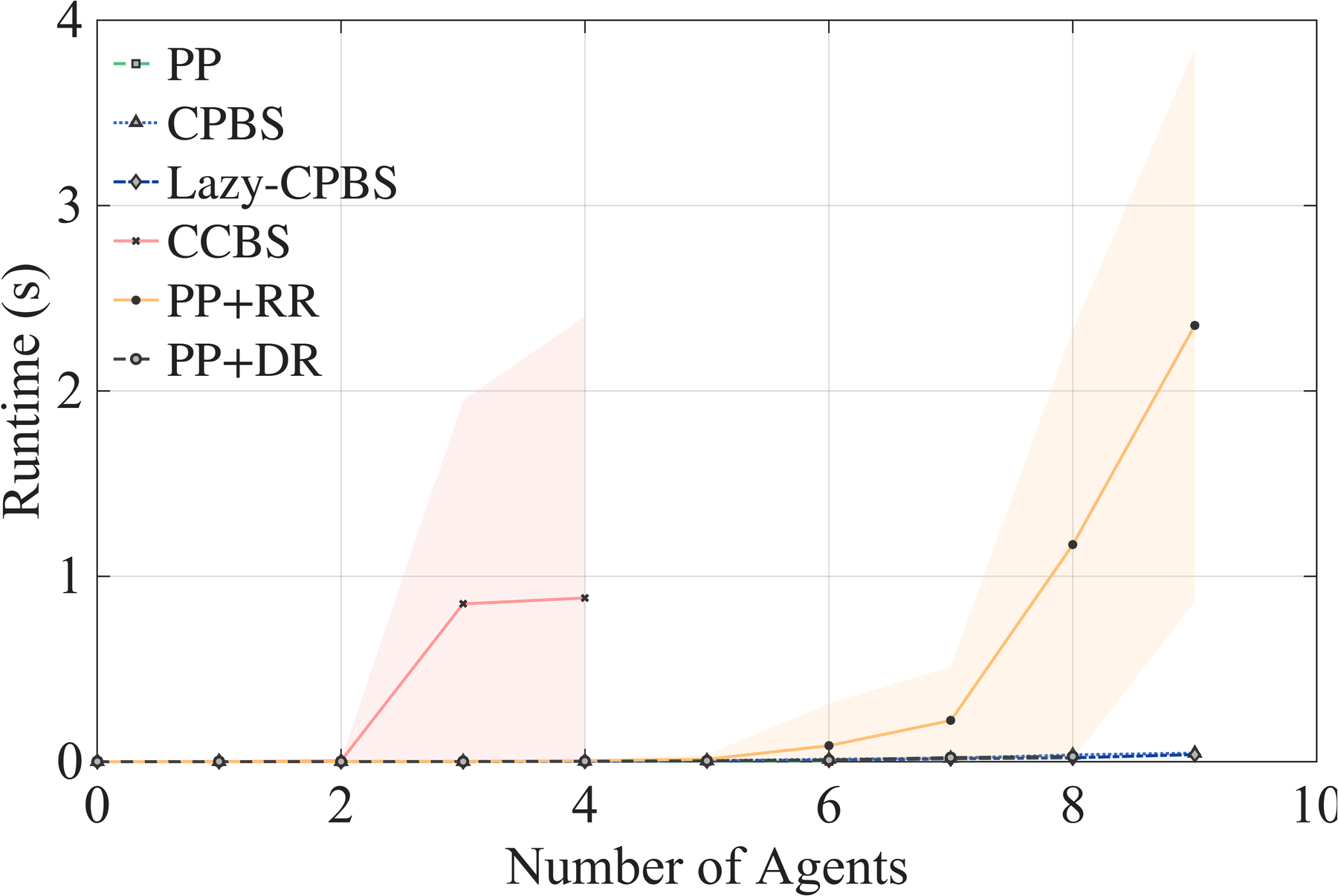}
\includegraphics[width=0.32\linewidth]{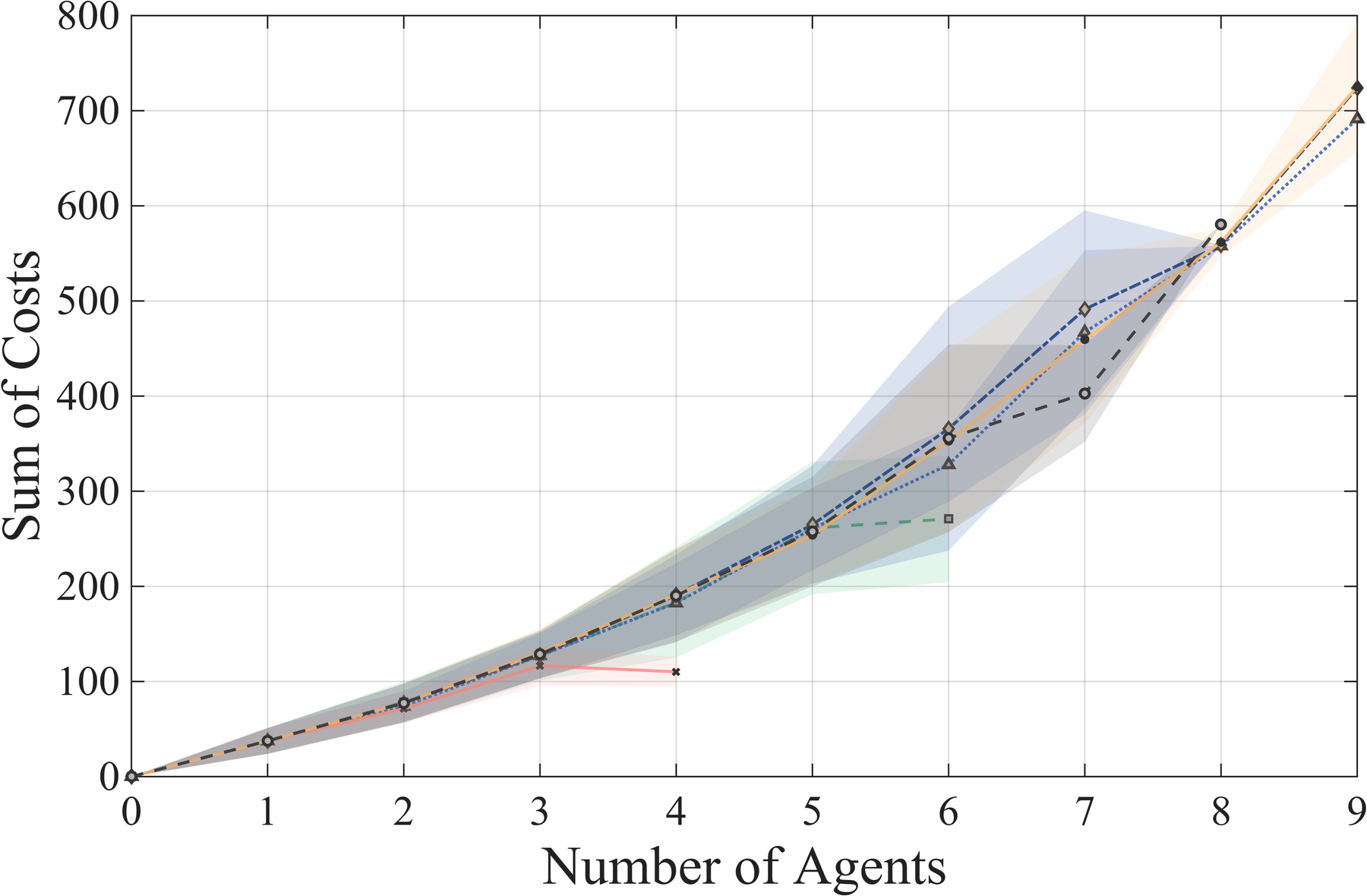}
\includegraphics[width=0.32\linewidth]{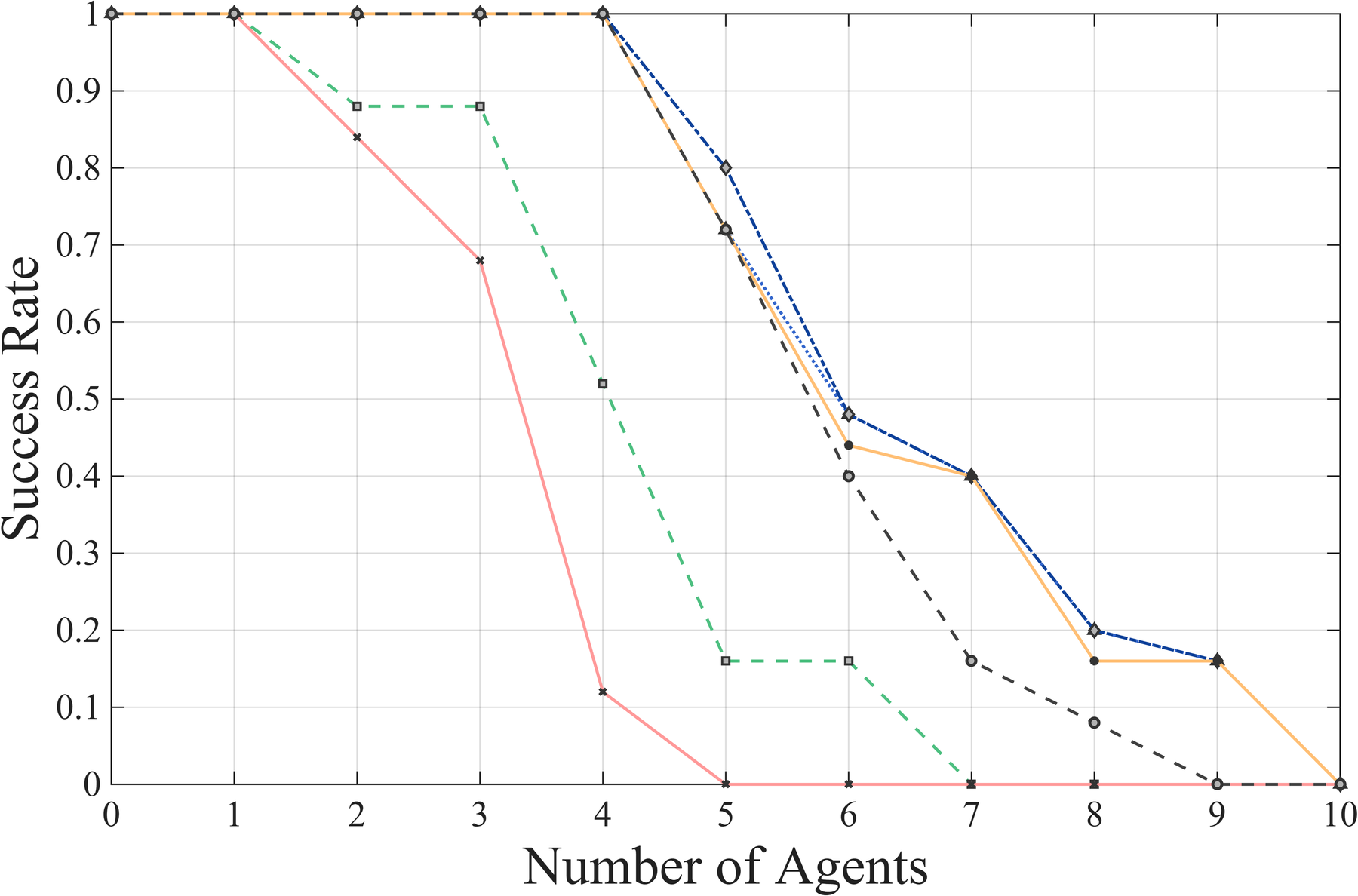}
} \\

\subfloat[Layout E]{
\includegraphics[width=0.32\linewidth]{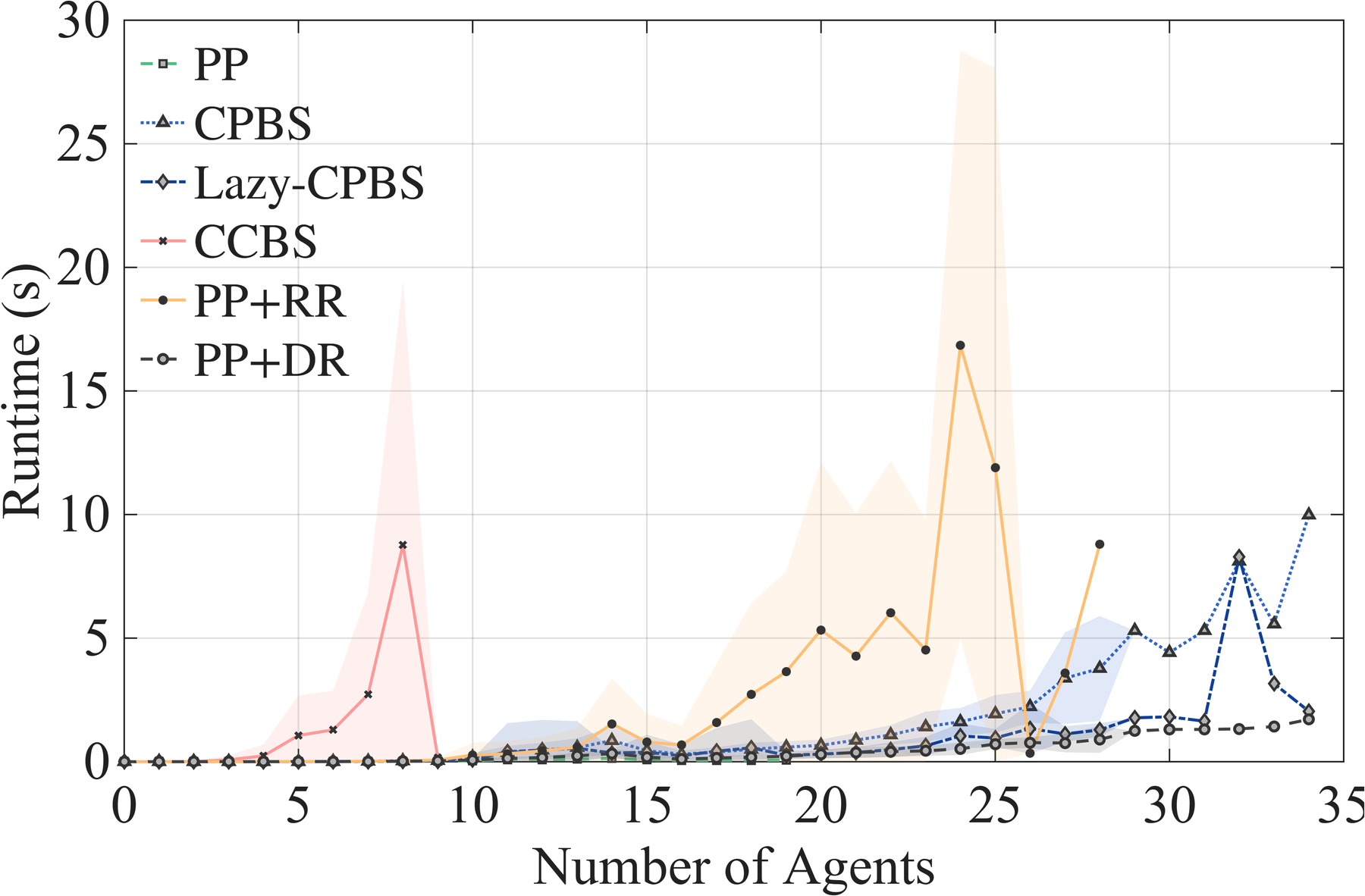}
\includegraphics[width=0.32\linewidth]{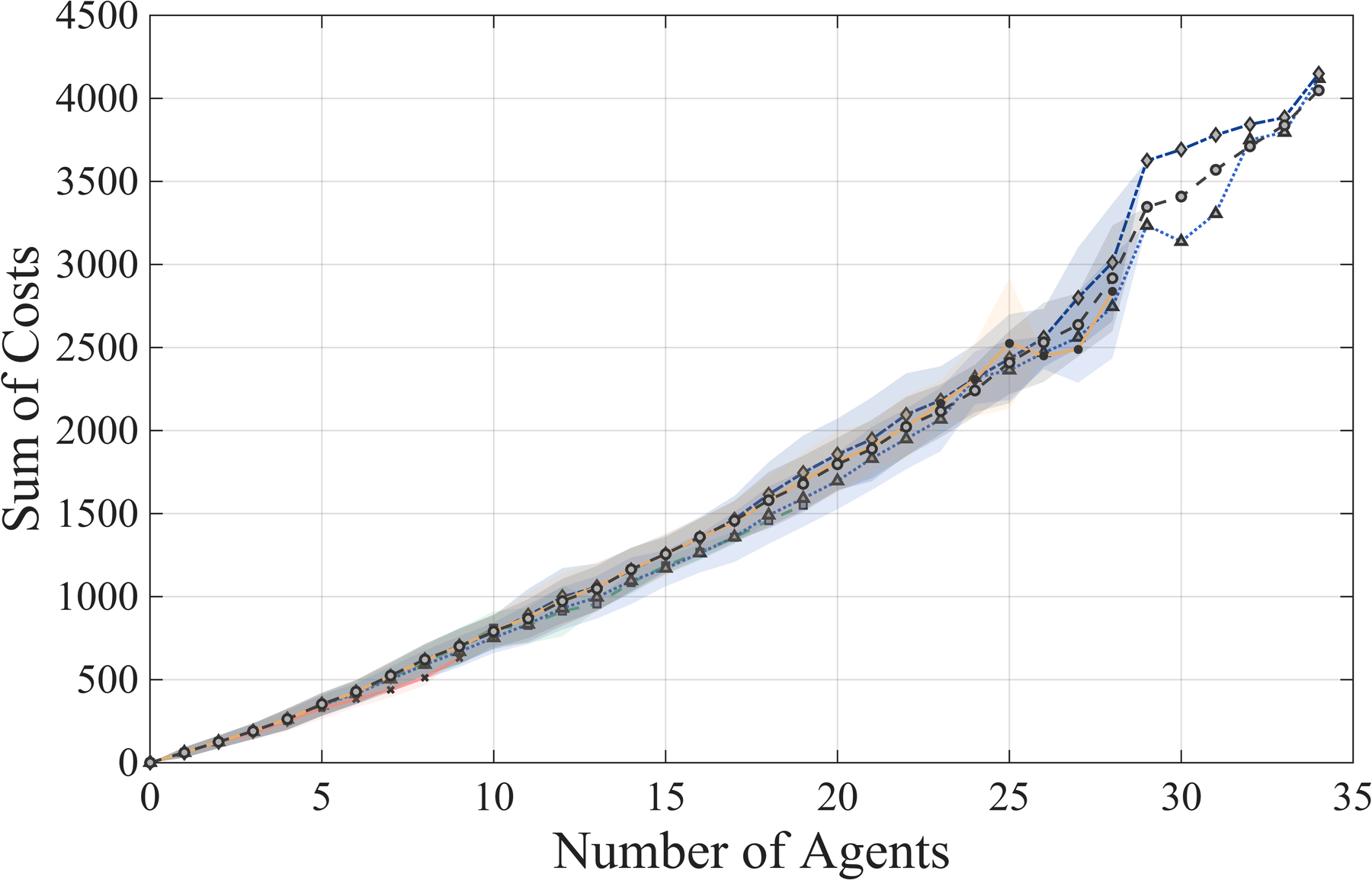}
\includegraphics[width=0.32\linewidth]{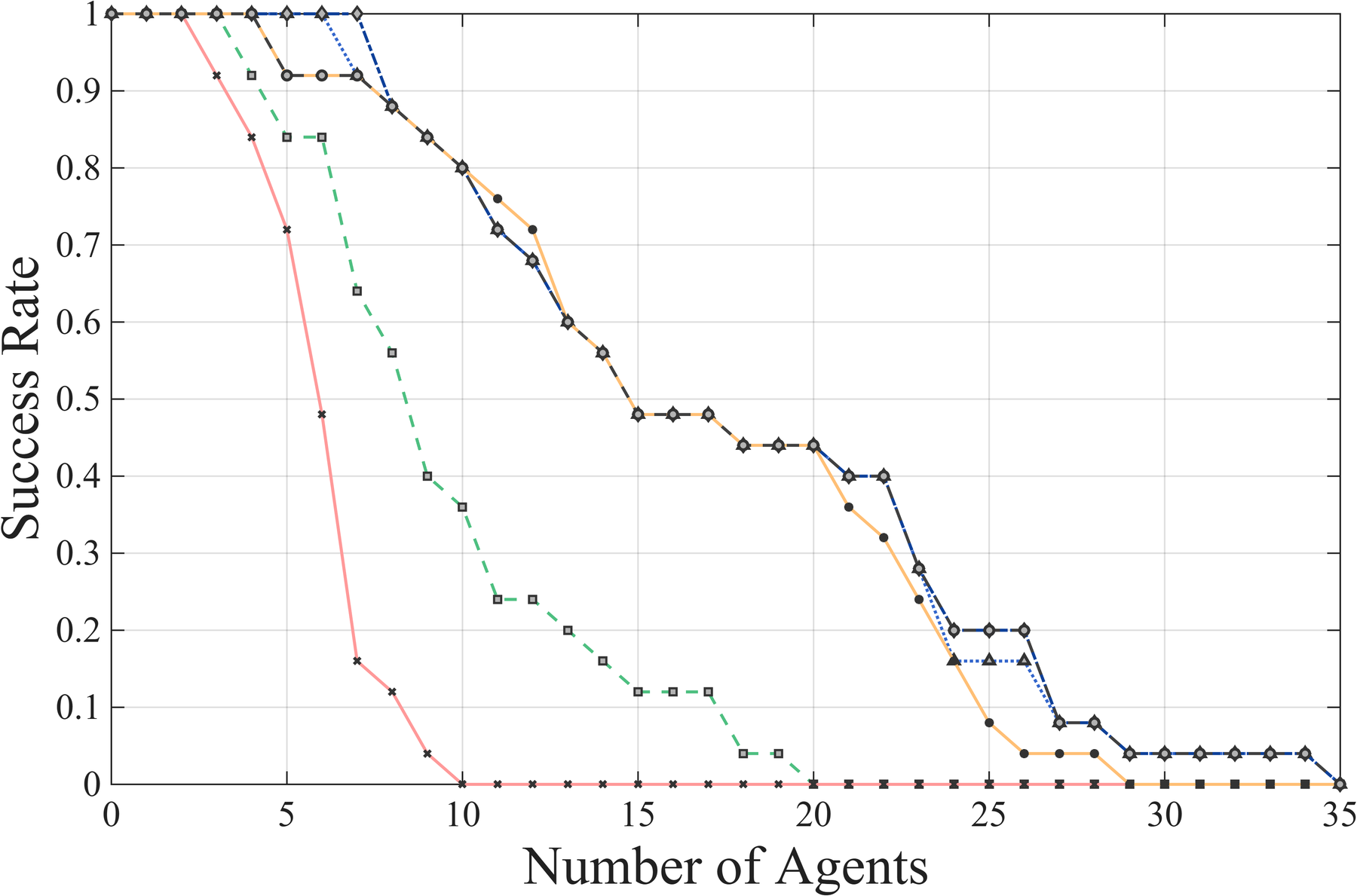}
} \\

\caption{Experimental results across five warehouse layouts. Each row corresponds to a different layout, showing runtime (left), SoC (middle), and success rate (right). Lines show the mean over solved instances and shaded bands $\pm 1$ standard deviation. \protect\footnotemark}
\label{fig:industrial-results}
\end{figure}

The results of the simulations on the five layouts are reported in Fig.~\ref{fig:industrial-results}. For each layout and algorithm, we analyze runtime, solution SoC, and success rate as the number of agents increases, with particular attention to the operating fleet sizes reported in Table~\ref{tab:e80_layouts}.

\footnotetext{Sample animations of instances solved by Lazy-CPBS are available at:
\url{https://www.youtube.com/watch?v=5NBPDKT-ud8}
(industrial animations displayed at 4$\times$ real-time speed).}

The first observation concerns CCBS.
Consistently with the benchmark-domain experiments, CCBS provides the lowest SoC when it succeeds, but its scalability remains limited in the industrial AGV-MAPF$_R$-CTC setting.
Its success rate degrades after only a few agents and vanishes by about ten agents on every layout.
As a result, CCBS does not reliably reach the operating fleet size on any layout: it solves no instance at the operating points of layouts~A and~B, and only a fraction of them on layouts~C, D, and~E.
This behavior is expected for CBS-style optimal solvers: resolving conflicts through a constraint tree becomes increasingly expensive on redundant roadmaps, where many alternative low-level paths traverse overlapping regions and generate additional interactions.
Thus, although CCBS remains the strongest method in terms of solution quality, it is not a viable option at the operating points of any of the considered layouts.

The second observation concerns prioritized planning.
PP is consistently the fastest method when it succeeds, but it also exhibits one of the lowest success rates across the industrial layouts.
Its fixed priority ordering makes it fragile under congestion: once the initial ordering is incompatible with the interaction structure of the instance, PP has no mechanism to revise it.
This limitation is more pronounced than in several standard MAPF benchmark settings, because industrial AGV roadmaps contain bottlenecks, shared passages, and larger effective vehicle footprints, so interactions arise already at moderate fleet sizes. PP never achieves full success at the operating points: it solves only a subset of instances on layouts~A, C, D, and~E, and none on layout~B, so its low runtime does not compensate for its limited reliability.

The restart-based variants PP+RR and PP+DR substantially improve over vanilla PP.
By testing multiple priority orderings, they recover feasible solutions in cases where a single fixed ordering fails, and around the operating points they match CPBS-based methods on layouts~C and~D, miss at most two instances on layouts~A and~E, and fall well behind on layout B.
Neither variant dominates the other: PP+RR is more reliable at intermediate densities on layout~A, whereas PP+DR degrades more gracefully as density grows and scales further on layouts~A and~B.
However, both methods remain weakly informed: they explore complete priority orderings without maintaining a structured representation of partial priority decisions or of previously failed priority prefixes.
As a consequence, their performance is more variable across layouts and density regimes.
In dense instances, where only a small fraction of priority orderings may be feasible, restart strategies can spend significant runtime testing orderings that do not lead to valid plans, whereas CPBS-based methods exploit partial priority constraints to guide the search more systematically.
Thus, although PP+RR and PP+DR are effective practical improvements over PP, their lower predictability near the operating points limits their suitability as standalone strategies for industrial AGV fleet coordination.

The strongest overall results are obtained by the CPBS-based methods.
Both CPBS and Lazy-CPBS maintain partial priority relations and revise priorities only when conflicts arise, which makes them substantially more robust than PP and more consistent than its restart variants across the congested regimes typical of industrial AGV roadmaps.
Lazy-CPBS further avoids materializing children that are never expanded. On layouts~A, B, and~E, this translates into markedly lower runtimes than CPBS, and on layouts~B and~C it allows Lazy-CPBS to solve instances with more agents than any other method.

The comparison across layouts is, however, more nuanced than a uniform ranking.
On layout~A, PP+RR is the most reliable method up to about 28 agents, CPBS from about 30 to 46 agents, and PP+DR at the highest densities; Lazy-CPBS stays close to CPBS up to 28 agents and is less robust beyond, but it is the fastest of the priority-revising methods.
On layout~B, Lazy-CPBS has the best success profile over most of the range and is the only method solving instances beyond 31 agents, with CPBS and PP+DR marginally ahead on some instances.
On layout~C, Lazy-CPBS remains fully reliable up to 8 agents, when all other methods have started to fail, and scales the furthest; at intermediate densities, the priority-based methods perform similarly.
On layout~D, CPBS, Lazy-CPBS, and PP+RR are nearly indistinguishable in success rate, with Lazy-CPBS slightly ahead at 5 agents, whereas PP+DR degrades earlier; however, PP+RR pays for its reliability with a sharp runtime increase, up to two orders of magnitude above the other priority-based methods, which all run in a fraction of a second.
On layout~E, Lazy-CPBS, CPBS, and PP+DR largely overlap, with Lazy-CPBS remaining fully reliable up to 7 agents and CPBS incurring higher runtimes at high densities.
Thus, Lazy-CPBS is not the best method on every layout and density, but it is the most consistent: every competitor falls clearly behind on at least one layout, namely CPBS in runtime, PP+RR through its collapse at higher densities on layouts~A and~B, and PP+DR through its lower reliability near the operating points of layouts~A and~B.

The advantage of Lazy-CPBS is particularly clear around the operating fleet sizes.
In this regime, Lazy-CPBS solves all instances on layouts~A, C, D, and~E, and about 80\% on layout~B, the highest rate among all methods there.
CPBS matches it on the first four layouts but is less reliable on layout~B, at more than twice the runtime, whereas the restart variants miss instances on layouts~B and~E, and PP+DR also on layout~A.
This makes Lazy-CPBS preferable for the considered industrial setting: it preserves the robustness of systematic priority revision while substantially reducing the replanning overhead that limits eager CPBS.

Finally, the SoC results follow the expected trade-off between optimality and greediness.
When several methods solve the same instance, CCBS returns the lowest-cost solution, typically followed by CPBS, Lazy-CPBS, and the prioritized-planning baselines.
The restart-based variants produce solutions of comparable quality to Lazy-CPBS, but with less predictable success.

\subsection{Overall Findings}
The industrial evaluation identifies CPBS-based planners as the most promising methods among the evaluated alternatives for realistic AGV roadmap coordination. In particular, Lazy-CPBS provides the strongest overall robustness/runtime compromise: it remains reliable under congestion and bottlenecks, where fixed or weakly informed priority strategies are less consistent across layouts, while substantially reducing the replanning overhead of CPBS and avoiding the scalability limitations of CCBS.

At the same time, the current implementation does not yet meet the strictest industrial requirements for real-time coordination of very large AGV fleets. However, CPBS and Lazy-CPBS provide concrete directions for further acceleration because they are grounded in a principled, systematic search structure. In particular, the structural properties discussed in Section~\ref{subsec:cpbs_properties} indicate that different regions of the priority-search space can be explored independently, suggesting that parallel high-level implementations could reach feasible priority structures earlier. This stands in contrast to heuristic prioritized-planning enhancements such as PP+RR and PP+DR, which may occasionally yield strong performance but tend to be less predictable and less consistent across layouts and traffic regimes.

\section{Ablation Study}
\label{sec:ablation}
Lazy-CPBS differs from CPBS in two respects: when it materializes nodes, and how it orders children, by $|\mathrm{ReplanList}|$ instead of SoC, which is unknown before materialization.
To isolate the contribution of each to the speed-up of Lazy-CPBS over CPBS, in this section we present an ablation study on the \texttt{100 CDT} benchmark roadmaps of Section~\ref{sec:experiments} and on the industrial layouts of Section~\ref{sec:industrial}.

\subsection{Experimental Design}

We introduce an intermediate CPBS-based variant, \textit{Eager-RL}, which materializes
both children at every expansion as CPBS does, but orders them by
$|\mathrm{ReplanList}|$ as Lazy-CPBS does. Comparing Eager-RL with Lazy-CPBS isolates the effect of lazy materialization. To make this comparison exact, Eager-RL is run here without the early goal test (i.e., plans
are checked for conflicts at node expansion rather than at materialization): the two algorithms then expand the same sequence of nodes, so their runtime difference is due to lazy materialization alone. Comparing CPBS with Eager-RL, both with the early goal test, isolates instead the effect of the child ordering.

Table~\ref{tab:ablation} reports means over the instances solved by all
variants. For the first comparison, $d$ is the depth of the solution node,
$\Delta_{\mathrm{m}}$ the number of materializations saved by Lazy-CPBS,
$\bar t_{\mathrm{m}}$ the mean time of one eager materialization, and
$\rho$ the runtime of Lazy-CPBS relative to Eager-RL. The predicted saving
$\Delta t^{\mathrm{pred}}=\Delta_{\mathrm{m}}\,\bar t_{\mathrm{m}}$ is
compared with the measured one, $\Delta t^{\mathrm{meas}}$. For the second
comparison, the table reports the number of expanded nodes and of SIPP calls
under SoC (S) and $\mathrm{ReplanList}$ (R) ordering.

\begin{table}[t]
\centering
\caption{Ablation of Lazy-CPBS on the \texttt{100 CDT} benchmark roadmaps
(top) and on the industrial layouts (bottom). Times in ms.}
\label{tab:ablation}
\footnotesize
\setlength{\tabcolsep}{1.8pt}
\resizebox{\ifdim\width>\columnwidth\columnwidth\else\width\fi}{!}{%
\begin{tabular}{lrrrcrcc}
\toprule
 & \multicolumn{5}{c}{Lazy materialization} & \multicolumn{2}{c}{Child ordering}\\
\cmidrule(lr){2-6}\cmidrule(lr){7-8}
Env. & $d$ & $\Delta_{\mathrm{m}}$ & $\bar t_{\mathrm{m}}$ & $\Delta t$ pred/meas & $\rho$ & Exp.\ S/R & SIPP S/R\\
\midrule
\texttt{Berlin} & 86.8 & 86.7 & 9.1 & 792/1192 & 0.25 & 74/86 & 678/527 \\
\texttt{den} & 92.7 & 92.7 & 15.0 & 1393/2058 & 0.27 & 78/92 & 910/720 \\
\texttt{empty16} & 62.4 & 58.6 & 5.1 & 298/328 & 0.49 & 61/66 & 534/516 \\
\texttt{empty256} & 118.6 & 118.5 & 4.0 & 479/701 & 0.29 & 102/118 & 1263/932 \\
\texttt{maze} & 73.3 & 68.9 & 13.5 & 927/1270 & 0.35 & 66/74 & 796/646 \\
\texttt{random} & 135.4 & 135.1 & 9.1 & 1225/1754 & 0.30 & 110/135 & 1487/1224 \\
\texttt{wareh.} & 55.2 & 55.2 & 10.7 & 589/822 & 0.31 & 56/55 & 367/260 \\
\midrule
A & 49.9 & 42.2 & 28.2 & 1193/1641 & 0.41 & 43/54 & 297/315 \\
B & 19.0 & 17.9 & 151.2 & 2707/3321 & 0.42 & 17/19 & 78/79 \\
C & 10.4 & 6.0 & 28.3 & 169/150 & 0.78 & 16/16 & 66/64 \\
D & 4.3 & 1.9 & 1.0 & 2/3 & 0.63 & 4/4 & 15/15 \\
E & 23.5 & 18.7 & 12.5 & 234/285 & 0.52 & 22/28 & 128/141 \\
\bottomrule
\end{tabular}}
\end{table}

\subsection{Results and Discussion}

When a node is expanded, Lazy-CPBS materializes the child it pops next and
leaves its sibling unmaterialized on the stack, whereas eager search
materializes both. A sibling that is popped later, upon backtracking, is
materialized by both algorithms. The savings therefore come from the siblings
that are never popped, of which there is at most one per level of the branch
from the root to the solution node, so $\Delta_{\mathrm{m}}\le d$.

On the benchmark roadmaps, backtracking is rare and $\Delta_{\mathrm{m}}$
almost coincides with $d$. On the industrial layouts, which are denser,
backtracking is more frequent and $\Delta_{\mathrm{m}}$ falls below $d$. In all
settings but one (layout C), $\Delta t^{\mathrm{pred}}$ underestimates the measured saving.
The reason is that the skipped sibling is the child with the longer
replan list, so its materialization costs more than the average
$\bar t_{\mathrm{m}}$. Overall, lazy materialization reduces the runtime by a
factor of 2--4 on the benchmark roadmaps and by 22--59\% on the industrial layouts.
The gain is smaller on the industrial layouts because backtracking and failed materializations
reduce the share of materializations that can be saved: $\Delta_{\mathrm{m}}$ is about half of
the eager materializations on the benchmark roadmaps, but only 25--47\% on the industrial layouts.

With eager materialization, SoC ordering expands fewer nodes than $\mathrm{ReplanList}$
ordering in most settings, up to about 20\% fewer, i.e., it selects slightly better
children. However, the child with the lower SoC often requires replanning a
larger part of the pDAG. On the benchmark roadmaps, SoC ordering therefore
performs up to 41\% more SIPP calls, each more expensive on average, and CPBS is
1.5--2.6$\times$ slower than Eager-RL. The only exception is \texttt{empty16},
where the two orderings are on par. On the industrial layouts, the two orderings
differ by approximately 10\% in SIPP calls and 11\% in runtime.

Hence, on the benchmark roadmaps the weaker guidance of $\mathrm{ReplanList}$ ordering is more than
compensated by cheaper materializations, and it is the only one of the two orderings that is
compatible with lazy materialization. There, the speed-up of Lazy-CPBS over CPBS combines lazy
materialization, the larger contribution, and the cheaper child ordering. On the industrial layouts,
it stems from lazy materialization alone: its absolute size is governed by
$\Delta_{\mathrm{m}}\,\bar t_{\mathrm{m}}$, and its relative size by the share of materializations saved.

\section{Conclusion} \label{sec:conclu}

In this paper, we studied priority-based continuous-time MAPF$_R$ on industrial AGV roadmaps. We first revisited the algorithmic foundations of this setting by analyzing the interaction between priority-induced reservations and the SIPP low-level planner. This clarified the SIPP boundary conditions required by priority-induced reservations and enabled earlier solution detection in CPBS.

As a key contribution, we analyzed the priority-based search space induced by partial precedence relations among agents. The resulting structural properties clarify how priority decisions propagate through the search, which agents are affected by a refinement, and why different branches of the priority space can be treated independently. Beyond their theoretical interest, these properties provide concrete guidance for engineering more efficient priority-based algorithms, including parallel high-level exploration.
Building on this analysis, we introduced Lazy-CPBS, a novel algorithm that delays node materialization until expansion. By avoiding redundant low-level replanning, Lazy-CPBS improves the practical scalability of priority-based continuous-time planning while preserving the robustness benefits of systematic priority revision under congestion. An ablation study showed that these savings are bounded by the depth of the solution in the priority tree, and that the replan-list ordering required by laziness is itself cheaper than SoC ordering on the benchmark roadmaps.

We grounded the proposed approach in industrial AGV coordination by formalizing AGV-MAPF$_R$-CTC and evaluating the methods on both
standard MAPF benchmarks and industrial AGV roadmaps imported from a proprietary digital-twin simulator. The results show that priority-based MAPF$_R$ solvers provide a strong robustness/runtime compromise and that Lazy-CPBS is a practical step toward centralized AGV fleet coordination on realistic warehouse roadmaps.

Future research will focus on further accelerating CPBS and Lazy-CPBS through structural speed-up techniques, including the design of zone-distributed variants to better exploit large and highly redundant roadmaps. We also plan to study theoretical guarantees of completeness and to develop a structurally efficient multi-threaded implementation. In addition, we aim to extend CPBS and Lazy-CPBS to a lifelong planning setting and integrate them into a novel AGV traffic manager capable of handling continuous streams of tasks under real-time industrial constraints. Taken together, these directions aim to further narrow the gap between theory and practice, advancing both the academic foundations and the industrial applicability of continuous-time MAPF.

\normalsize
\bibliography{references}

\end{document}